\documentclass[12pt]{article}

\usepackage{mathtools}
\usepackage{booktabs}
\usepackage[english]{babel} % English language/hyphenation
\usepackage{amsmath,bm}
\usepackage{bbm}
\usepackage{ amssymb }
\usepackage{graphicx}
\usepackage{array}
\usepackage{multirow}
\usepackage{hhline}
\usepackage[titletoc]{appendix}
\usepackage{subfigure}
\usepackage[normalem]{ulem}

\usepackage{color}
\usepackage[CJKbookmarks=true,
            bookmarksnumbered=true,
			bookmarksopen=true,
            colorlinks=true,
			citecolor=blue,
			linkcolor=blue,
			anchorcolor=red,
			urlcolor=blue]{hyperref}

\usepackage{eqparbox}
\usepackage{footnote}

\usepackage{smileXX}

\makeatletter
\renewcommand*{\thanks}[1]{%
  \footnotemark
  \protected@xdef\@thanks{\@thanks
    \protect\footnotetext[\arabic{footnote}]{#1}}%
}
\makeatother

\usepackage{setspace}

\usepackage{algpseudocode}

\usepackage{eqparbox}

\usepackage{tabularx}
\usepackage[font = small,labelfont=bf,textfont=it]{caption} % Custom captions under/above floats in tables or figures
\usepackage{footnote}
\usepackage{algorithm}% http://ctan.org/pkg/algorithms

\usepackage[dvipsnames]{xcolor}

\makeatletter
\renewcommand*{\thanks}[1]{%
  \footnotemark
  \protected@xdef\@thanks{\@thanks
    \protect\footnotetext[\arabic{footnote}]{#1}}%
}
\makeatother

\usepackage[utf8]{inputenc}
\usepackage{mathtools}                                  
\usepackage{booktabs}
\usepackage[english]{babel} % English language/hyphenation
\usepackage{amsmath,amsthm}
\usepackage{ amssymb }
\usepackage{graphicx}
\usepackage{array}
\usepackage{multirow}
\usepackage{hhline}
\usepackage[titletoc]{appendix}

\usepackage[margin=1in,hmarginratio=1:1,top=20mm,columnsep=20pt]{geometry} % Document margins
\usepackage{booktabs} % Horizontal rules in tables
\usepackage{natbib}
\usepackage{setspace}

\usepackage{graphicx}
\usepackage{subfigure}

\usepackage{footnote}
\newtheorem{lemma}{Lemma}[section]
\newtheorem{definition}{Definition}[section]
\newtheorem{assumption}{Assumption}[section]

\newtheorem{remark}{Remark}[section]
\newtheorem{theorem}{Theorem}[section]
\newtheorem{example}{Example}[section]

\newcommand{\BlackBox}{\rule{1.5ex}{1.5ex}}  % end of proof

\begin{document}

\title{Random Smoothing Regularization in Kernel Gradient Descent Learning}

\author{ Liang Ding\thanks{The authors' names are sorted alphabetically. Corresponding author: Wenjia Wang} \\
        Fudan University
       \and 
       Tianyang Hu  \\  Purdue University
       \and
       Jiahang Jiang \\  The Hong Kong University of Science and Technology
       \and
       Donghao Li  \\ The Hong Kong University of Science and Technology 
       \and 
       Wenjia Wang \\ The Hong Kong University of Science and Technology (Guangzhou)\\ and The Hong Kong University of Science and Technology 
       \and
       Yuan Yao \\ The Hong Kong University of Science and Technology }

\maketitle

\begin{abstract}
Random smoothing data augmentation is a unique form of regularization that can prevent overfitting by introducing noise to the input data, encouraging the model to learn more generalized features. Despite its success in various applications, there has been a lack of systematic study on the regularization ability of random smoothing. In this paper, we aim to bridge this gap by presenting a framework for random smoothing regularization that can adaptively and effectively learn a wide range of ground truth functions belonging to the classical Sobolev spaces. Specifically, we investigate two underlying function spaces: the Sobolev space of low intrinsic dimension, which includes the Sobolev space in $D$-dimensional Euclidean space or low-dimensional sub-manifolds as special cases, and the mixed smooth Sobolev space with a tensor structure. By using random smoothing regularization as novel convolution-based smoothing kernels, we can attain optimal convergence rates in these cases using a kernel gradient descent algorithm, either with early stopping or weight decay. It is noteworthy that our estimator can adapt to the structural assumptions of the underlying data and avoid the curse of dimensionality. This is achieved through various choices of injected noise distributions such as Gaussian, Laplace, or general polynomial noises, allowing for broad adaptation to the aforementioned structural assumptions of the underlying data. The convergence rate depends only on the effective dimension, which may be significantly smaller than the actual data dimension. We conduct numerical experiments on simulated data to validate our theoretical results. 
%    In nonparametric regression with kernel methods, statistical optimality demands delicate control of the smoothness of the kernel function so that it matches that of the ground truth.
    % In case of a mismatch, regularization is mandatory, e.g., ridge penalty when the kernel is over-smooth, or early stopping when the opposite. 
    % However, different means of regularization lack synergy, each with its own restrictions, and there is no universal solution that can provably handle all cases. 
%    In this work, inspired by the empirical success of random smoothing data augmentation,
    % \yy{`random smoothing data augmentation' is more specific?}
 %   we propose a unified framework based on random smoothing data augmentation that can effectively and adaptively learn a wide range of ground truth functions. 
\end{abstract}

% \tableofcontents

\section{Introduction}\label{sec:intro}

Random smoothing data augmentation is a technique used to improve the generalization and robustness of machine learning models, particularly in the context of deep learning. This method involves adding random noise, such as Gaussian or Laplace noise, to the input data during the training process. %By introducing such noise, the model is encouraged to learn more generalized representations of the data, making it less likely to overfit to the training set and more capable of handling unseen data. 
The idea behind random smoothing is to make the model more robust to small perturbations in the input data, as the added noise simulates variations that may occur naturally in real-world data. This augmentation approach has proven to be an effective regularization technique, contributing to the empirical success of deep learning models across various applications. For instance, random flip, random crop, and color jitter can significantly improve the classification accuracy in natural images \citep{goodfellow2016deep,shorten2019survey}. Random smoothing has been proven effective for improving model robustness and generalization \citep{Blum20,rosenfeld2020certified,mehra2021robust, wang2020certifying,gao2020certified}. For example, random smoothing with Gaussian noise injection is introduced to address the adversarial vulnerability \citep{cohen2019certified,salman2019provably}, and by encouraging the feature map to be invariant under data augmentations, self-supervised contrastive learning methods \citep{he2020momentum,chen2020simple,grill2020bootstrap,chen2021exploring, he2021masked} can achieve state-of-the-art performance for various downstream tasks.

Random smoothing can be viewed as a form of regularization \citep{grandvalet1997noise}. %because it helps prevent overfitting by introducing noise to the input data, encouraging the model to learn more generalized features. 
Regularization techniques generally aim to reduce the complexity of a model, making it less prone to fitting the noise in the training data and, consequently, improving its performance on unseen data. Random smoothing can be considered an implicit form of regularization, as it does not directly modify the model’s parameters or loss function, unlike explicit regularization techniques such as $\ell_1$ or $\ell_2$ regularization. Instead, it indirectly influences the model’s behavior by altering the input data during training. By adding random noise to the input data, random smoothing forces the model to focus on the underlying structure of the data rather than memorizing specific instances. This leads to more robust and generalizable models that can better handle variations in real-world data. As a result, random smoothing acts as a regularizer, improving the model’s ability to generalize from the training set to unseen data.  Such a regularization perspective at least starts with  \cite{grandvalet1997noise}. However, in spite of the empirical success of random smoothing in various applications, there is a lack of systematic research on the regularization effect of random smoothing in the literature. 

In this paper, we address this gap by examining the classic nonparametric regression problem from the perspective of random smoothing regularization. 
% In supervised learning, the goal is to understand the function relationship between the input and output. Kernel ridge regression is one prominent supervised learning method which has been extensively studied in the literature (cite???). It has been shown that the optimal convergence rate of the kernel ridge regression depends on both the smoothness of the underlying function, denoted by $m_f$, and the smoothness of the kernel, denoted by $m$. Specifically, it has been shown that if $m \in [m_f/2,\infty)$, with appropriate regularization strength, the optimal convergence rate under $L_2$ metric can be obtained, which is $n^{-m_f/(2m_f+D)}$, where $n$ is the sample size, and $D$ is the dimension. However, if we choose the kernel with the smoothness less than $m_f/2$, it has not been shown whether the optimal convergence rate can be obtained, while a less smooth kernel is ubiquitous in practice.
% nonparametric regression
In nonparametric regression, the primary objective is to uncover the functional relationship between input and output variables. By making appropriate assumptions about the underlying truth function and selecting the appropriate estimator, we focus on understanding the efficiency of the estimation, specifically, the rate at which the estimation error converges to zero as the sample size $n$ increases.
%In nonparametric regression, the goal is to recover the functional relationship between the input and output. With suitable assumptions on the ground truth function and proper choices of the estimator, we are interested in the estimation efficiency, i.e., how fast the estimation error converges to zero with the sample size. 
% statistical optimality? 
The optimal convergence rate is typically dictated by the problem’s inherent complexity. The actual achievable convergence rates depend on the specific estimation methods employed. 
%The optimal convergence is usually determined by the problem itself. The more complicated the ground truth, or the higher the dimension, the slower the best possible rate. 
%The actual achievable convergence rates are up to the estimation methods. 
% kernel methods
%Among others, kernel ridge regression is a prominent family of kernel-based methods that have been extensively studied in the literature
Among various techniques, we consider kernel methods that have been extensively investigated in the research literature \citep{wahba1990spline,els}.

In this study, we present a unified framework that can learn a wide range of $D$-dimensional ground truth functions belonging to the classical Sobolev spaces ($\cW^{m_f}$) in an effective and adaptive manner.
The framework incorporates random smoothing as a central component. Our hypothesis space is a reproducing kernel Hilbert space that is associated with a kernel function of smoothness denoted by $m_0$. Random smoothing regularization leads to a novel convolution between the kernel function and a probability density function for the injected input noise. This injected noise is governed by either short or long-tail distributions, namely Gaussian and polynomial (including Laplace) noises, respectively. 
The resulting convolution-based random smoothing kernel enables us to adapt to the smoothness of the target functions more efficiently. Notably, we establish that for any $m_0$ and $m_f$ greater than $D/2$, optimal convergence rates can be achieved by utilizing random smoothing regularization and appropriate early stopping and/or weight decay techniques. 

To be specific, we investigate two possible function spaces that may contain the target function. In Section \ref{sec:fullsobo}, we analyze the Sobolev space with a low intrinsic dimension, which is denoted by $d$. This space covers both $D$-dimensional Euclidean spaces (when $d = D$) and low-dimensional sub-manifolds as specific examples. In Section \ref{sec:mixsobo}, we explore the mixed smooth Sobolev spaces, which possess a tensor structure. Our principal findings are summarized below.

\begin{itemize}
    \item In case of Sobolev space of low intrinsic dimensionality $d\leq D$: 
    \subitem When using Gaussian random smoothing, an upper bound of the convergence rate is achieved at $n^{-m_f/(2m_f+d)}(\log n)^{D+1}$, which recovers the results presented in \cite{hamm2021adaptive} and is hypothetically optimal up to a logarithmic factor. However, in contrast to \cite{hamm2021adaptive}, we present a different approach that allows us to analyze polynomial smoothing;
    \subitem When using polynomial random smoothing with data size adaptive smoothing degree, a convergence rate of $n^{-m_f/(2m_f+d)}(\log n)^{2m_f+1}$ is achieved, which is again, hypothetically optimal up to a logarithmic factor. %These two convergence rates are  in \cite{hamm2021adaptive}.
    \item In case of mixed smooth Sobolev spaces, using polynomial random smoothing of degree $m_\varepsilon$, a fast convergence rate of $n^{-2m_f/(2m_f + 1)}(\log n)^{\frac{2m_f}{2m_f+1}\left(D-1+\frac{1}{2(m_0+m_\varepsilon)}\right)}$ is achieved, which is optimal up to a logarithmic factor.
\end{itemize}

To the best of our knowledge, such results have not been studied in the literature so far. They have various implications below. 

First of all, these results enhance the convergence rates in the context of kernel ridge regression by incorporating random smoothing data augmentation with two other popular techniques, early stopping and weight decay.
% match smoothness
In kernel ridge regression, it is crucial to balance the smoothness of the kernel function ($m_0$) with that of the ground truth ($m_f$). In practice, it is common for $m_0$ to be unequal to $m_f$.  In cases of mismatch, regularization becomes essential. Specifically, if $m_0 \in [m_f/2,\infty)$, the optimal convergence rate $n^{-m_f/(2m_f+D)}$ can be achieved by employing an appropriate ridge penalty strength. This result can be generalized to low intrinsic dimensionality $d\leq D$, where the hypothetically optimal convergence rate is $n^{-m_f/(2m_f+d)}$ \citep{hamm2021adaptive}. However, when the chosen kernel has a smoothness $m_0$ less than $m_f/2$, the optimal adaptation is not well studied in kernel ridge regression. In contrast, our findings demonstrate optimal adaptation for arbitrary $m_0$ and $m_f\geq D/2$ without such a constraint. This highlights the broad adaptation ability of random smoothing regularization. 
%\yy{I have some concerns about the discussions on `saturation issue' as the motivation. First, saturation is about integral operator approach for function complexity, i.e. the ground truth $f_\rho \in L_K^r(B_R(L_2))$. Second, it is about when $r>0$, KRR only achieves $\|f_\lambda - f_\rho\|_2\leq O(\lambda^{\min(r,1)})$, while early stopping achieves $\|f_t - f_\rho\|_2 \leq O(t^{-r})$ for all $r>0$. In particular, when $r>1$, KRR approximation error saturates at $\lambda^{1}$ while early stopping goes $t^{-r}$ for large $r$, where $t\sim 1/\lambda$. In cases that $r=s/D=m_f/D$, this is about when $m_f>D$, the optimal approximation rates of KRR might be worse than early stopping. However, it seems that one can always adapt $m_0\in[m_f/2,\infty)$ such that optimal rates are possible for KRR as above. Therefore, \emph{saturation} might not be the best selling point for this work? What do you think? }\ww{What we originally thought is that even with a fixed low smooth kernel, random smoothing can still achieve fast (and optimal) convergence rate in various scenarios (that is why we used NTK). However, I agree with you that this may not be the best selling point since early stopping can also achieve fast convergence rate with low smooth kernel, and random smoothing regularization is better. I think what we might sell is that random smoothing can adapt to wide range of smoothness and various scenarios as you mentioned?}

% link to neural networks
Moreover, the optimal adaptation of polynomial random smoothing has an implication for neural networks via the (generalized) Laplace random smoothing. It is known that the training of neural networks, with enough overparametrization, can be characterized by kernel methods with a special family of kernels called the ``neural tangent kernel'' (NTK). Due to the low smoothness of the ReLU activation function, the corresponding NTK also has a low smoothness that is the same as a Laplace kernel \citep{chen2020deep, geifman2020similarity}. To the best of our knowledge, the estimation error is at the rate $n^{-\frac{D}{2D-1}}$ \citep{hu2021regularization}. Our results, using the polynomial random smoothing with (generalized) Laplace distributions, show that the convergence rate can be improved, which sheds light on understanding non-smooth augmentations such as random crop and mask. Based on this understanding, 
% which prevents the DNN estimator to achieve the optimal convergence rate. 
% Even for a smooth function, to the best of our knowledge, the estimation error is at the rate $n^{-\frac{d}{2d-1}}$ \citep{hu2021regularization}, which is far from the optimal convergence rate $n^{-m_f/(2m_f+D)}$ if $m_f$ is large.
numerical experiments with neural networks are conducted
on simulated data to corroborate our theoretical results.

Finally, it is worth mentioning that with random smoothing, the convergence rates mentioned above can be obtained by early stopping. However, if one applies weight decay, the number of iterations can be reduced from polynomial$(n)$ to polynomial$(\log n)$. Additionally, our estimator can adapt to the low-dimensional assumptions mentioned earlier, as the convergence rates depend on $D$ at most logarithmically, alleviating the curse of dimensionality. It is also important to note that we do not employ the spectrum of integral operator technique \citep{yao2007early,lin2016iterative,lin2017optimal}, but instead use Fourier analysis, which provides a universal basis for kernels of different smoothness, and avoids imposing conditions on the eigenvalues and eigenfunctions of the kernel function. This is because there is no clear relationship between the low intrinsic dimension and the eigenvalues of the integral operator. Furthermore, our theoretical analysis can be applied to the widely used Mat\'ern kernel functions.

% If one choose using weight decay instead of early-stopping, we show that for a fixed kernel with any smoothness, the optimal convergence rate can still be 

% obtained by utilizing proper data augmentation (adding noise) and early stopping. 

% \ww{add treatment}

% It should be noted that (state our difference with previous works, like different proof techniques and general noise)

The remainder of this paper is structured as follows. In Section \ref{sec_related_works}, we provide a review of related works. Section \ref{sec:settings} introduces the settings considered in this work, which include early stopping with a random smoothing kernel, as well as the conditions and assumptions utilized in this work. The main theoretical results are presented in Section \ref{sec_theory}, and numerical studies are conducted in Section \ref{sec:num}. Conclusions and a discussion are provided in Section \ref{sec:conclusion}. Technical proofs are included in the Appendix.

%The rest of this paper is arranged as follows. In Section \ref{sec_related_works}, we review some related works. Section \ref{sec:settings} introduces the settings we consider in this work, i.e., early stopping with random smoothing kernel, as well as conditions and assumptions used in this work. Main theoretical results are presented in Section \ref{sec_theory} and numerical studies are conducted in Section \ref{sec:num}. Conclusions and discussion are made in Section \ref{sec:conclusion}. Technical proofs are provided in Appendix.

% Add a figure of cifar10 showing the motivation here.

\section{Related Works}\label{sec_related_works}
% kernel regression and regularization
Various means of regularization have been proposed for kernel methods to better recover the underlying function, among which, ridge penalty and early stopping are the most popular.
% ridge penalty and early stopping
Kernel ridge regression has been extensively studied in the literature, see \cite{blanchard2018optimal,dicker2017kernel,guo2017learning,lin2017distributed,steinwart2009optimal,tuo2020improved,wu2006learning} for example. 
Early stopping treats the number of training iterations as a hyperparameter in the optimization process, which has been extensively studied by the applied mathematics community \citep{dieuleveut2016nonparametric, yao2007early,pillaud2018statistical,raskutti2014early}. Various forms of early stopping also have been studied including boosting \citep{ZhaYu02,bartlett2007adaboost}, conjugate gradient algorithm \citep{blanchard2016convergence} and kernel gradient descent \citep{BuhYu02,caponnetto2006adaptation,yao2007early,wei2017early,lin2016iterative}.
Some works (e.g. \cite{lin2016iterative,lin2017optimal,pillaud2018statistical}) have explored early stopping by employing the integral operator induced by the kernel, imposing conditions on the eigenvalues and eigenfunctions of the kernel function. Smoothness or regularity of functions thus implicitly depends on the measure that defines the spectrum of the integral operator, whereas classical smoothness like Sobolev spaces is not explicitly handled.

In kernel regression with gradient descent, \cite{raskutti2014early} showed that early stopping and ridge penalty both can achieve the optimal convergence rate if the smoothness is well-specified. Yet, kernel ridge regression might suffer the ``saturation issues'' while early stopping does not \citep{EngHanNeu96,yao2007early}. 
% curse of dimensionality, tensor RKHS
In regression problems, it is usually assumed that the domain of interest has a positive Lebesgue measure, while in practice, the data generating distribution is supported on some low-dimensional smooth sub-manifold \citep{scott2006minimax,yang2016bayesian,ye2008learning,ye2009svm,hamm2021intrinsic,hamm2021adaptive}. 
Kernel methods can circumvent the curse of dimensionality and adapt to various low-dimensional assumptions of the underlying function. 
In particular, \cite{hamm2021intrinsic,hamm2021adaptive} generalized the manifold assumption by applying the box-counting dimension of the support of the data distribution, and derived upper bounds on the convergence rate of the prediction error. 
Another simplifying assumption is tensor product kernels \citep{gretton2015simpler,szabo2017characteristic}, whose product forms allow efficient computation of Gaussian process regression \citep{saatcci2012scalable,wilson2015kernel,ding2022sample,chen2022kernel} and analysis of independent component \citep{bach2002kernel, gretton2005measuring,gretton2007kernel}. 
The RKHS induced by a tensor product kernel is simply tensored RKHS \citep{paulsen2016introduction}. Tensor product kernels we consider induce the tensored Sobolev spaces \citep{dung2018hyperbolic}.  

% relate to deep learning's empirical success -> NTK, training neural networks
For complicated high-dimensional data, deep learning models seem to perform extremely well, which has sparked numerous investigations into their generalization ability. As it turns out, the training of neural networks has deep connections to kernel methods with neural tangent kernels (NTK).  
Under proper initialization, training sufficiently wide DNN with gradient descent equates to kernel regression using NTK.
First introduced by \cite{jacot2018neural}, the correspondence has been significantly extended \citep{du2018gradient,li2018learning,arora2019fine,cao2020generalization, arora2019harnessing, li2019enhanced, huang2020deep, kanoh2021neural, hu2022understanding}.
From the NTK point of view, ridge penalty and early stopping are also vital in training neural networks. The former is equivalent to weight decay \citep{hu2021regularization}, which is applied by default in training deep learning models for better generalization, so is early stopping \citep{prechelt1998early}.
\cite{zhang2021understanding, hardt2016train} revealed that longer training can harm the generalization performance of deep models. 
\cite{li2020gradient, bai2021understanding} utilized early stopping to improve robustness to label noises. 

% 5. random smoothing and data augmentation/noise injection
% noise in inputs (?)
Besides NTK, various data augmentation techniques in deep learning that are proven effective in improving model generalization can also provide inspiration for kernel methods. 
\cite{grandvalet1997noise} studied from a regularization perspective how noise injection can improve generalization. 
Data augmentation is particularly important for handling natural images \citep{shorten2019survey}, where horizontal flip, random crop, color jitter can significantly improve the classification accuracy. 
By applying the above augmentations, self-supervised contrastive learning methods \citep{he2020momentum,chen2020simple,grill2020bootstrap,chen2021exploring, he2021masked} can achieve state-of-the-art performance for various downstream tasks. 
Randomized smoothing \citep{cohen2019certified,salman2019provably} is a special data augmentation, first proposed to address the adversarial vulnerability \citep{goodfellow2014explaining,carlini2017towards} of deep learning models. The key idea is to perturb the input with random noise injection and make predictions by aggregating the outputs from all augmented inputs. 
Random smoothing has been proven effective 
for improving model robustness and generalization \citep{rosenfeld2020certified,mehra2021robust, wang2020certifying,gao2020certified}. 
% Deep learning models are known to be vulnerable to adversarial attacks, where injecting human-imperceptible noise into the model input significantly changes model prediction \citep{goodfellow2014explaining,carlini2017towards}. 
% Random smoothing is the state-of-the-art method for $l_2$ certified defense against adversarial attacks \citep{cohen2019certified,salman2019provably}, which guarantees that the model prediction is constant when the adversarial noise is within $l_2$ balls. Specifically, random smoothing replaces the original classifier $f$ with a smoothed classifier $g$, which predicts the majority class when the model input $x$ is perturbed with noise $\bvarepsilon$. That is, $g(x)=\arg\max_{c\in y} \mathbb{P}(f(x+\bvarepsilon)=c )$. When $\bvarepsilon$ is isotropic Gaussian noise, \cite{lecuyer2019certified} proved that $g$ is certified $l_2$ robust, and \cite{cohen2019certified} improved the result by giving a tight bound for the certified radius. Apart from certified robustness against adversarial examples, random smoothing has also been wildly used to defend other attacks against machine learning models, including data poisoning attacks \citep{rosenfeld2020certified,mehra2021robust}, backdoor attacks \citep{wang2020certifying}, and topology attacks \citep{gao2020certified}. 
Our proposed framework incorporates random smoothing, together with weight decay and early stopping, to provide a unified solution for the smoothness mismatch problem in kernel regression. 
%
% Noise injection has also been widely used in deep generative models. In training generative adversarial networks (GAN) \citep{goodfellow2020generative}, instance noise has been proven useful for stabilizing the training process \citep{jenni2019stabilizing, zhao2020image, feng2021understanding}. \cite{liang2018HowWell} showed that for density estimation, GAN without noise injection can smooth the empirical data distribution and can achieve a faster convergence rate. In denoising score matching \citep{Song2019dsm}, multi-scale Gaussian noise injection is considered to make Fisher divergence minimization more efficient. Such a denoising score-matching training approach gave rise to the booming of a new type of generative model called diffusion probabilistic models \citep{ho2020denoising}.  
%
% 6. recent works most related to this work?
It is worth clarifying the difference between our method and the ``errors in variables" literature \citep{zhou2019gaussian,wang2022gaussian,cressie2003spatial,cervone2015gaussian}. 
Though the formulations seem similar, i.e., the inputs in both cases are corrupted with noises, the two are fundamentally different. 
In our setting, both the input $\bx$ and added noise $\bvarepsilon$ are known (we control the noises in our estimator) while in the other setting, the input is noisy and only $\bx+\bvarepsilon$ is observed. 

% the input variables are also corrupted by noise. Noisy or uncertain inputs are quite common in spatial statistics, because geostatistical data are often indexed by imprecise locations.

% Other studies also draw connections between over-parametrized neural networks and NTKs: \citep{li2019enhanced} and \citep{huang2020deep} define similar kernels for convolutional and residual networks, respectively; \citep{du2018gradient,li2018learning,arora2019fine} focus on one-hidden-layer ReLu neural networks and conclude that with enough overparametrized DNN, model parameters and corresponding NTK do not move far away from its initialization for all iterations. 

% Yu's paper does not consider misspecification. The kernel space may be smaller than the function smoothness.

% Random smoothing has been wildly used by machine learning community to defence different attacks against machine learning models, including adversarial example attacks \citep{cohen2019certified,salman2019provably}, data poisoning attacks \citep{rosenfeld2020certified,mehra2021robust}, backdoor attacks \citep{wang2020certifying}, and topology attacks \citep{gao2020certified}. 

% In Neural Radiance Fields (NeRF), \cite{tancik2020fourier,mildenhall2021nerf}

% \section{\yy{\sout{Problem Settings and}} Methodology}\label{sec:settings}

\section{Random Smoothing Kernel Regression}\label{sec:settings}

In this section, we introduce the problem of interest, our methodology, and the necessary conditions used in this work.

% {\bf Can we just consider adding Gaussian noise?} We can consider many kinds of noise. Not all augmentations are smooth.

% {\bf Change notation  $d\rightarrow D$; $\rho \rightarrow d$}

% \subsection{\yy{Problem Setting \sout{Early Stopping with Random Smoothing Kernel}}}

\subsection{{Problem Setting}}\label{subsec:ProblemSettings}

Suppose we have observed data $(\bx_j,y_j)$ for $j=1,...,n$, which follows the relationship given by
\begin{align}\label{eq:model}
    y_j = f^*(\bx_j)+\epsilon_j.
\end{align}
Here, $\bx_j$'s are independent and identically distributed (i.i.d.) following a marginal distribution $P_\Xb$ with support supp$(P_\Xb) = \Omega \subset\RR^D$. The function $f^* \in \cH(\Omega)$, where $\cH(\Omega)$ denotes a function space, and $\epsilon_j$'s are i.i.d. noise variables with mean zero and finite variance. Our objective is to recover the function $f^*$ based on the noisy observations.
% where $\bx_j$'s are independent and identically distributed (i.i.d.) following a marginal distribution $P_\Xb$, with supp$(P_\Xb) = \Omega \subset\RR^D$, $f^* \in \cH(\Omega)$ with $\cH(\Omega)$ denoting a function space, and $\epsilon_j$'s are i.i.d. noise with mean zero and finite variance. 
% The underlying function $f^*$ is assumed to lie in a general function space $\cH(\Omega)$. 
% Our goal is to recover the function $f^*$ based on noisy observations.

In this work, we consider two cases. In the first case (Section \ref{sec:fullsobo}), the function space $\cH(\Omega)$ is a Sobolev space with smoothness $m$, denoted by $\cW^{m}(\Omega)$, and the data is of low intrinsic dimension. In the second case (Section \ref{sec:mixsobo}), the function space $\cH(\Omega)$ is a tensor Sobolev space. Throughout this work, we assume without loss of generality that $P_\Xb$ follows a uniform distribution. Note that our theoretical analysis can be easily extended to the case where $P_\Xb$ is upper and lower bounded by positive constants.
% In this work, we consider two cases, where the function space $\cH(\Omega)$ can be a Sobolev space with smoothness $m$, denoted by $\cW^{m}(\Omega)$, with data on low intrinsic dimension (Section \ref{sec:fullsobo}), or a tensor Sobolev space (Section \ref{sec:mixsobo}). 
% Without loss of generality, we assume $P_\Xb$ to be uniform distribution throughout this work. It should be noted that our theoretical analysis can be easily generalized to the case that $P_\Xb$ is upper and lower bounded by some positive constants. 

In order to recover the function $f^*$, we use reproducing kernel Hilbert spaces (RKHSs). We briefly introduce the RKHSs and their relationship with Sobolev spaces in the following, and refer to \cite{wendland2004scattered} and \cite{adams2003sobolev} for details. Let $K:\Omega \times \Omega \rightarrow \RR$ be a symmetric positive definite kernel function. Define the linear space
\begin{eqnarray}\label{FPhi}
F_{K}(\Omega)=\left\{\sum_{k=1}^n\beta_k K(\cdot,\bx_k):\beta_k\in \RR,\bx_k\in \Omega,n\in\mathbb{N}\right\},
\end{eqnarray}
and equip this space with the bilinear form
\begin{eqnarray*}
\left\langle\sum_{k=1}^n\beta_k K(\cdot,\bx_k),\sum_{j=1}^m\gamma_j K(\cdot, \bx'_j)\right\rangle_K:=\sum_{k=1}^n\sum_{j=1}^m\beta_k\gamma_j K(\bx_k, \bx'_j).
%\label{eq1.2}
\end{eqnarray*}
Then the reproducing kernel Hilbert space $\cH_{K}(\Omega)$ generated by the kernel function $K$ is defined as the closure of $F_{K}(\Omega)$ under the inner product $\langle\cdot,\cdot\rangle_{K}$, and the norm of  $\cH_{K}(\Omega)$  is $\| f\|_{\cH_{K}(\Omega)}=\sqrt{\langle f,f\rangle_{\cH_{K}(\Omega)}}$, where $\langle\cdot,\cdot\rangle_{\cH_{K}(\Omega)}$ is induced by $\langle \cdot,\cdot\rangle_{K}$. 
The following theorem gives another characterization of the reproducing kernel Hilbert space when $K$ is stationary, via the Fourier transform. Our notion of the Fourier transform is
\begin{eqnarray*}
    \cF(g)(\bomega)=(2\pi)^{-D/2}\int_{\RR^D} g(\bx)e^{-i\bomega^T\bx} {\rm d}\bx,
\end{eqnarray*}
for a function $g\in L_1(\RR^D)$. Note that a kernel function $K$ is said to be stationary if the value $K(\bx,\bx')$ only depends on the difference $\bx-\bx'$. Thus, we can write $K(\bx-\bx') :=K(\bx,\bx')$. 

\begin{theorem}[Theorem 10.12 of \cite{wendland2004scattered}]\label{thm:NativeSpace}
	Let $K$ be a positive definite kernel function that is stationary, continuous, and integrable in $\RR^D$. Define 
	$$\mathcal{G}:=\{f\in L_2(\RR^D)\cap C(\RR^D):\mathcal{F}(f)/\sqrt{\mathcal{F}(K)}\in L_2(\RR^D)\},$$
	with the inner product
	$$\langle f,g\rangle_{\cH_K(\RR^D)}=(2\pi)^{-d/2}\int_{\RR^d}\frac{\mathcal{F}(f)(\bomega)\overline{\mathcal{F}(g)(\bomega)}}{\mathcal{F}(K)(\bomega)}{\rm d} \bomega.$$ Then $\mathcal{G} = \cH_K(\RR^D)$, and both inner products coincide.
\end{theorem}

For $m>D/2$, the (fractional) Sobolev norm for function $g$ on $\mathbb{R}^D$ is defined by
\begin{align}\label{Sobolev}
\|g\|^2_{\cW^m(\mathbb{R}^D)}=\int_{\mathbb{R}^d} |\mathcal{F} (g)(\bomega)|^2 (1+\|\bomega\|_2^2)^m {\rm d} \bomega,
\end{align}
and the inner product of a Sobolev space $\cW^m(\mathbb{R}^D)$ is defined by
$$\langle f,g\rangle_{\cW^m(\mathbb{R}^D)}=\int_{\RR^D}\mathcal{F}(f)(\bomega)\overline{\mathcal{F}(g)(\bomega)}(1+\|\omega\|_2^2)^m {\rm d} \bomega.$$
% This definition can be naturally extended to Sobolev spaces with non-integer orders, which are commonly known as the \textit{fractional Sobolev spaces}, denoted by $H^m(\mathbb{R}^d)$ with a non-integer $m$. 
\begin{remark}\label{remarksob}
In this work, we are only interested in Sobolev spaces with $m > D/2$ because these spaces contain only continuous functions according to the Sobolev embedding theorem. 
\end{remark}

It can be shown that if $m$ is an integer, the norm defined in \eqref{Sobolev} is equivalent to that of the usual Sobolev space \citep{adams2003sobolev}. If $m$ is not an integer, then the corresponding Sobolev space is called a Bessel potential space \citep{almeida2006characterization,gurka2007bessel}. 
The Sobolev space on a region $\tilde \Omega$ with a positive Lebesgue measure can be defined via restrictions as
\begin{eqnarray*}%\label{restriction}
\|f\|_{\cW^m(\tilde \Omega)}=\inf\{\|f_E\|_{\cW^m(\RR^D)}:f_E\in \cW^m(\RR^D),f_E|_{\tilde \Omega}=f\},
\end{eqnarray*}
where $f_E|_{\tilde \Omega}$ denotes the restriction of $f_E$ to $\tilde \Omega$. 

Comparing Theorem \ref{thm:NativeSpace} and \eqref{Sobolev}, it can be seen that if
\begin{align*}
    c_1(1+\|\bomega\|_2^2)^{-m} \leq  \mathcal{F}(K)(\bomega)\leq c_2(1+\|\bomega\|_2^2)^{-m}, \forall \bomega\in\mathbb{R}^D,
\end{align*}
for some two constants $c_1,c_2>0$, then $\cW^{m}(\mathbb{R}^D)$ coincides with the reproducing kernel Hilbert space $\cH_K(\RR^D)$ with equivalent norms (also see \cite{wendland2004scattered}, Corollary 10.13). By the extension theorem \citep{devore1993besov}, $\cH_{K}(\Omega)$ also coincides with $\cW^{m}(\Omega)$, and two norms are equivalent.

% Let $\cH_K(\Omega)$ be a RKHS generated by a positive definite kernel function $K(\cdot,\cdot): \Omega\times \Omega \mapsto \RR$. For the ease of mathematical treatment, we assume that $K(\cdot,\cdot)$ is stationary, i.e., the value of $K(\bx,\bx')$ only depends on $\bx-\bx'$ for all $\bx,\bx'\in \Omega$ such that we can write $K(\bx,\bx')$ as $K(\bx-\bx')$. 

% For a brief introduction to the Sobolev spaces and RKHSs, we refer to Appendix \ref{subsec:introrkhs}. 

% With the RKHS $\cH_K(\Omega)$, the kernel ridge regression estimates $f^*$ by solving
% \begin{align}\label{KRRest}
% \hat f_K = \operatorname*{argmin}_{\hat f\in \cH_K(\Omega)}\bigg( \frac{1}{n}\sum_{k=1}^n(y_k-\hat f(\bx_k))^2 + \lambda_K \|\hat f\|^2_{\cH_K(\Omega)}\bigg),
% \end{align}
% where $\lambda_K\geq 0$ is a regularization parameter, and $\|\cdot\|_{\cH_K(\Omega)}$ is the norm of the RKHS $\cH_K(\Omega)$. If $\cH_K(\Omega) = \cH(\Omega) = \cW^{m_0}(\Omega)$ and $\Omega$ has a positive Lebesgue measure satisfying certain regularization conditions, then it has been shown in the literature \citep{wahba1990spline,geer2000empirical} that the optimal convergence rate under $L_2$ measure is $O_\PP(n^{-\frac{m_0}{m_0+D}})$. 

% \subsection{\yy{Random Smoothing Kernel Regression with Early Stopping?}}

\subsection{Random Smoothing Kernel Regression with Early Stopping}

In this study, we systematically investigate the efficiency of random smoothing data augmentation, which is a widely used technique in deep learning, in improving the estimation efficiency (i.e., convergence rate) for $f^*\in\cH(\Omega)$ without assuming any relationship between $\cH(\Omega)$ and $\cH_K(\Omega)$ and considering a wide context of $\Omega$ that may have Lebesgue measure zero. To overcome the lack of smoothness in $\cH_K(\Omega)$, we construct $N$ augmentations for each observed input point $\bx_j$ by adding i.i.d. noise $\bvarepsilon_{jk}$ with a continuous probability density function $p_\varepsilon$. We can generate $\bvarepsilon_{jk}$ independently for each $j$, or we can generate $\bvarepsilon_k$ for $k=1,...,N$, and apply them to all $\bx_j$, $j=1,...,n$ simultaneously. While the latter is easier to implement, the former is easier to theoretically justify. Due to its lower computational complexity, we only consider the latter method in this work.

% \ww{asdfk,jasldkf;jkldf}

% In this work, we do not assume any relationship between $\cH(\Omega)$ and $\cH_K(\Omega)$, and consider a wide context of $\Omega$, where it is possible that $\Omega$ has Lebesgue measure zero. 
% To overcome the lack of smoothness in $\cH_K(\Omega)$, we consider a widely used technique in deep learning --- data augmentation, and investigate whether it can improve the estimation efficiency (i.e., convergence rate) for $f^*\in\cH(\Omega)$. 
% For the ease of mathematical treatment, we consider the easiest way to construct augmented data by adding random noise. Specifically, for each observed input point $\bx_j$, we construct $N$ augmentations by $\bx_j+\bvarepsilon_{jk}$, $k=1,...,N$, where $\bvarepsilon_{jk}$ are i.i.d. noise, with continuous probability density function $p_\varepsilon$. We can generate $\bvarepsilon_{jk}$ independently for each $j$, but can also generate $\bvarepsilon_k$ for $k=1,...,N$, and apply them to all $\bx_j$, $j=1,...,n$ simultaneously. 
% The latter is easier to implement, but the mathematical development is more tricky, while the former is easier to be theoretically justified. 
% In this work, we only consider the latter one, due to a lower computational complexity. 

\begin{remark}[Adding non-smooth noise and practical data augmentation techniques]
It should be noted that we do not assume $p_\varepsilon$ to be Gaussian, and can be non-smooth.
While applying Gaussian noise is a common practice, not all data augmentation techniques involve smooth noise, such as random crop, random mask, and random flip. In this work, we investigate various types of noise, including non-smooth Laplace noise and smooth Gaussian noise.
% In this work, we consider various types of noise, from non-smooth Laplace noise to smooth Gaussian noise. 
Although adding non-smooth noise still cannot
capture the effects of complex data augmentation techniques such as random mask or random crop, we aim to use it as a tool to gain insights into the success of these more complicated data augmentations.
\end{remark}

% \yy{I'm thinking that the following part is probably moved to after the model in the beginning as motivations, Assumptions may be just before the theorems in analysis.}

With augmented data, we proceed to the estimation of the function $f^*$. For any point $\bx\in \Omega$, we obtain the estimator by computing the average of the function values evaluated at the $N$ augmented inputs. Specifically, the estimator is constructed as
\begin{align}\label{eqn:f}
     f(\bx) = \frac{1}{N}\sum_{k=1}^N h(\bx+\bvarepsilon_k)
\end{align}
for $h\in \cH_K(\Omega)$. By properties of the RKHS, $f$ as in \eqref{eqn:f} is also inside $\cH_K(\Omega)$. 
We consider the following $l_2$ loss function defined as
\begin{align}\label{eq:loss}
   L_n(f)=\frac{1}{2n}\sum_{j=1}^n\rbr{f(\bx_j)-y_j}^2,
\end{align}
or equivalently,
\begin{align*}
    L_n(h) = \frac{1}{2n}\sum_{j=1}^n \left(\frac{1}{N}\sum_{k=1}^N h(\bx_j+\bvarepsilon_k)-y_j\right)^2.
\end{align*}

\begin{remark} 
% \yy{$L_n(f)$ leads to Reproducing Kernel smoothing, while $L'_n(f)$ leads to Nadaraya-Watson kernel smoothing?}
The loss function $L_n(h)$ is slightly different from the loss function used in practice, i.e., 
\begin{align*}
    L_n'(h) = \frac{1}{2n}\sum_{j=1}^n \frac{1}{N}\sum_{k=1}^N \left(h(\bx_j+\bvarepsilon_k)-y_j\right)^2.
\end{align*}
However, it can be shown that $L_n(h)$ is close to $L_n'(h)$. To see this, note that
\begin{align}\label{eq:remLnpLn}
    L_n'(h) - L_n(h) = & \frac{1}{2n}\sum_{j=1}^n\frac{1}{2N^2}\sum_{k=1}^N\sum_{l=1}^N \left(h(\bx_j+\bvarepsilon_k)-h(\bx_j+\bvarepsilon_l)\right)^2.
\end{align}
As we will see later in Section \ref{sec_theory}, we require that the variance of $\bvarepsilon_k$ to converge to zero, which implies that the right-hand side in \eqref{eq:remLnpLn} is close to zero. 
% Nevertheless, numerical examples in ??? illustrate that the difference between $L_n(f)$ and $L_n'(f)$ is small.
\end{remark}

In order to minimize \eqref{eq:loss}, we apply the gradient descent method. Since we impose a restriction that the estimator $f$ is in the RKHS $\cH_{K}(\Omega)$, by the representer theorem, it suffices to consider the function space 
\begin{align*}
   \cF_0 = \left\{f: f(\cdot) = \sum_{j=1}^n\sum_{k=1}^N w_{jk} K(\cdot-(\bx_j+\bvarepsilon_k)), w_{jk}\in \RR\right\}.
\end{align*}
Because the number of parameters in $\cF_0$ scales as $n\times N$, which can be prohibitively large if there are too many augmentations, it is often necessary to reduce the flexibility of $\cF_0$ in order to minimize the loss function \eqref{eq:loss}. To achieve this, we consider a subspace of $\cF_0$, denoted by 
\begin{align*}
   \cF = \left\{f: f(\cdot) = \sum_{j=1}^n\sum_{k=1}^N w_j K(\cdot-(\bx_j+\bvarepsilon_k)), w_j\in \RR\right\},
\end{align*}
i.e., all the weights for the different augmented data from the same input $\bx_j$ are the same. Define an empirical random smoothing kernel function by
\begin{align}\label{eq:empirical_kernel}
    K_S(\bx_l-\bx_j) := \frac{1}{N^2}\sum_{k_1=1}^N \sum_{k_2=1}^N  K(\bx_l+\bvarepsilon_{k_1}-(\bx_j+\bvarepsilon_{k_2})),
\end{align}
whose expectation leads to the following random smoothing kernel function, which plays an important role in the convergence analysis.

\begin{definition}[Random smoothing kernel function] \label{eq:rskernel}
   The kernel function $K_S$ defined in \eqref{eq:empirical_kernel} is the empirical \textit{random smoothing} kernel function corresponding to the original kernel $K$. The expectation of $K_S$ with respect to the noise $\bvarepsilon_k$ is the convoluted kernel function $K*p_\varepsilon$, where $*$ is a convolution operator defined by
\begin{align*}
    (g_1*g_2)(\bs) = \int g_1(\bt)g_2(\bs - \bt) {\rm d}\bt,
\end{align*}
for two functions $g_1$ and $g_2$. We call the convoluted kernel function $K*p_\varepsilon$ as the random smoothing kernel function. %and its further characterization is provided in Appendix ???. 
\end{definition} 

Now we can rewrite the loss function $L_n(f)$ in \eqref{eq:loss} as 
% \yy{$\omega$ or $w$?} \ww{$w$.}
\begin{align}\label{eq_lossX}
    L_n(\bw) = \frac{1}{2}\left\|\by - \Kb \bw\right\|_2^2,
\end{align}
where $\Kb=(K_S(\bx_j-\bx_k))_{jk}$, $\bw=(w_1,...,w_n)^T$, and $\by=(y_1,...,y_n)^T$. As stated in \cite{raskutti2014early}, it is more natural to perform gradient descent on the transformed vector $\btheta = \sqrt{\Kb}\bw$, where the square root can be taken because $\Kb$ is positive (semi-)definite. Then, we apply gradient descent on the square loss \eqref{eq_lossX} with the transformed vector $\btheta$. Initialize $\btheta_{0}=\bw_0 = 0$. Taking gradient with respect to $\btheta$, direct computation shows that the gradient update is 
\begin{align}\label{eq_GD_nowd}
    \btheta_{t+1} = \btheta_t - \beta_t \left(\Kb\btheta_t-\sqrt{\Kb}\by\right),
\end{align}
where $\beta_t>0$, $t=0,1,2,\ldots$ is the learning rate (step size). With parameter $\bw_t$ obtained at the $t$-th iteration, the corresponding estimator of $f^*(\bx)$ for any point $\bx\in \Omega$ is defined by 
\begin{align}\label{eq:predictor}
     f_t(\bx) = \bw_t^T \kb(\bx),
\end{align}
where $\kb(\bx) = (K_S(\bx-\bx_1),\ldots,K_S(\bx-\bx_n))^T$.

In practice, gradient descent is often paired with weight decay \citep{krogh1992simple} to prevent overfitting and improve generalization \citep{hu2021regularization}. 
Therefore, we also consider the gradient descent with weight decay, where the parameter $\btheta$ is updated by
\begin{align}\label{eq_GD_wd}
    \btheta_{t+1} = \btheta_t - \beta_t \left(\Kb\btheta_t-\sqrt{\Kb}\by\right) - \alpha_t\btheta_t,
\end{align}
with $\alpha_t>0$, $t=0,1,2,\ldots$ being the strength of weight decay. The learning rate $\beta_t$ and weight decay parameter $\alpha_t$ can be varied with $t$, but for mathematical convenience, we assume that the step sizes $\beta_t$ and the weights decay parameter $\alpha_t$ are not related to the iteration number $t$, i.e., $\beta_t=\beta$ and $\alpha_t=\alpha$ for all $t=0,1,2,\ldots$. 
% \ww{For TH: add something saying that this is also what we use in practice?}
 
In this work, we are interested in the prediction error 
\begin{align}\label{eq:prederror}
    \|f^* - f_t\|_{L_2(P_\Xb)}.
\end{align}

% \subsection{Notation}
In the rest of this paper, the following definitions are used. For two positive sequences $a_n$ and $b_n$, we write $a_n\asymp b_n$ if, for some $C,C'>0$, $C\leq a_n/b_n \leq C'$. Similarly, we write $a_n\gtrsim b_n$ if $a_n\geq Cb_n$ for some constant $C>0$, and $a_n\lesssim b_n$ if $a_n\leq C'b_n$ for some constant $C'>0$. Also, $C,C',c_j,C_j, j\geq 0$ are generic positive constants, of which value can change from line to line. 

% \yy{In ML, usually RK uses notation $K$ instead of $K$ and basis uses $\psi$. 
% Why not use $\cH_K$ rather than $\cH_K$?} 

\section{Main Results}\label{sec_theory}

% In this section, we present our main theoretical results. 
% under two cases of $\cH(\Omega)$. 
In this section, we present our main theoretical results. We begin by collecting all the assumptions that will be used throughout the paper in Section \ref{sec:assumptions}. Then, in Section \ref{sec:fullsobo}, we consider the case where $\Omega$ has a finite intrinsic dimension. Finally, in Section \ref{sec:mixsobo}, we consider the case where $\cH(\Omega)$ is a tensor RKHS.

%We also present a comparison theorem in Section \ref{sec:connecttoKRR}, which states that the gradient descent algorithm with early stopping has a faster (at least the same) convergence rate than that of the kernel ridge regression.

\subsection{Assumptions} \label{sec:assumptions}

% \yy{Kernels in this paper: 1) Sobolev kernel $K$ for $\cW^{m_0}(\Omega)$ (why not using $K_m$?); 2) Noise kernel, e.g. Gaussian $K_\gamma = \exp(-\|x - x'\|^2 /\gamma)$ or Laplace $K_{\gamma} = \exp(-\|x - x'\|_1/\gamma) $? 3) Random smoothed/convolved kernel $\widetilde{K}= K\ast K_\gamma$? Is there a distribution $d$ on $X\times Y$ such that $supp(d_X) = \Omega$? Or just uniform measure? }

In this work, we will use the following assumptions. 
\begin{assumption}\label{assum:sub-G}
The error $\epsilon_j$'s in \eqref{eq:model} are i.i.d. sub-Gaussian \citep{geer2000empirical}, i.e., satisfying
\begin{align*}%\label{noiseCond}
	C^2 (\mathbb{E} e^{|\epsilon_j|^2/C^2}-1)\leq C', \quad j=1,...,n.
\end{align*}
% for some positive constants $C$ and $C'$. 
\end{assumption}

\begin{assumption}\label{assum:PsiDecay}
There exists $m_0 > D/2$ such that
\begin{align}\label{eq:kernelsmoothness}
    c_1(1+\|\bomega\|_2^2)^{-m_0} \leq  \mathcal{F}(K)(\bomega)\leq c_2(1+\|\bomega\|_2^2)^{-m_0}, \forall \bomega\in\mathbb{R}^D.
\end{align} 
\end{assumption}

\begin{assumption}[Tensor kernel function]\label{assum:PsiDecay_tensor}
The kernel function $K$ can be expressed as $K=\prod_{j=1}^DK_j$, where $K_j$'s are one-dimensional kernel functions. There exists $m_0>1/2$ such that for $j=1,\ldots,D$,
\begin{align}\label{eq:kernelsmoothness}
    c_1(1+\omega_j^2)^{-m_0} \leq  \mathcal{F}(K_j)(\omega)\leq c_2(1+\omega_j^2)^{-m_0}, \forall \omega_j\in\mathbb{R}.
\end{align} 
\end{assumption}

\begin{example}
    A class of kernel functions satisfying Assumption \ref{assum:PsiDecay} is the isotropic Mat\'ern kernel functions \citep{williams2006gaussian}. With reparameterization, the Mat\'ern kernel function is given by \begin{eqnarray}\label{eq_matern}
        K(\bx) = \frac{(2\phi\sqrt{m_0-D/2} \|\bx\|_2)^{m_0-D/2}}{\Gamma(m_0-D/2)2^{m_0-D/2-1}} B_{m_0-D/2}(2\phi\sqrt{m_0-D/2}\|\bx\|_2),
    \end{eqnarray}
    with the Fourier transform \citep{tuo2015theoretical}
    \begin{eqnarray}\label{maternspectral}
        \mathcal{F}(K)(\bomega) = \pi^{-D/2}\frac{\Gamma(m_0)}{\Gamma(m_0-D/2)}(4\phi^2(m_0-D/2) )^{m_0-D/2} (4\phi^2(m_0-D/2)+\|{\bomega}\|^2)^{-m_0},
    \end{eqnarray}
    where $\phi > 0$, and $B_{m_0-D/2}$ is the modified Bessel function of the second kind. It can be seen that \eqref{maternspectral} is bounded above and below by $(1+\|\bomega\|_2^2)^{-m_0}$, up to a constant multiplier.
    
    Another example satisfying Assumption \ref{assum:PsiDecay} is the generalized Wendland kernel function \citep{wendland2004scattered,gneitingstationary,chernih2014closed, bevilacqua2019estimation,fasshauer2015kernel}, defined as
    \begin{align}\label{GWcorr}
        K_{GW}(\bx) = \left\{
        \begin{array}{lc}
             \frac{1}{{\rm Beta}(2\kappa,\mu+1)}\int_{\|\phi \bx\|_2}^1 u(u^2-\|\phi \bx\|_2^2)^{\kappa - 1}(1-u)^\mu {\rm d}u, & 0\leq \|\bx\|<\frac{1}{\phi}, \\
             0, & \|\bx\|_2\geq \frac{1}{\phi},
        \end{array}\right.
    \end{align}
    where $\phi,\kappa > 0$ and $\mu \geq (D+1)/2 + \kappa$, and ${\rm Beta}$ denotes the beta function. Theorem 1 of \cite{bevilacqua2019estimation} shows that \eqref{GWcorr} satisfies Assumption \ref{assum:PsiDecay} with $m_0 = (D+1)/2 + \kappa$.

    If the kernel function $K=\prod_{j=1}^DK_j$, and each $K_j$ is a one-dimensional Mat\'ern kernel function or generalized Wendland kernel function, then Assumption \ref{assum:PsiDecay_tensor} is satisfied.
\end{example}

% , with $\cW^{m_f}(\Omega)$ the Sobolev space with smoothness $m_f$. 
% We do not assume any relationship between $m_f$ and $m_0$, but only assume $\min(m_f,m_0)>d/2$, such that all functions in $\cW^{m_f}(\Omega)$ or $\cW^{m_0}(\Omega)$ are continuous according to the Sobolev embedding theorem \citep{adams2003sobolev}.

% $m_f$ and $m_0$, but only assume $\min(m_f,m_0)>D/2$, such that all functions in $\cW^{m_f}(\Omega)$ or $\cW^{m_0}(\Omega)$ are continuous according to the Sobolev embedding theorem \citep{adams2003sobolev}.

% If $m_0<m_f/2$, then whether the optimal convergence rate can be achieved has not been discussed to the best of our knowledge. In order to improve the convergence rate, we consider a widely used technique in deep learning: data augmentation. 

\begin{assumption}[Random smoothing noise]\label{assum:augnoise}
The elements of $\bvarepsilon_k$ are i.i.d. mean zero sub-Gaussian random variables. 
% with the same distribution  {\color{red} satisfying $\PP(\|\bvarepsilon_k\|>t)\leq L_1e^{-L_2t^a}$} where $L_1,L_2,a$ are any positive constants, and has mean zero. 
Furthermore, we consider three cases of $\bvarepsilon_k$ as follows, where %\yy{smoothing scaling?} 
$\sigma_n^2$'s are positive parameters to be specified later in Section \ref{sec_theory}.
\begin{itemize}
    \item[(C1)] (Polynomial noise) There exists $m_{\varepsilon}>D/2$ such that the characteristic function of $\bvarepsilon_k$ satisfies 
    \begin{align*}
        c_1(1+\sigma_n^2\|\bomega\|_2^2)^{-m_{\varepsilon}} \leq  \EE(e^{i\bomega^T\bvarepsilon_k})\leq c_2(1+\sigma_n^2\|\bomega\|_2^2)^{-m_{\varepsilon}}, \forall \bomega\in\mathbb{R}^D.
    \end{align*}
    \item[(C2)] (Tensor Polynomial noise) There exists $m_{\varepsilon}>1/2$ such that the characteristic function of $\bvarepsilon_k$ satisfies 
    \begin{align*}
        c_1\prod_{j=1}^D(1+\sigma_n^2\omega_j^2)^{-m_{\varepsilon}} \leq  \EE(e^{i\bomega^T\bvarepsilon_k})\leq c_2\prod_{j=1}^D(1+\sigma_n^2\omega_j^2)^{-m_{\varepsilon}}, \forall \bomega = (\omega_1,\ldots,\omega_D)\in\mathbb{R}^D.
    \end{align*}
    \item[(C3)] (Gaussian noise) The elements of $\bvarepsilon_k$ are normally distributed with variance $\sigma_n^2$.
\end{itemize}
Here the constants $c_1$ and $c_2$ do not depend on $\sigma_n$ and $m_\varepsilon$. We call $\sigma_n$ the smoothing scale in this work.
\end{assumption}

\begin{example}
It is easy to construct distributions satisfying (C1) or (C2). For example, the generalized Laplace distribution with parameter $s$ has a density function \citep{kozubowski2013multivariate,kotz2001laplace}
\begin{align}\label{eq_GLdensity}
    p_\varepsilon(\bx)  = \frac{2^{1-s}}{(2\pi)^{D/2}\Gamma(s)}(\sqrt{2}\|\bx\|_2)^{s+D/2}B_{s-D/2}\left(\sqrt{2}\|\bx\|_2\right),
\end{align}
where $\Gamma$ is the Gamma function, and $B_{s-D/2}$ is the modified Bessel function of the second kind. It can be shown that the generalized Laplace distribution 
% \yy{shall we give a precise definition here for completeness?}  
has the characteristic function 
\begin{align*}
    \EE_{\bX}(e^{i\bomega^T\bX}) = \left(1+\frac{1}{2}\bomega^T\bomega\right)^{-s}.
\end{align*}
Then $\bvarepsilon_k=\sigma_n\bX$ satisfies Assumption \ref{assum:augnoise} (C1). 
% As a special case, the multivariate Laplace distribution also satisfies Assumption \ref{assum:augnoise} (C1).
% the (isotropic) Mat\'ern family \citep{stein1999interpolation}, which, after a proper reparametrization, can be expressed as
% \begin{align}\label{kernelfunctionPsi}
% 	K_{m_{\varepsilon}}(\bx)=\frac{1}{\Gamma(m_{\varepsilon} - d/2)2^{m_{\varepsilon} - d/2 -1}}\|\bx\|_2^{m_{\varepsilon} - d/2} B_{m_{\varepsilon} - d/2}(\|\bx\|_2),
% \end{align}
% where $B_{m_{\varepsilon} - d/2}$ is the modified Bessel function of the second kind. The Fourier transform of the Mat\'ern kernel is 
% \begin{eqnarray}\label{maternfourier}
% \mathcal{F}(K_{m_{\varepsilon}})(\omega)=2^{d/2}\frac{\Gamma(m_{\varepsilon})}{\Gamma(m_{\varepsilon}-d/2)}(1+\|\bomega\|_2^2)^{-m_{\varepsilon}}.
% \end{eqnarray}
% Then $\bvarepsilon_k/\sigma_n$ with probability density function proportional to $K_{m_{\varepsilon}}(\bx)$ 

If each component of $\bvarepsilon_k/\sigma_n$ has a univariate generalized Laplace distribution and all components are independent, then Assumption \ref{assum:augnoise} (C2) is satisfied. 

% In particular, if each component follows the univariate Laplace distribution, Assumption \ref{assum:augnoise} (C2) holds.
\end{example}

%\yy{Add another example with Mat\'ern kernel?} \ww{I added an example after Assumption 6, where I put two kernel functions there.}

% \begin{assumption}\label{assum_stepsize}
% Let $\eta_1$ be the largest eigenvalue of $\Kb$, where $\Kb$ is as in \eqref{eq_lossX}. The learning rate $\beta$ satisfies $\beta\eta_1 +\alpha < 1$, where $\alpha=0$ if there is no weight decay, and $\alpha>0$ if there is weight decay.

% % If there is no weight decay, the learning rate $\beta$ satisfies $\beta\lambda_1 < 1$ and $1 -\beta\lambda_n < C < 1$. If there is weight decay, the learning rate $\beta$ and the weight decay parameter $\alpha$ satisfies $\beta\lambda_1 + \alpha < 1$ and $1-\alpha -\beta\lambda_n < C < 1$.
% % We assume that $\beta\lambda_1 + \alpha < 1$ and $1-\alpha -\beta\lambda_n =s < 1$, where $s$ is some constant independent of $n$, $\lambda_1$ is the largest eigenvalue of $\Kb$, and $\lambda_n$ is the smallest eigenvalue of $\Kb$. 
% \end{assumption}

% \ww{kl;asdfjkasdjfkasdfjl}

Assumption \ref{assum:sub-G} assumes that the observation error is sub-Gaussian, which is a standard assumption in nonparametric literature. See \cite{geer2000empirical} for example. Assumption \ref{assum:PsiDecay} assumes that the Fourier transform of the kernel function $K(\cdot-\cdot)$ has an algebraic decay. Under this assumption, Corollary 10.13 of \cite{wendland2004scattered} shows that the reproducing kernel Hilbert space $\cH_K(\RR^D)$ coincides with the Sobolev space $\cW^{m_0}(\RR^D)$, with equivalent norms. More details on this can be found in Section \ref{subsec:ProblemSettings}.
% We are only interested in the case of $m_0 > D/2$ because if $m_0 > D/2$, all functions in $\cW^{m_0}(\RR^D)$ are continuous according to the Sobolev embedding theorem \citep{adams2003sobolev}. 
Assumption \ref{assum:PsiDecay_tensor} states that the kernel function $K$ has a tensor structure, and the Fourier transform of each component $K_j$ has an algebraic decay. Assumptions \ref{assum:PsiDecay} and \ref{assum:PsiDecay_tensor} will be used in Sections \ref{sec:fullsobo} and \ref{sec:mixsobo}, respectively.
Assumption \ref{assum:augnoise} imposes conditions on the noise $\bvarepsilon_k$'s and considers three types of augmentations: polynomial noise, tensor polynomial noise, and Gaussian noise. The corresponding smoothing techniques are referred to as \textit{polynomial smoothing}, \textit{tensor polynomial smoothing}, and \textit{Gaussian smoothing}, respectively.

\subsection{Low Intrinsic Dimension Space}\label{sec:fullsobo}

We first consider $\Omega$ with finite intrinsic dimension. The intrinsic dimension provides a ``measure of the complexity'' for the region of interest $\Omega$.
% support of the $P_\Xb$, i.e., $\Omega$. 
The definition of the intrinsic dimension depends on the covering number; see Definition 2.1 of \cite{geer2000empirical} for example.

\begin{definition}[Covering number]
Consider a subset $\cA\subset \cG$ where $\cG$ is a normed space. For a given $\delta>0$, the covering number of $\cA$, denoted by $\cN_{\cG}(\delta,\cA)$, is defined by the smallest integer $M$ such that $\cA$ can be covered by $M$ balls with radius $\delta$ and centers $\bx_1,...,\bx_M\in \cG$. 
% Given a second normed space $F$ and a bounded linear operator $T:E\rightarrow F$, the covering numbers of $T$ are defined by $\cH(T,\epsilon) = \cH_F(TB_E,\epsilon)$. 
\end{definition}

\begin{assumption}[Low intrinsic dimension]\label{assum:intriD}
There exist positive constants $c_1$ and $d\le D$ such that for all $\delta\in (0,1)$, we have
\begin{align*}
    \cN_{\ell_\infty^D}(\delta,\Omega)\leq c_1\delta^{-d},
\end{align*}
where $\ell_\infty^D$ is the $\RR^D$ space equipped with $\ell_\infty$ norm.
\end{assumption}
For discussion and examples of regions that satisfy Assumption \ref{assum:intriD}, we refer to \cite{hamm2021adaptive}. In particular, if $\Omega\subset \RR^D$ is a bounded region with positive Lebesgue measure or a bounded $D'$-dimensional differentiable manifold, then Assumption \ref{assum:intriD} holds with $d=D$ and $d=D'$, respectively. 

% \ww{add new paragraph: Our notion of smoothness}

Besides the low intrinsic dimension, our theoretical results depend on the smoothness of the underlying function. Because we are considering function space on a finite intrinsic dimensional space, which may have Lebesgue measure zero, the usual definition of (fractional) Sobolev space via Fourier transform stated in Section \ref{subsec:ProblemSettings} cannot be directly applied in our case. Thus, we need to introduce our notion of the smoothness of functions on finite intrinsic dimension space.
% For an integrable function $f\in L_1(\mathbb{R}^D)$, its Fourier transform is defined as $$\mathcal{F}(f)(\bomega)=(2\pi)^{-D/2}\int_{\mathbb{R}^D} f(\bx) e^{-i\bx^{\mathrm{T}}\bomega}{\rm d}\bomega.$$ 
% Let $\Omega_1 = \{\bx\in \RR^D: \inf_{\bx'\in \Omega}\|\bx-\bx'\|_2\leq \delta\}$ be a $\delta$-neighborhood of $\Omega$. Clearly, $\Omega_1$ has a positive Lebesgue measure. We further assume that $\Omega_1$ has 
% Suppose there exists a region $\Omega_1$ with positive Lebesgue measure and a Lipschitz boundary \citep{leoni2017first}, such that $\Omega \subset \Omega_1$. Roughly speaking, we assume that the boundary of $\Omega_1$ is ``sufficiently regular'' and $\Omega$ can be contained by $\Omega_1$. Thus, the extension theorem \citep{devore1993besov} ensures that there exists an extension operator from $L_2(\Omega_1)$ to $L_2(\mathbb{R}^D)$ and the smoothness of each function is maintained. 
Specifically, we impose the following assumption on the underlying true function $f^*$.

\begin{assumption}\label{assum:fstar}
There exists a region $\Omega_1$ with positive Lebesgue measure and a Lipschitz boundary such that $\Omega \subset \Omega_1$.
The underlying true function $f^*$ is well-defined on $\Omega_1$ and $m_f = \arginf_{m>D/2} \{m:f^*\in \cW^{m}(\Omega_1)\}$ with $f^*\in \cW^{m_f}(\Omega_1)$, and $m_f>D/2$.
\end{assumption}

In Assumption \ref{assum:fstar}, we assume that the boundary of $\Omega_1$ is ``sufficiently regular'' (see \cite{leoni2017first} for the definition of Lipschitz boundary) and $\Omega$ can be contained by $\Omega_1$. Thus, the extension theorem \citep{devore1993besov} ensures that there exists an extension operator from $L_2(\Omega_1)$ to $L_2(\mathbb{R}^D)$ and the smoothness of each function is maintained. With Assumption \ref{assum:fstar}, we use $m_f$ to denote the smoothness of $f^*$. By some well-known extension theorems (see, for example, \cite{devore1993besov,evans2009partial,stein1970singular}), if $D=d$, then our notion of smoothness coincides with the smoothness of functions on the whole space $\RR^D$.

Now we are ready to present the main theorems in this subsection. Theorems \ref{thm:soboNN} and \ref{thm:soboGN} state the convergence rates when applying polynomial smoothing and Gaussian smoothing, respectively. 

% \ww{akl;sdfjdkl;asjf;kldasjfkl;sd}

\begin{theorem}[Polynomial smoothing]\label{thm:soboNN}
Suppose Assumptions \ref{assum:sub-G}, \ref{assum:PsiDecay}, \ref{assum:augnoise} (C1), \ref{assum:intriD} and \ref{assum:fstar} are satisfied. 
% Suppose there exists $\Omega_1$ with positive Lebesgue measure and a Lipschitz boundary such that $\Omega\subset \Omega_1$ and $f^*\in \cW^{m_f}(\Omega_1)$. 
Let $f_t(\bx)$ be as in \eqref{eq:predictor} and $\beta=n^{-1}C_1$ with $C_1\leq (2\sup_{\bx\in \RR^D}K_S(\bx))^{-1}$. Suppose the smoothing scale
% \yy{for polynomial smoothing, $\sigma_n$ appears in the scaling of characteristic functions. Is that the variance?} of the augmentation noise 
$\sigma_n\asymp n^{\nu}$ with $\nu \leq 0$. Suppose one of the following holds:
\begin{enumerate}
    \item There is no weight decay in the gradient descent, and the iteration number $t$ satisfies
    $
        t \asymp  n^{\frac{2(m_0+m_\varepsilon)}{2m_f+d}}\sigma_n^{2m_\varepsilon}
    $
    \item There is weight decay in the gradient descent with $\alpha \asymp n^{-1-\frac{2(m_0+m_\varepsilon)}{2m_f+d}}\sigma_n^{-2m_\varepsilon}$, and the iteration number satisfies $t\geq C_2(\frac{m_f}{2m_f+d}+1/2)\log n/(\log (1-\alpha))$.
\end{enumerate}
Then by setting $m_\varepsilon = 2d^{-1}(2D \max(m_0,m_f)+ m_0d)\log n - m_0$ and 
    \begin{align*}
        \nu =\left\{\begin{array}{ll}
             -\frac{2(2m_0+2m_\varepsilon)D-(2m_0+2m_\varepsilon-D)d}{(2m_f+d)(4m_\varepsilon D-(2m_0+2(1-(\log n)^{-1})m_\varepsilon -D)d)} < 0, & D>d, \\
            0, & D=d,
        \end{array}\right.                
    \end{align*}
    we have
    \begin{align*}
        \|f_t - f^*\|_{L_2(P_\Xb)}^2 = & O_{\PP}\left(n^{-\frac{2m_f}{2m_f + d}}(\log n)^{2m_f+1}\right).
    \end{align*}
for $N>N_0$, where $N$ is the number of augmentations, and $N_0$ depends on $n$ (specified in \eqref{eq_N0require}). 
% \begin{enumerate}
%     \item For any $a>0$, there exists an $m_\varepsilon$ such that when 
%     \begin{align*}
%         \nu =\left\{\begin{array}{ll}
%             -\frac{2(2m_0+2m_\varepsilon)D-(2m_0+2m_\varepsilon-D)d}{(2m_f+d)(4m_\varepsilon D-(2m_0+2(1-d^{-1}(2m_f+d)a)m_\varepsilon -D)d)}, & D>d, \\
%             0, & D=d,
%         \end{array}\right.        
%     \end{align*}
%     we have
%     \begin{align*}
%         \|f_t - f^*\|_{L_2(P_\Xb)}^2 = & O_{\PP}\left(n^{-\frac{2m_f}{2m_f + d}+a}\right).
%     \end{align*}
%     \item Set $m_\varepsilon = 2d^{-1}(2D \max(m_0,m_f)+ m_0d)\log n - m_0$. Then by choosing 
%     \begin{align*}
%         \nu =\left\{\begin{array}{ll}
%              -\frac{2(2m_0+2m_\varepsilon)D-(2m_0+2m_\varepsilon-D)d}{(2m_f+d)(4m_\varepsilon D-(2m_0+2(1-(\log n)^{-1})m_\varepsilon -D)d)} < 0, & D>d, \\
%             0, & D=d,
%         \end{array}\right.                
%     \end{align*}
%     we have
%     \begin{align*}
%         \|f_t - f^*\|_{L_2(P_\Xb)}^2 = & O_{\PP}\left(n^{-\frac{2m_f}{2m_f + d}}(\log n)^{2m_f+1}\right).
%     \end{align*}
% \end{enumerate}
\end{theorem}

\begin{theorem}[Gaussian smoothing]\label{thm:soboGN}
Suppose Assumptions \ref{assum:sub-G}, \ref{assum:PsiDecay}, \ref{assum:augnoise} (C3), \ref{assum:intriD}, and \ref{assum:fstar} are satisfied. 
% Suppose there exists $\Omega_1$ with positive Lebesgue measure and a Lipschitz boundary such that $\Omega\subset \Omega_1$ and $f^*\in \cW^{m_f}(\Omega_1)$. 
Let $f_t(\bx)$ be as in \eqref{eq:predictor}, $\beta=n^{-1}C_1$ with $C_1\leq (2\sup_{\bx\in \RR^D}K_S(\bx))^{-1}$, and $\sigma_n\asymp n^{-\frac{1}{2m_f+d}}$.
Suppose one of the following holds:
\begin{enumerate}
    \item There is no weight decay in the gradient descent, and the iteration number $t$ satisfies $t \asymp n^{\frac{2m_0+2m_f}{2m_f+d}}$
    \item There is weight decay in the gradient descent with $\alpha \asymp n^{-1-\frac{2(m_0+m_\varepsilon)}{2m_f+d}}$, and the iteration number satisfies $t\geq C_2(\frac{m_f}{2m_f+d}+1/2)\log n/(\log (1-\alpha))$.
\end{enumerate}
Then we have
\begin{align}\label{eq:efullsobonowd1}
        \|f^* - \hat f_t\|_{L_2(P_\Xb)}^2 = O_{\PP}(n^{-\frac{2m_f}{2m_f+d}}(\log n)^{D+1}),
    \end{align}  
when $N>N_0$, where $N$ is the number of augmentations, and $N_0$ depends on $n$ (specified in \eqref{eq_N0requireG}).
% Then the following statements are true with $N>N_0$, where $N$ is the number of augmentations, and $N_0$ depends on $n$ (specified in \eqref{eq_N0requireG}).
% \begin{enumerate}
%     \item (Without weight decay). If the iteration number $ t \asymp n^{\frac{2m_0+2m_f}{2m_f+d}}$, then 
%     \begin{align}\label{eq:efullsobonowd1}
%         \|f^* - \hat f_t\|_{L_2(P_\Xb)}^2 = O_{\PP}(n^{-\frac{2m_f}{2m_f+d}}(\log n)^{D+1}).
%     \end{align}
%     \item (With weight decay). Let $\alpha \asymp n^{-1-\frac{2(m_0+m_\varepsilon)}{2m_f+d}}$, and the iteration number satisfies $t\geq C_2(\frac{m_f}{2m_f+d}+1/2)\log n/(\log (1-\alpha))$. Then we have 
%     \begin{align}\label{eq:efullsobonowd2}
%         \|f^* - \hat f_t\|_{L_2(P_\Xb)}^2 = O_{\PP}(n^{-\frac{2m_f}{2m_f+d}}(\log n)^{D+1}).
%     \end{align}
% \end{enumerate}
% \yy{Perhaps just one equation for the common rates? Like previous theorem} 
\end{theorem}

\begin{remark}
    We require $\beta=n^{-1}C_1$ with $C_1\leq (2\sup_{\bx\in \RR^D}K_S(\bx))^{-1}$ in both Theorems \ref{thm:soboNN} and \ref{thm:soboGN} is because by Gershgorin's theorem \citep{varga}, we have for sufficiently large $n$,     
    \begin{align*}
    \beta\eta_1(\Kb) +\alpha \leq \beta n\max_{j,k}|K_S(\bx_j,\bx_k)|+\alpha < 1,
    \end{align*}
    which ensures that the gradient descent algorithm can converge.
\end{remark}

If the region $\Omega$ has a positive Lebesgue measure, then it has been shown that the optimal convergence rate is $n^{-m_f/(2m_f+D)}$ \citep{stone1982optimal}. By random smoothing, the gradient descent with early stopping can achieve the optimal convergence rate in this case, up to a logarithm term. Furthermore, it can adapt to the low intrinsic dimension case, where $\Omega$ can have Lebesgue measure zero. In \cite{hamm2021adaptive}, it is strongly hypothesized that the convergence rate $n^{-m_f/(2m_f+d)}$ is optimal. Although our definition of the smoothness is different, we have the same hypothesis and leave its exploration as a future work.

It is worth noting that our approach differs from that in \cite{hamm2021adaptive}, and therefore, we can investigate the effects of polynomial smoothing, which may have its own interest.  Such non-smooth noise can shed light on non-smooth augmentations commonly used in practice.  Furthermore, we obtain an identical result as in \cite{hamm2021adaptive} if we use Gaussian smoothing. Comparing the convergence rates in Theorems \ref{thm:soboNN} and \ref{thm:soboGN}, we find that the convergence rate by polynomial smoothing is slightly worse than that of Gaussian smoothing, since $m_f>D/2$ (Assumption \ref{assum:fstar}). In comparison, \cite{eberts2013optimal} achieved convergence rate of the similar form $n^{-2 m_f/(2 m_f+d) +\xi}$ by applying kernel ridge regression with Gaussian kernel functions, where $\xi$ can be any value strictly larger than zero. Clearly, this rate is slower than those in \cite{hamm2021adaptive} and ours.
% As $\log (n) \lesssim n^\xi$ holds, our rate is better than theirs, no matter in no-smooth noise or Gaussian noise cases. 
Under additional assumptions such as a compact Riemannian manifold input space and the underlying function having Lipschitz continuity $m_f\in(0,1]$, \cite{ye2008learning} derived convergence rates of the form $\left(\log ^2 (n)/n \right)^{m_f/(8 m_f+4 d)}$. 
% In comparison, our results in Theorems \ref{thm:soboNN} and \ref{thm:soboGN} 
Instead of kernel ridge regression, \cite{yang2016bayesian} focused on Bayesian regression with Gaussian process and proved the convergence rate $n^{-2m_f /(2 m_f+d)}(\log n)^{d+1}$. However, their theorem is limited by a compact low dimensional differentiable manifold input space, and the condition $m_f \leq 2$. As a comparison, we do not require such restrictive assumptions.

From a different perspective of early stopping, we consider both cases with and without weight decay, while existing studies only consider the case without weight decay. With weight decay, one can achieve the same convergence rate but with a much smaller iteration number. Specifically, the iteration number should be polynomial in $n$ without weight decay, which can be reduced to polynomial in $\log n$ if one applies weight decay. This also justifies the use of weight decay in practice. Besides, the random smoothing kernel enables us to establish connections with data augmentation and we further explain the effectiveness of using augmentation, which may lead to a new interpretation of using augmentations in deep learning.

% \ww{as;lkfjaskljflaskdjfkls}

% \ww{added the following paragraph}

% Moreover, our approach considers both cases with and without weight decay, whereas previous studies only consider the case without weight decay. With weight decay, we can achieve the same convergence rate as without weight decay, but with a much smaller iteration number, which can be reduced to polynomial$(\log n)$ instead of polynomial$(n)$. This has important practical implications and also enables us to establish connections between early stopping and data augmentation. We provide an explanation of the effectiveness of using augmentations, which may lead to a new interpretation of their use in deep learning.

Our approach to studying early stopping is distinct from previous studies in the literature (see, e.g., \cite{dieuleveut2016nonparametric, yao2007early,pillaud2018statistical,raskutti2014early}), which typically use integral operator techniques and impose assumptions on the eigenvalues of the kernel function (which always exists by Mercer's theorem). However, such assumptions cannot be easily applied to the low intrinsic dimension case, as it is unclear how eigenvalues behave in this regime. Additionally, previous studies often impose a ``source condition'' that requires the kernel function to have finite smoothness, which is not satisfied when using Gaussian smoothing to construct the random smoothing kernel. Therefore, even for the special case where the intrinsic dimension is equal to the ambient dimension, Theorems \ref{thm:soboNN} and \ref{thm:soboGN} improve upon previous results in the early stopping literature.

\begin{remark}
In general, the Bessel potential space used in our work is different from the Besov space used in \cite{hamm2021adaptive}. Specifically, the Bessel potential space is obtained via complex interpolation, while the Besov space is constructed by real interpolation. For a more thorough explanation, readers may refer to \cite{edmunds2008function}. We chose to use the Bessel potential space because of its natural connection to the Fourier transform and the characteristic function of a random variable, which allowed us to study the impact of the augmentations considered in our work.
\end{remark}

\begin{remark}
There are some other notions of smoothness in the literature. For example, \cite{hamm2021adaptive} define the smoothness induced by the Besov spaces, and \cite{yang2016bayesian} assume $f^*$ has $k$-th continuous derivatives. Another alternative definition of the Sobolev space on $\Omega$ is via Sobolev–Slobodeckij spaces. For simplicity, let $\Omega \subset \RR^{D-1}$. For a function $f$, $\theta\in (0,1)$, and $s>0$, define the Slobodeckij seminorm
\begin{align*}
    |f|_{\theta,\Omega} = \left(\int_{\Omega\times\Omega}\frac{|f(\bx)-f(\bx')|^2}{\|\bx-\bx'\|_2^{2\theta+D-1}}{\rm d}\bx{\rm d}\bx'\right)^{1/2}.
\end{align*}
Then the Sobolev–Slobodeckij space on $\Omega$, denoted by $W^s(\Omega)$, is defined by 
\begin{align*}
    W^s(\Omega) = \left\{f: f\in W^{\lfloor s \rfloor}(\Omega): \sup_{\alpha = \lfloor s \rfloor} |D^\alpha f|_{\theta,\Omega} <\infty \right\},
\end{align*}
with norm
\begin{align*}
    \|f\|_{W^s(\Omega)} = \|f\|_{W^{\lfloor s \rfloor}(\Omega)} +\sup_{\alpha = \lfloor s \rfloor} |D^\alpha f|_{\theta,\Omega},
\end{align*}
and $D^\alpha f:=\frac{\partial^{|\alpha|}}{\partial x_1^{\alpha_1}\ldots x_d^{\alpha_d}} f$ denotes the $\alpha$-th (weak) derivative of a function $f$ with $|\alpha|=\alpha_1+\ldots+\alpha_d$ for a multi-index $\alpha=(\alpha_1,\ldots,\alpha_d)\in \mathbb{N}_0^d$. By the trace extension theorem \citep{Triebel}, there exists an extension operator such that the extended function $f_E\in W^{s+1/2}(\RR^D)$ and $f_E|_{\RR^{D-1}}=f$, which implies $m_f=s+1/2$ if $\Omega$ has a positive Lebesgue measure in $\RR^{D-1}$.
\end{remark}

% In order to achieve 

% In the full Sobolev case, suppose the function class of interest is $\cH(\Omega) = \cW^{m_f}(\Omega)$ with $m_f>D/2$. It has been shown that the optimal convergence rate is $n^{-m_f/(2m_f+D)}$ \citep{stone1982optimal}. With sufficient number of augmentations, we show that even if the kernel function of the RKHS has a low smoothness, this optimal convergence rate can still be achieved in the full Sobolev space, as stated in the following two theorems.

% Our theorems blablabla...

% By adding non-smooth noise, the corresponding early stopping algorithm can be adapted to the low-intrinsic noise case.

% \begin{remark}
% Improve Steinwart 13 EJS from $n^{-\frac{2m_f}{2m_f+D}+\xi}$ to $n^{-\frac{2m_f}{2m_f+D}}(\log n)^{D+1}$. Improve Steinwart 21 Annals from bounded $y$ to sub-Gaussian noise. The non-smooth noise improve the convergence rate. \yy{perhaps say more about what's the key difference? a selling point?}
% \end{remark}

% \begin{remark}
% Consider early stopping with weight decay and no weight decay.
% \end{remark}

% \begin{remark}
% Consider random smoothing kernel, establish relationship with data augmentation.
% \end{remark}

\subsection{Tensor Reproducing Kernel Hilbert Space}\label{sec:mixsobo}

In this section, we consider a low-dimensional structure for the function class, specifically a \textit{tensor reproducing kernel Hilbert space}. Let $K = \prod_{j=1}^DK_j$ be kernel functions that satisfy Assumption \ref{assum:PsiDecay_tensor}, while $\Omega$ can have a low intrinsic dimensional structure, as discussed in Section \ref{sec:fullsobo}, or have a positive Lebesgue measure in $\RR^D$. 

Our theoretical results in this section are based on mixed smooth Sobolev spaces, denoted by $\cMW^m(\RR^D)$, where $m>1/2$. For a function $f$ defined on $\RR^D$, the mixed smooth Sobolev norm is defined as
\begin{eqnarray}\label{eq_mixSobolev}
\|f\|_{\cMW^m(\mathbb{R}^D)}=\left(\int_{\mathbb{R}^D} |\mathcal{F} (f)(\bomega)|^2 \prod_{j=1}^D(1+|\omega_j|^2)^{m} {\rm d} \bomega\right)^{1/2},
\end{eqnarray}
and the mixed smooth Sobolev spaces on $\Omega$ can be defined via restriction similar to the Sobolev spaces. In fact, the mixed smooth Sobolev space is a tensor product of one-dimensional Sobolev spaces, and it can be shown that $\cMW^{m_0}(\mathbb{R}^D)$ is equivalent to the tensor reproducing kernel Hilbert space generated by kernel function $K=\prod_{j=1}^DK_j$ satisfying Assumption \ref{assum:PsiDecay_tensor}. Because of such a tensor structure, it is often considered as a reasonable model reducing the complexity in high-dimensional spaces \citep{kuhn2015approximation,dung2021deep}. For instance, the mixed smooth Sobolev spaces are utilized in high-dimensional approximation and numerical methods of PDE \citep{bungartz1999note}, data mining \citep{garcke2001data}, and deep neural networks \citep{dung2021deep}.

If the underlying function belongs to some mixed smooth Sobolev space, then it can be shown that by applying appropriate augmentations, we can achieve a fast convergence rate, which nearly coincides with the minimax rate in the one-dimensional case, up to a logarithmic term. Similar to Assumption \ref{assum:fstar}, we assume that $f^*$ can be extended to some ``regular space'' with positive Lebesgue measure, as follows.
% Specifically, we assume the following.

\begin{assumption}\label{assum:fstarten}
There exists a region $\Omega_1$ with positive Lebesgue measure and a Lipschitz boundary such that $\Omega \subset \Omega_1$, and the underlying true function $f^*$ is well-defined on $\Omega_1$ and  $f^*\in \cMW^{m_f}(\Omega_1)$. 
\end{assumption}

The following theorem states the convergence rate when applying tensor polynomial smoothing in the tensor RKHS case. 
% The main theorem in this section is as follows.

\begin{theorem}[Tensor polynomial smoothing]\label{thm:soboNN_tensor}
Suppose Assumptions \ref{assum:sub-G}, \ref{assum:PsiDecay_tensor}, \ref{assum:augnoise} (C2), \ref{assum:intriD}, and \ref{assum:fstarten} are satisfied. 
% Suppose there exists $\Omega_1$ with positive Lebesgue measure and a Lipschitz boundary such that $\Omega\subset \Omega_1$ and $f^*\in \cW^{m_f}(\Omega_1)$.  
Let $f_t(\bx)$ be as in \eqref{eq:predictor} and $\beta=n^{-1}C_1$ with $C_1\leq (2\sup_{\bx\in \RR^D}K_S(\bx))^{-1}$. Let $m_\varepsilon +m_0\geq m_f$, and the smoothing scale  $\sigma_n\asymp 1$.
    % where $C$ is a constant depending on $m_0,m_f,d,D$. 

% \begin{align*}
%     \lambda_n \asymp  n^{-\frac{2(m_0+m_\varepsilon)}{2m_f+d}}\sigma_n^{-2m_\varepsilon},
%     \sigma_n \asymp  n^{-\frac{(m_0+m_\varepsilon)(D-d)}{(2m_f+d)(m_\varepsilon D-m_0 d -(1-\delta_0) m_\varepsilon d)}},
%     n^{-1}(\beta t)^{-1} \asymp  \lambda_n,
%     \beta \asymp n^{-1}.
% \end{align*}
% Therefore, if 
% \begin{align}\label{eq:mainthmnn1111}
%     & \lambda_n \leq C(\lambda_n\sigma_n^{2m_\varepsilon})^{\frac{m_f}{m_0+m_\varepsilon}}\nonumber\\
%     \Leftrightarrow & n^{-\frac{2(m_0+m_\varepsilon)}{2m_f+d}}n^{\frac{2m_\varepsilon(m_0+m_\varepsilon)(D-d)}{(2m_f+d)(m_\varepsilon D-m_0 d -(1-\delta_0) m_\varepsilon d)}} \leq Cn^{-\frac{2m_f}{2m_f+d}}\nonumber\\
%     % \Leftrightarrow & -\frac{2(m_0+m_\varepsilon)}{2m_f+d}-\frac{2m_\varepsilon(m_0+m_\varepsilon)(D-d)}{(2m_f+d)(m_\varepsilon D-m_0 d -(1-\delta_0) m_\varepsilon d)}\leq -\frac{2(m_0+m_\varepsilon)}{2m_f+d}\nonumber\\
%     \Leftarrow & (m_0+m_\varepsilon)(\delta_0 m_\varepsilon d - m_0 d) > m_f(m_\varepsilon D-m_0 d -(1-\delta_0) m_\varepsilon d)\nonumber\\
%     \Leftarrow & m_\varepsilon > \frac{m_fD+m_0d}{\delta_0 d},
% \end{align}
% $m_\varepsilon \geq C>m_0$, 
Then the following statements are true with $N>N_0$, where $N$ is the number of augmentations, and $N_0$ depends on $n$ (specified in \eqref{eq_N0requirete}). Suppose one of the following holds:
\begin{enumerate}
    \item There is no weight decay in the gradient descent, and the iteration number $t$ satisfies
    $
        t \asymp  n^{\frac{2(m_0+m_\varepsilon)}{2m_f+1}}(\log n)^{\frac{2(D-1)(m_0+m_\varepsilon)+1}{2m_f+1}}
    $
    \item There is weight decay in the gradient descent with $\alpha \asymp n^{-1-\frac{2(m_0+m_\varepsilon)}{2m_f+d}}(\log n)^{\frac{2(D-1)(m_0+m_\varepsilon)+1}{2m_f+1}}$, and the iteration number satisfies $t\geq C_2(\frac{m_f}{2m_f+1}+1/2)\log n/(\log (1-\alpha))$.
\end{enumerate}
Then we have 
\begin{align}\label{eq_thmtensor}
        \|f_t - f^*\|_{L_2(P_\Xb)}^2 = & O_{\PP}\left(n^{-\frac{2m_f}{2m_f + 1}}(\log n)^{\frac{2m_f}{2m_f+1}\left(D-1+\frac{1}{2(m_0+m_\varepsilon)}\right)}\right).
    \end{align}
% \begin{enumerate}
%     \item (Without weight decay).
%     If the iteration number $t$ satisfies
%     $
%         t \asymp n^{\frac{2(m_0+m_\varepsilon)}{2m_f+d}-1}\sigma_n^{2m_\varepsilon},
%     $
%     we have
%     \begin{align*}
%         \|f_t - f^*\|_{L_2(P_\Xb)}^2 = & O_{\PP}\left(n^{-\frac{2m_f}{2m_f + d}}\right).
%     \end{align*}
%     \item (With weight decay). Let $\alpha \asymp n^{-\frac{2(m_0+m_\varepsilon)}{2m_f+d}}\sigma_n^{-2m_\varepsilon}$. If the iteration number satisfies $t\geq C_2(\frac{m_f}{2m_f+d}+1/2)\log n/(\log (1-\alpha))$, we have 
%     \begin{align*}
%         \|f_t - f^*\|_{L_2(P_\Xb)}^2 = & O_{\PP}\left(n^{-\frac{2m_f}{2m_f + d}}\right).
%     \end{align*}
% \end{enumerate}
\end{theorem}

% Comparing Theorem \ref{thm:soboNN_tensor} with Theorem \ref{thm:soboNN}, we can see that if Assumption \ref{assum:fstarten} holds, the requirements of the tensor polynomial smoothing are much milder than those in Theorem \ref{thm:soboNN}. First, in Theorem \ref{thm:soboNN_tensor}, the smoothness of the tensor polynomial smoothing $m_\varepsilon$ can be a constant as long as $m_\varepsilon +m_0\geq m_f$, while in Theorem \ref{thm:soboNN}, $m_\varepsilon$ needs to be comparable with $\log n$. Second, the smoothing scale $\sigma_n$ can be a constant in Theorem \ref{thm:soboNN_tensor}, while the smoothing scale in Theorem \ref{thm:soboNN} needs to be carefully chosen. These differences indicates that the tensor RKHS has a much simpler structure than the RKHS even on a low intrinsic dimension space. In fact, the convergence rate in Theorem \ref{thm:soboNN_tensor} does not depend on the low intrinsic dimension of $\Omega$, and is almost dimension-free. Furthermore, because the power of the logarithmic term in \eqref{eq_thmtensor} decreases as $m_\varepsilon$ increases, we can see that if $m_\varepsilon$ is large, the convergence rate in Theorem \ref{thm:soboNN_tensor} is faster, which encourages using a smoother tensor polynomial smoothing. This statement is align with the results in Theorem \ref{thm:soboNN} and Theorem \ref{thm:soboGN}, because applying Gaussian smoothing can yield a faster convergence rate than polynomial smoothing. As far as we know, few results are presented in existing literature for tensor RKHSs with early stopping, and our results can shed light on it. 

Based on Theorem \ref{thm:soboNN_tensor}, tensor polynomial smoothing leads to a convergence rate of tensor RKHS, which is $O_{\PP}(n^{-\frac{2m_f}{2m_f + 1}}(\log n)^{\frac{2m_f}{2m_f+1}(D-1+\frac{1}{2(m_0+m_\varepsilon)})})$. This convergence rate is almost the same as the optimal convergence rate in the one-dimensional case $O_{\PP}(n^{-\frac{2m_f}{2m_f + 1}})$, differing only by a logarithmic term.

Moreover, compared to Theorem \ref{thm:soboNN}, Theorem \ref{thm:soboNN_tensor} has less stringent requirements for tensor polynomial smoothing when Assumption \ref{assum:fstarten} holds. Specifically, Theorem \ref{thm:soboNN_tensor} allows for $m_\varepsilon$ to be a constant as long as $m_\varepsilon +m_0\geq m_f$, whereas Theorem \ref{thm:soboNN} requires $m_\varepsilon$ to be comparable to $\log n$. Additionally, while the smoothing scale $\sigma_n$ in Theorem \ref{thm:soboNN} demands careful selection, Theorem \ref{thm:soboNN_tensor} permits a constant smoothing scale $\sigma_n$. These differences suggest that the tensor RKHS has a simpler structure than the RKHS even in a low intrinsic dimension space. The convergence rate in Theorem \ref{thm:soboNN_tensor} does not depend on the low intrinsic dimension of $\Omega$, and is almost dimension-free.  Moreover, because the power of the logarithmic term in \eqref{eq_thmtensor} decreases as $m_\varepsilon$ increases, the convergence rate in Theorem \ref{thm:soboNN_tensor} decreases as $m_\varepsilon$ increases, encouraging the use of a smoother tensor polynomial smoothing for faster convergence. This aligns with the results in Theorem \ref{thm:soboNN} and Theorem \ref{thm:soboGN}, as Gaussian smoothing may yield faster convergence rates than polynomial smoothing. Few studies have explored tensor RKHSs with early stopping, and our findings can provide valuable insights into this area.

% , which we believe is unavoidable \yy{?}, because the tensor RKHS, albeit having a low dimensional structure, still lies in a $D$-dimensional Euclidean space. 

\begin{remark}
    For any $\cW^{m_f}(\RR^D)$ with $m_f>D/2$, there exist $m^{*}>1/2$ such that $\cW^{m_f}(\RR^D)\hookrightarrow \cMW^{m^*}(\RR^D)$ and $ \cMW^{m^*}(\RR^D)\hookrightarrow\mathcal{C}(\RR^D)$. Thus, the capacity of $\cMW^{m^*}(\RR^D)$ is high-enough for any approximation problem which can be solved by assuming that the underlying true function lies in some Sobolev space. % The assumption that the underlying function is in a $\cMW^{m^*}(\RR^D)$ is not a strong assumption.
\end{remark}

\section{Numerical Studies}\label{sec:num}
\begin{figure}[!h]
    \centering
    \includegraphics[width=16cm]{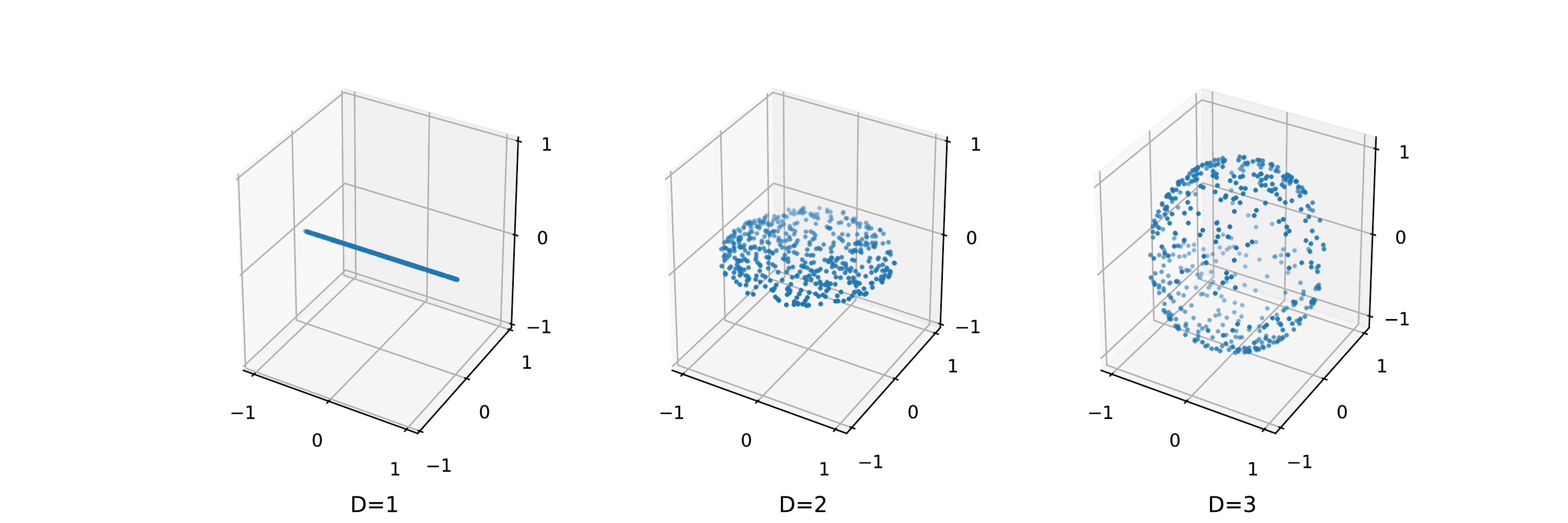}
    \caption{Simulated data spaces in the forms of: line ($D=1$), circle ($D=2$) and sphere ($D=3$).}
    \label{fig:ori_data}
\end{figure}

In this section, we enhance our theoretical findings by experimentally validating the effectiveness of the random smoothing kernel with data augmentation and early stopping on synthetic datasets. We focus on three data spaces with dimensions $D=1,2,3$, as illustrated in Figure \ref{fig:ori_data}, where $\bx_j$ samples are uniformly drawn.

%In this section, we complement our theoretical results by experimentally verifying the performance of the random smoothing kernel with data augmentation and early stopping on synthetic datasets. Specifically,
% \subsection{Synthetic Data and Training Settings}\label{sec:data}
% synthetic: Full Sobo, intrinsic, tensor
%we consider three data spaces with dimension $D=1,2,3$, respectively, demonstrated in Figure \ref{fig:ori_data}, where $\bx_j$'s are sampled uniformly. 

In our experiments, the underlying function $f^*$ is obtained by drawing random sample paths from the Gaussian process with the Matérn covariance function. This covariance function is widely used in Gaussian process modeling. We adopt the Matérn covariance function with the following form:
\begin{align}\label{eq:matern}
    K_\nu(\bx)=\sigma^2 \frac{2^{1-\nu}}{\Gamma(\nu)}\left(\sqrt{2 \nu} \frac{\|\bx\|_2}{\rho}\right)^\nu B_\nu\left(\sqrt{2 \nu} \frac{\|\bx\|_2}{\rho}\right),
\end{align}
where $\sigma, \phi,\nu>0$, $\Gamma$ is the Gamma function, and $B_\nu$ is the modified Bessel function of the second kind. 
In order to make $f^*$ smoother, we set the smoothness parameter $\nu=5.0$ for Matérn kernel \eqref{eq:matern}. The error $\epsilon_j$'s are i.i.d. Gaussian with mean zero and variance 0.01.

We utilize two-hidden-layer neural networks with ReLU activation \citep{nair2010relu} as our predictor. Each hidden layer of the neural network comprises 100 nodes, and all weights are initialized using Kaiming Initialization \citep{he2015delving}. For random smoothing, we experiment with both non-smooth Laplace noise and smooth Gaussian noise. To be precise, each element of $\bvarepsilon_k$ is randomly sampled from either $\mathcal{N}(0,\sigma^2)$ or $Laplace(0,b)$.
% The underlying function is obtained by random sample paths drawing from Gaussian processes with Matérn correlation functions. 
% We adopt a two-hidden-layer neural network with ReLU activation \citep{nair2010relu} as our predictor. 
For more experiment details and additional results, we refer to Appendix \ref{app:experiments}.

% \begin{figure}[!h]
%     \centering
%     \includegraphics[width=16cm]{img/data/underlying truth.pdf}
%     \caption{Underlying truth generated by Gaussian process. $X_1$ in the x-axis denotes the first dimension of generated data.}
%     \label{fig:underlying_truth}
% \end{figure}

% \subsection{With Weight Decay}

\begin{figure}[!t]
    \centering
    \includegraphics[width=16cm]{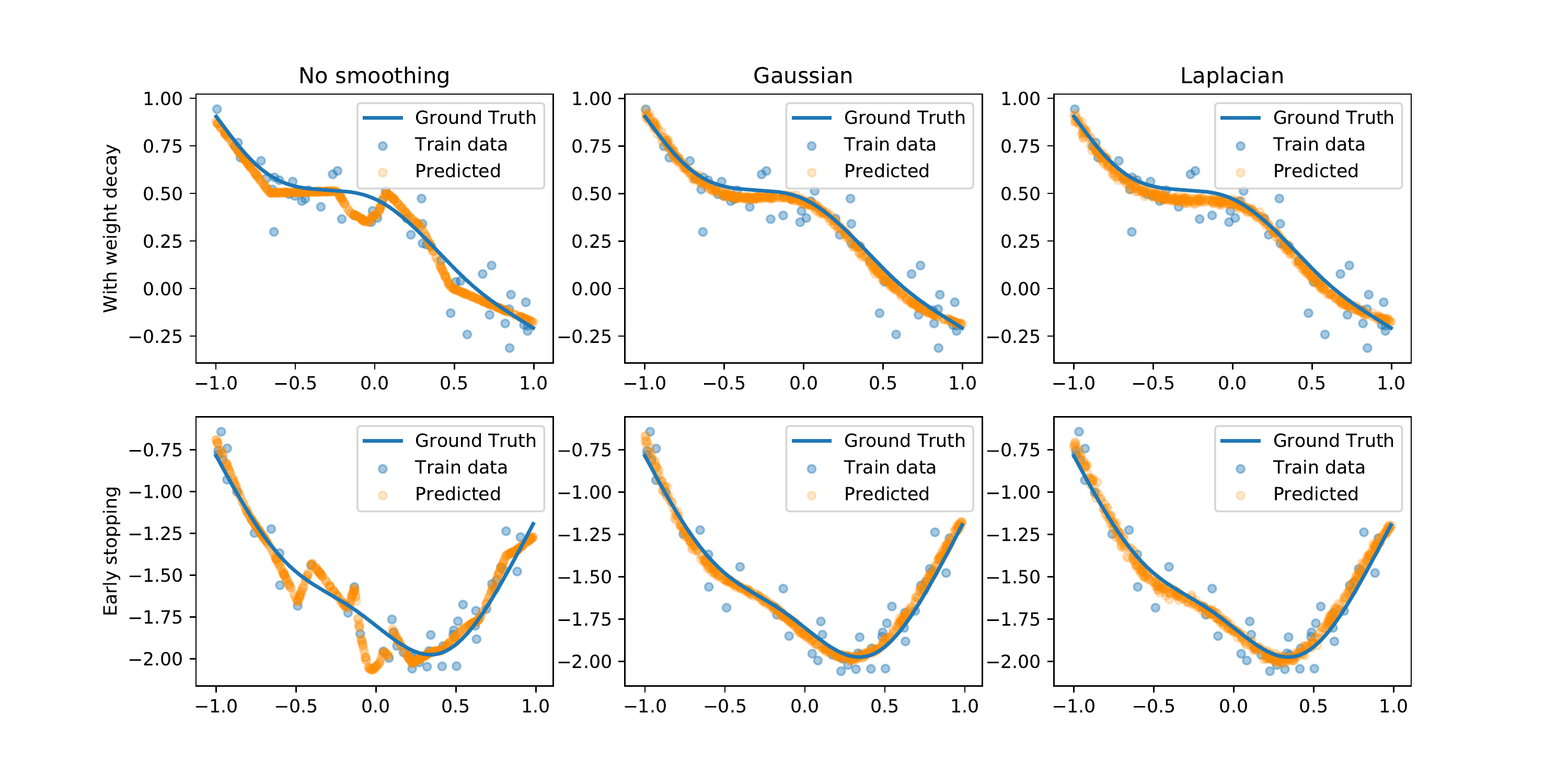}
    \caption{Visualization of the underlying truth (blue curve), training data (blue dots), and neural network predictions (orange dots) when training size is 50, where the first and second rows represent cases with weight decay and early stopping, respectively. It is obvious to see that the optimization without random smoothing will be more vulnerable to noise.
    % We refer to Appendix \ref{app:experiments} for size 100 and 200.
    } 
    \label{fig:fitted_1d}
\end{figure}

Figure \ref{fig:fitted_1d} presents a visualization of the underlying truth (blue curve), training data (blue dots), and neural network predictions (orange dots) when the training size is 50. The underlying truth is smooth since we use a smooth kernel. However, the neural network predictions without random smoothing are not smooth due to the low smoothness of the ReLU activation function and tend to overfit the noise. Upon applying random smoothing, the neural network predictions become smoother and approach the underlying truth.

% 1) Random smoothing is better than no smoothing
% 2) U shape curve and an optimal scale.
% 3) 1d fitted curve, smooth
% 4) Early stopping and weight decay are similar?

\begin{figure}[!h]
    \centering
    \includegraphics[width=16cm]{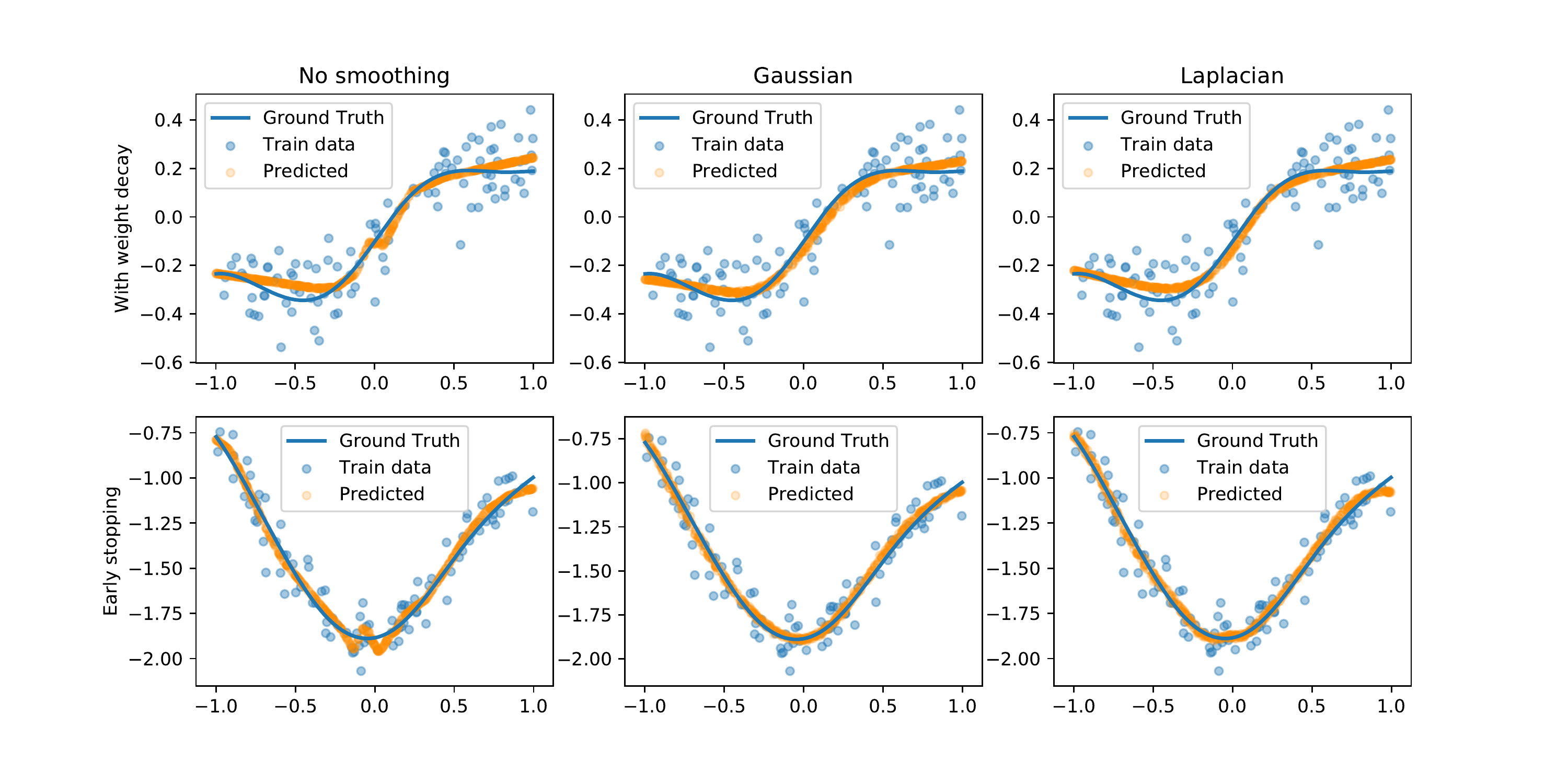}
    \caption{Underlying truth (blue curve), training data (blue dots), and neural network predictions (orange dots) when training size is 100.}
    \label{fig:fitted_1d_size100}
\end{figure}

\begin{figure}[!h]
    \centering
    \includegraphics[width=16cm]{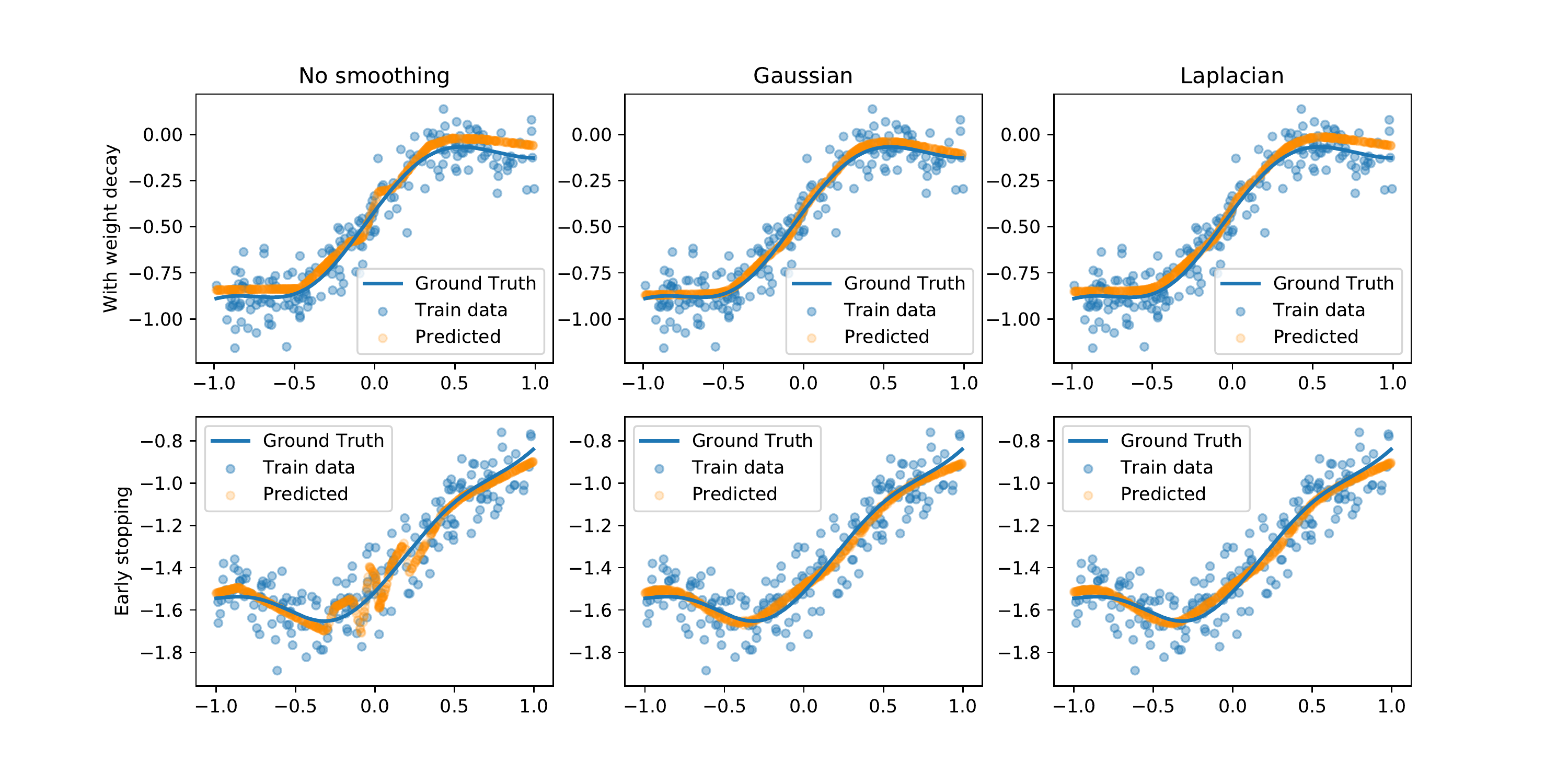}
    \caption{Underlying truth (blue curve), training data (blue dots), and neural network predictions (orange dots) when training size is 200.}
    \label{fig:fitted_1d_size200}
\end{figure}

Figure \ref{fig:fitted_1d_size100} and Figure \ref{fig:fitted_1d_size200} further show the underlying truth (blue curve), training data (blue dots), and neural network predictions (orange dots) when the training size is 100 and 200, respectively. Although increasing the training size improves smoothness in cases like size 200 with weight decay, the fitted curve still experiences a perturbation from overfitted noise compared to examples where random smoothing is applied.

% Table generated by Excel2LaTeX from sheet 'mse'

\begin{figure}[!h]
    \centering
    \begin{tabular}{c|c}
        With Weight Decay & Early Stopping \\ 
        \includegraphics[width=7cm]{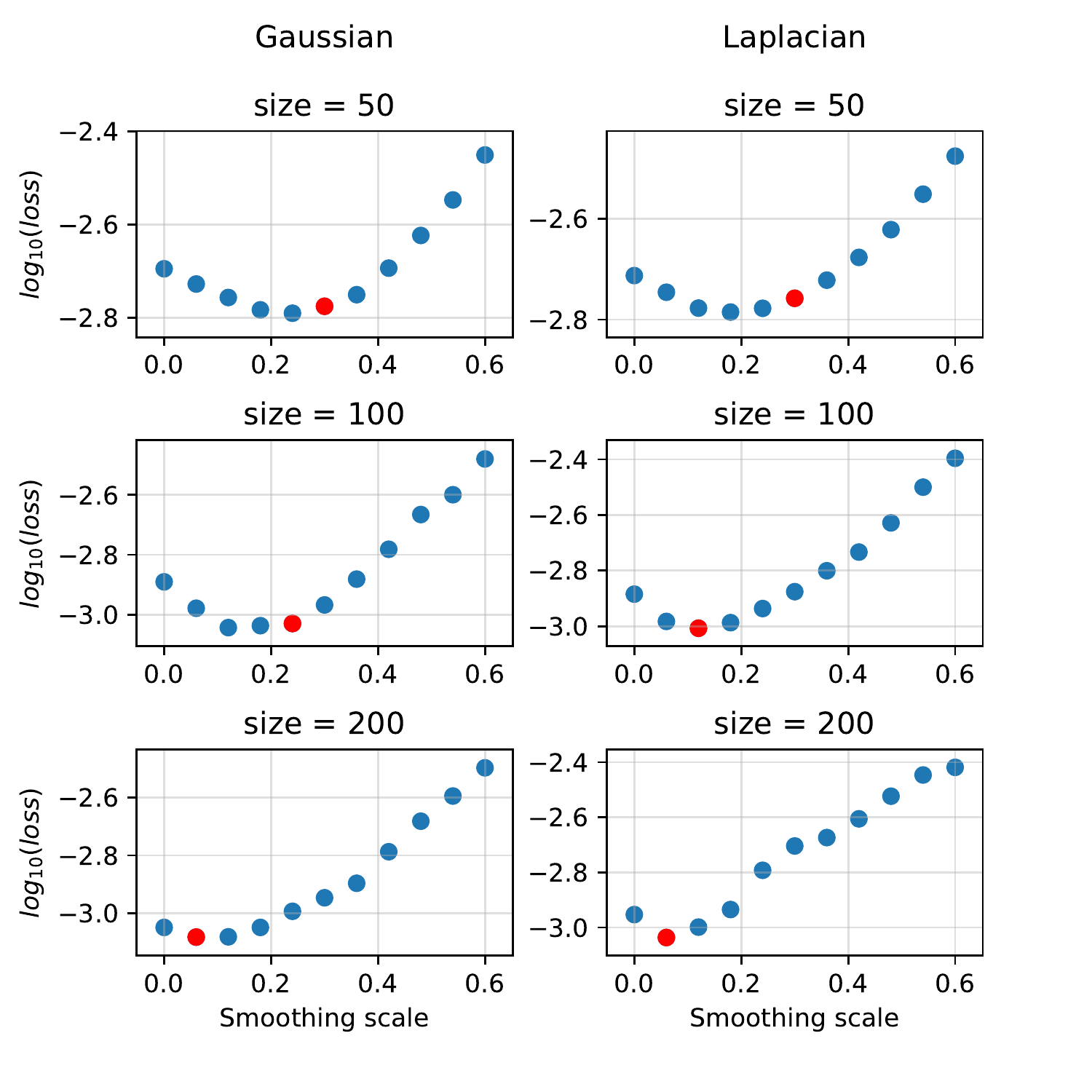} & \includegraphics[width=7cm]{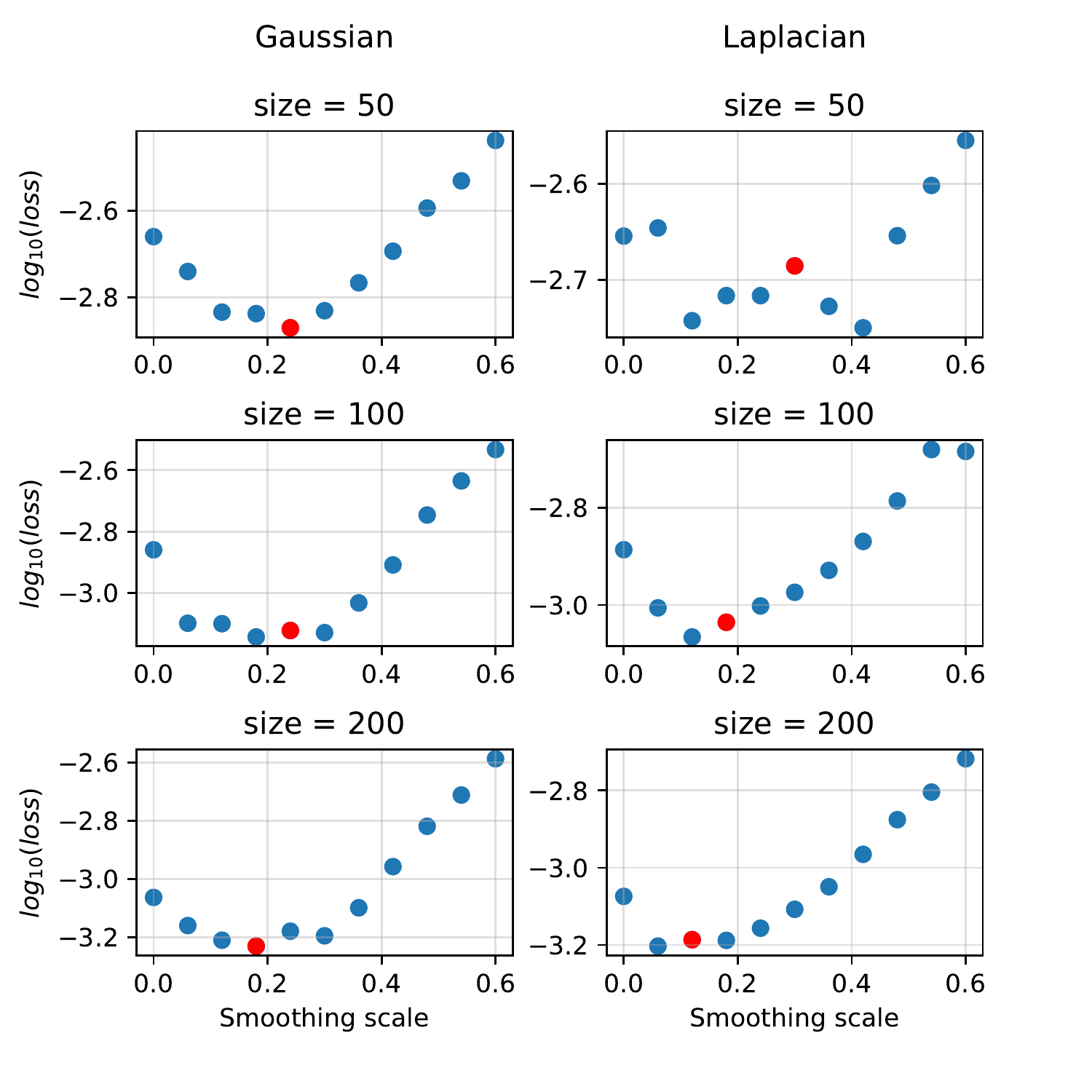}
    \end{tabular}
    \caption{Loss changes according to smoothing scale with training size increase from 50 to 200 in 1d data space. 
    % For D=2 and D=3, we refer to Appendix \ref{app:experiments}. 
    The red points represent the optimal smoothing scales selected based on the validation set.} 
    \label{fig:MSE_scale1d}
\end{figure}

\begin{figure}[!h]
    \centering
    \begin{tabular}{c|c}
        With Weight Decay & Early Stopping \\ 
        \includegraphics[width=7cm]{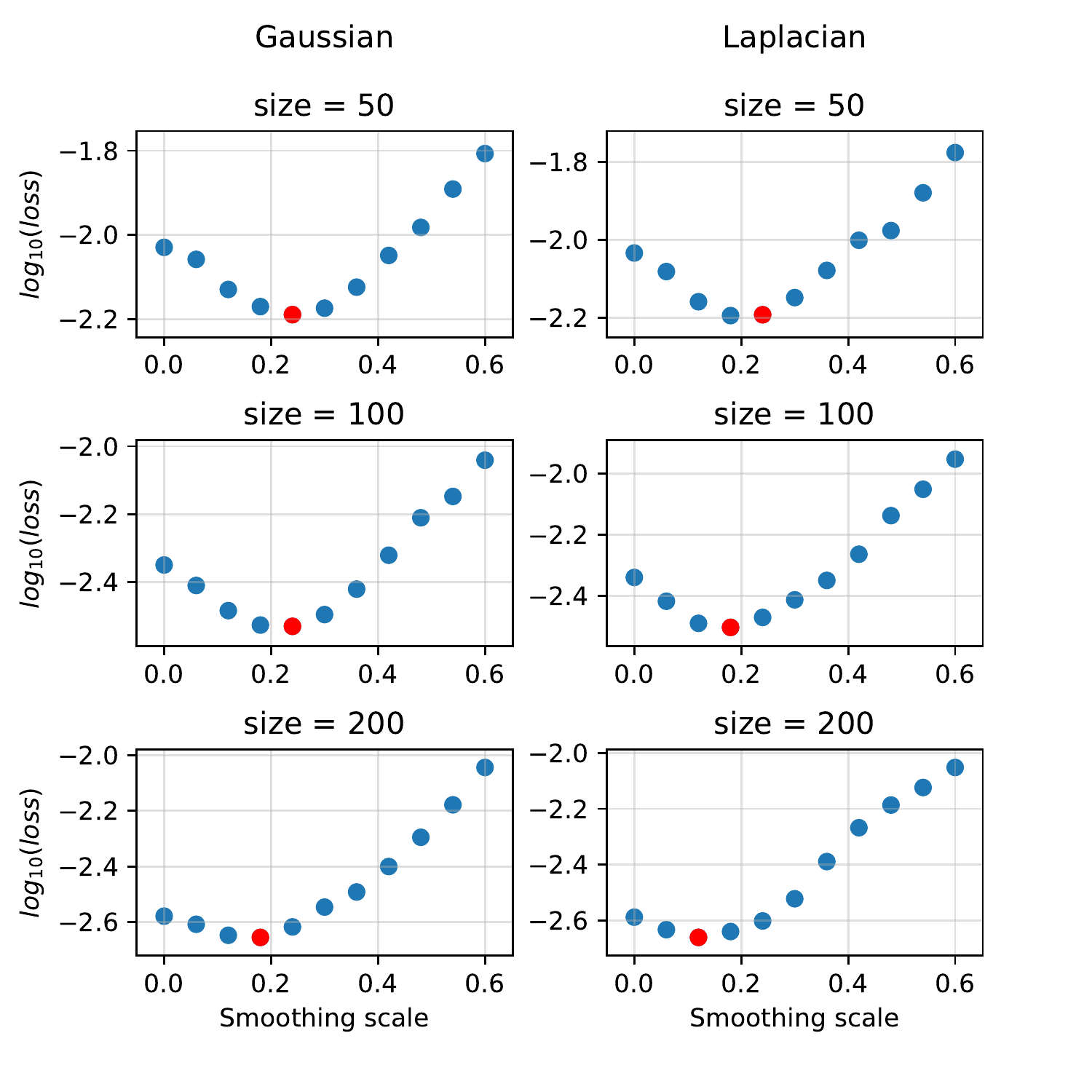} & \includegraphics[width=7cm]{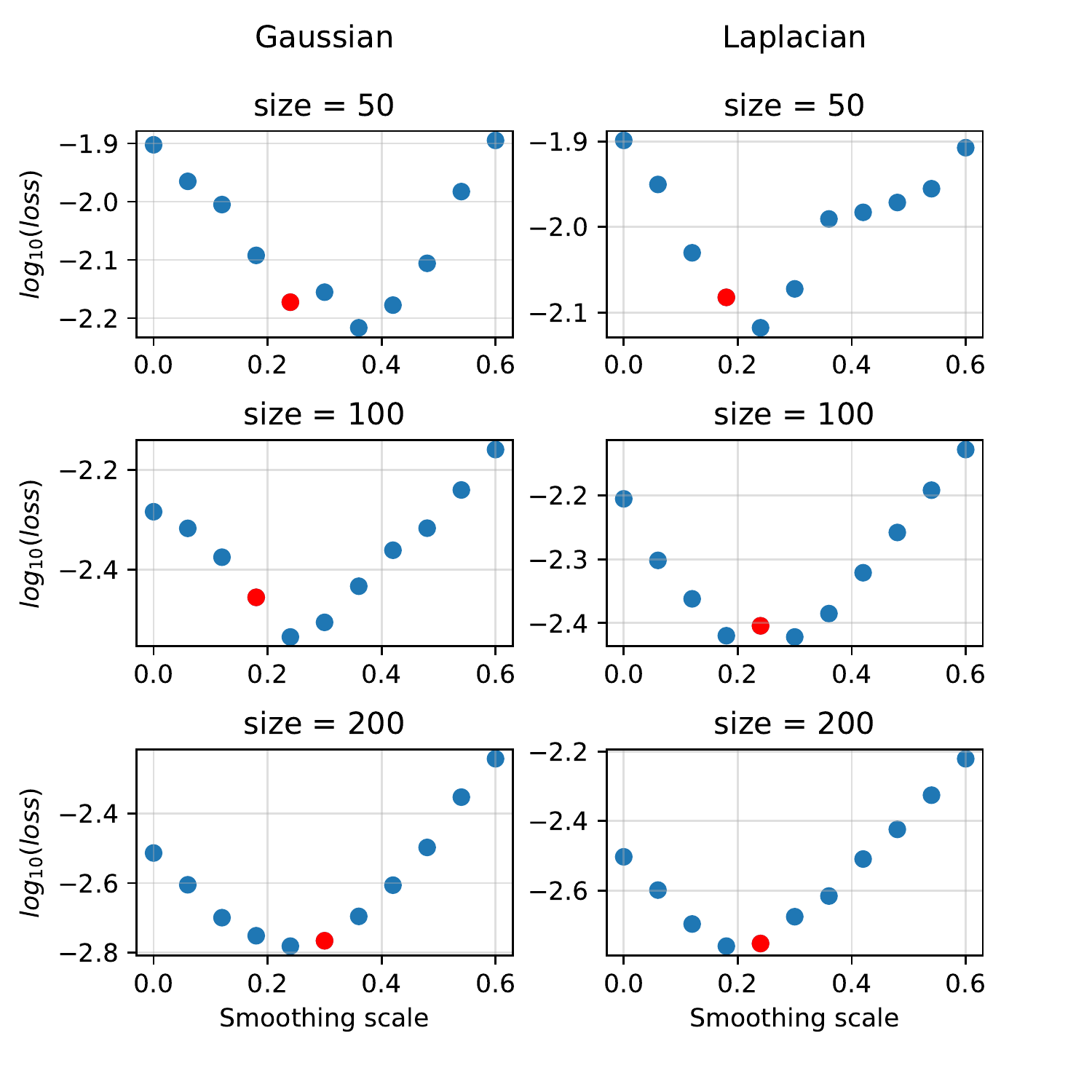}
    \end{tabular}
    \caption{Loss changes according to smoothing scale with training size increase from 50 to 200 in 2d data space. 
    % For D=2 and D=3, we refer to Appendix \ref{app:experiments}. 
    The red points represent the optimal smoothing scales selected based on the validation set.} 
    \label{fig:MSE_scale2d}
\end{figure}

\begin{figure}[!h]
    \centering
    \begin{tabular}{c|c}
        With Weight Decay & Early Stopping \\ 
        \includegraphics[width=7cm]{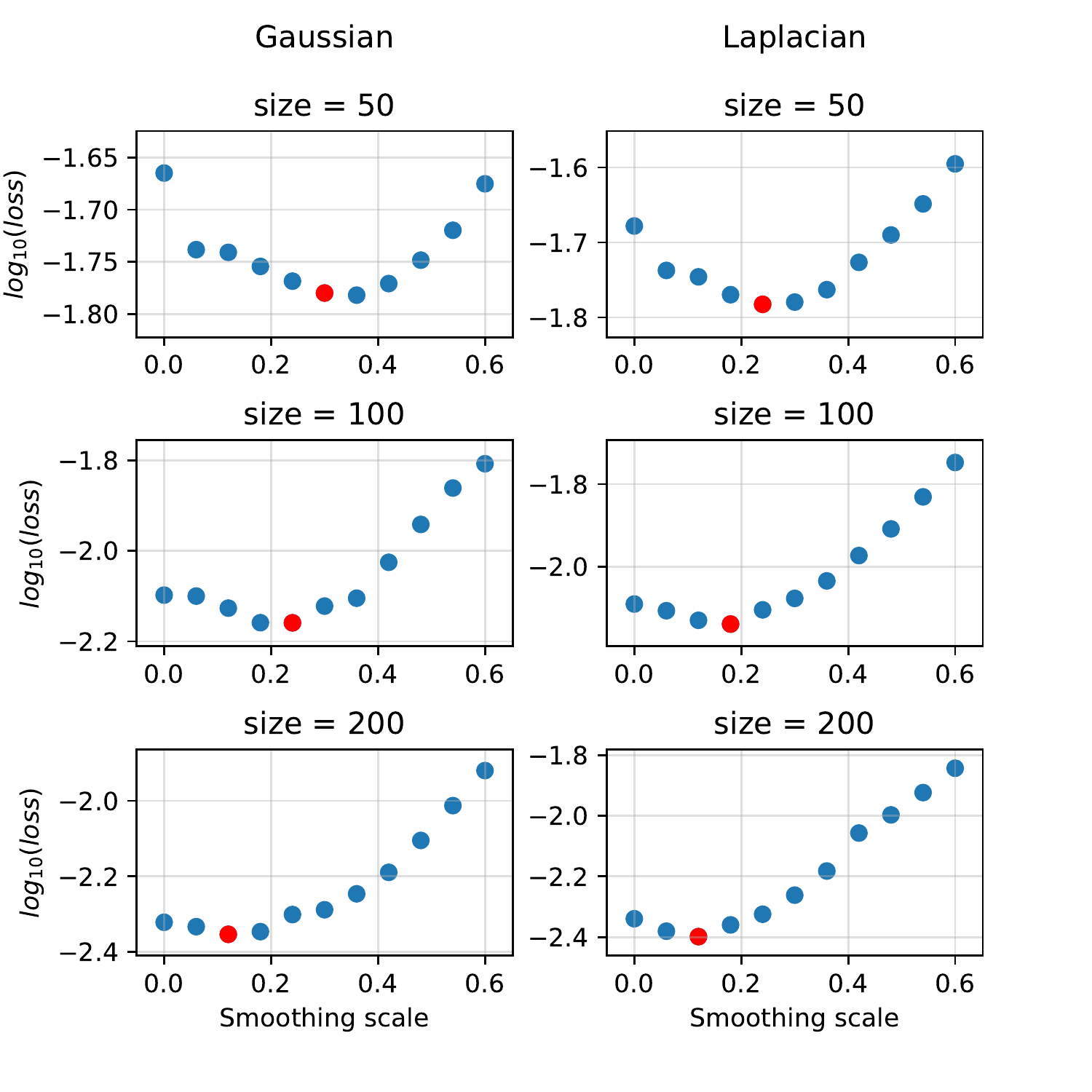} & \includegraphics[width=7cm]{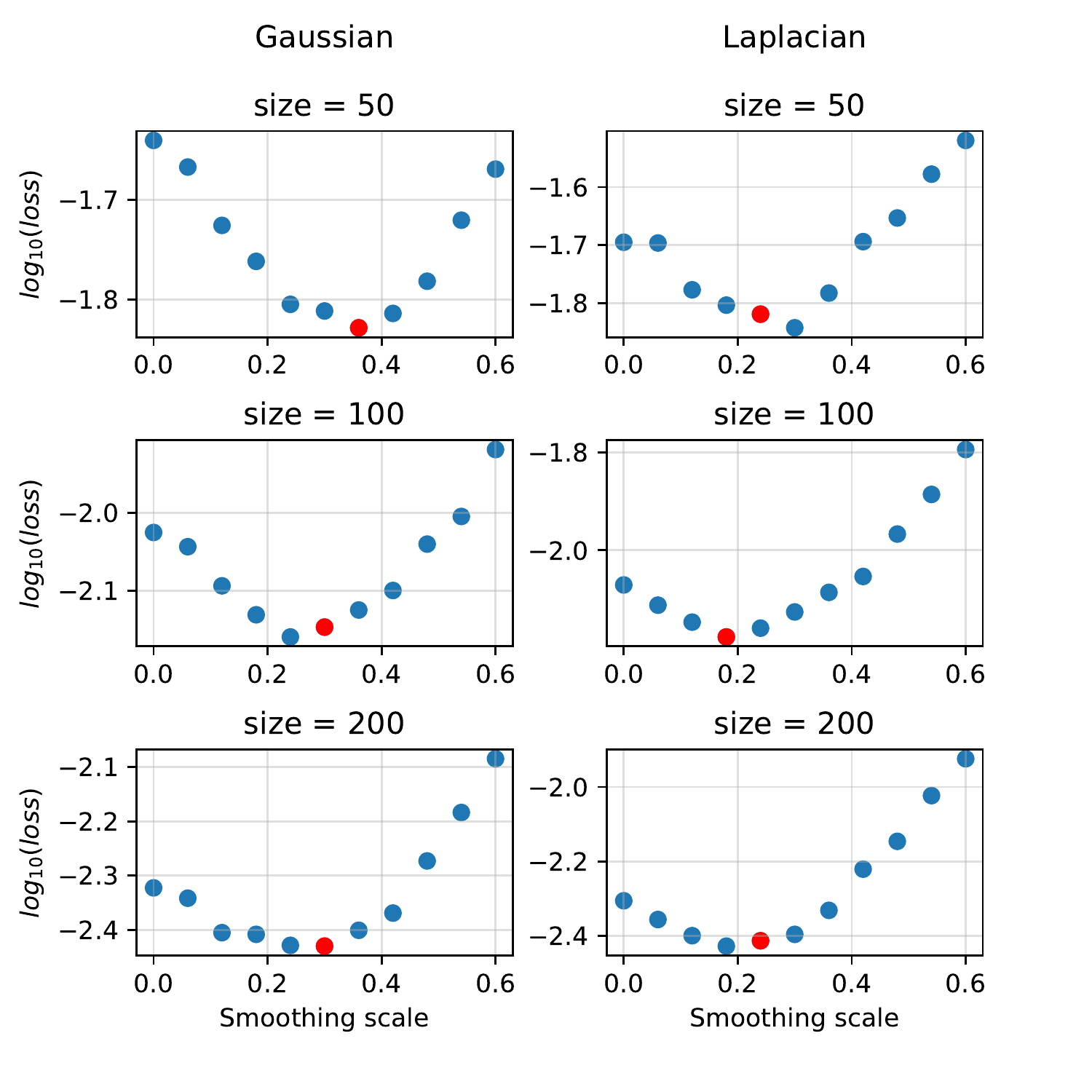}
    \end{tabular}
    \caption{Loss changes according to smoothing scale with training size increase from 50 to 200 in 3d data space. 
    % For D=2 and D=3, we refer to Appendix \ref{app:experiments}. 
    The red points represent the optimal smoothing scales selected based on the validation set.} 
    \label{fig:MSE_scale3d}
\end{figure}

Table \ref{tab:L2 Loss} presents a summary of the test $l_2$ loss under different settings. Both Gaussian smoothing and polynomial smoothing (random smoothing with Laplacian noise) improve the $l_2$ loss in all settings, demonstrating the effectiveness of random smoothing. Figure \ref{fig:MSE_scale1d} further investigates how the $l_2$ loss changes concerning the smoothing scale $\sigma_n$ %\yy{is it $\sigma_n$ or $m_\varepsilon$?}
when $D=1$. The plot shows a U-shaped curve, indicating that an optimal smoothing can minimize the $l_2$ loss, while either smaller or larger values will result in a larger $l_2$ loss. 
% The same properties can also be found when $D=2$ and $D=3$ (refer to Appendix \ref{app:experiments}). 
It is worth noting that when the training size is small, such as size 50, the U-shape curve in Figure \ref{fig:MSE_scale1d} may be less distinct due to noise introduced by early stopping based on a small validation set. Another observation from Figure \ref{fig:MSE_scale1d} is that the optimal smoothing scales exhibit a decreasing trend as the sample size increases, as indicated by Theorem \ref{thm:soboNN} and Theorem \ref{thm:soboGN}. Additionally, Figure \ref{fig:MSE_scale2d} and Figure \ref{fig:MSE_scale3d} depict the U-shaped curves of $l_2$ loss changes concerning smoothing scale when $D=2$ and $D=3$, respectively. While it is possible that some red points may not be accurately placed due to a small validation set, the optimal smoothing scales exhibit a decreasing trend with respect to training size, which is consistent with the trend observed in $D=1$ as depicted in Figure \ref{fig:MSE_scale1d}.  

\begin{table}[!h]
  \centering
  \caption{Test $l_2$ loss of SGD with early stopping. ``G'', ``L'', and ``N'' correspond to random smoothing with Gaussian noise, random smoothing with Laplacian noise, and no random smoothing. The smallest losses are underlined.}
    \begin{tabular}{|c|c|l|l|l|l|l|l|}
    \hline
    \multirow{3}[0]{*}{Dim} & \multirow{3}[0]{*}{Type} & \multicolumn{3}{c|}{With weight decay} & \multicolumn{3}{c|}{Early stopping} \\ 
        \cline{3-8}
          &       & \multicolumn{3}{c|}{Training size} & \multicolumn{3}{c|}{Training size} \\
          \cline{3-8}
          &       & \multicolumn{1}{c|}{50} & \multicolumn{1}{c|}{100} & \multicolumn{1}{c|}{200} & \multicolumn{1}{c|}{50} & \multicolumn{1}{c|}{100} & \multicolumn{1}{c|}{200} \\\hline
    \multirow{3}[0]{*}{D=1} & G     & 1.7466e-03 & 9.8343e-04 & 9.1924e-04 & \underline{1.3468e-03} & \underline{7.5579e-04} & \underline{5.8775e-04} \\
          & L     & \underline{1.6765e-03} & \underline{9.3367e-04} & \underline{8.2806e-04} & 2.0638e-03 & 9.2128e-04 & 6.5118e-04 \\
          & N     & 1.9381e-03 & 1.3045e-03 & 1.1135e-03 & 2.2168e-03 & 1.2985e-03 & 8.4292e-04 \\\hline
    \multirow{3}[0]{*}{D=2} & G     & \underline{6.4208e-03} & 3.1423e-03 & \underline{2.1842e-03} & \underline{6.7205e-03} & \underline{3.5027e-03} & \underline{1.7132e-03} \\
          & L     & 6.4676e-03 & \underline{2.9491e-03} & 2.2136e-03 & 8.2725e-03 & 3.9418e-03 & 1.7674e-03 \\
          & N     & 9.2474e-03 & 4.5782e-03 & 2.5810e-03 & 1.2628e-02 & 6.2301e-03 & 3.1396e-03 \\\hline
    \multirow{3}[0]{*}{D=3} & G     & \underline{1.6498e-02} & 7.2578e-03 & \underline{3.9938e-03} & \underline{1.4852e-02} & 7.1306e-03 & \underline{3.7147e-03} \\
          & L     & 1.6599e-02 & \underline{6.9336e-03} & 4.4334e-03 & 1.5167e-02 & \underline{6.6471e-03} & 3.8615e-03 \\
          & N     & 2.0987e-02 & 8.1158e-03 & 4.5752e-03 & 2.0178e-02 & 8.4932e-03 & 4.9460e-03 \\\hline
    \end{tabular}%
  \label{tab:L2 Loss}%
\end{table}%

\section{Conclusions and Discussion}\label{sec:conclusion}

This work studies random smoothing kernel and random smoothing regularization, which have a natural relationship with data augmentations. We consider two cases: when the region $\Omega$ has a low intrinsic dimension, or when the kernel function can be presented as a product of one-dimensional kernel functions. In both cases, we show that by applying random smoothing, with appropriate early stopping and/or weight decay techniques, the resulting estimator can achieve fast convergence rates, regardless of the kernel function used in the construction of the random smoothing kernel estimator.

There are several directions that could be pursued in future research. First, while we consider noise injection to construct augmentations and use non-smooth noise to interpret practical non-smooth augmentation techniques, such as random crop, random mask, and random flip, this interpretation may not be perfect. For example, the behavior of adding noise may differ from that of random crop. Furthermore, these practical techniques may also introduce some prior knowledge on the geometry of the low intrinsic dimension. A sharper characterization of practical augmentation techniques is needed and will be pursued in future work.

Second, while we consider gradient descent, we believe that our results can be generalized to the stochastic gradient descent method. However, the discussion of the latter is beyond the scope of the current work.

Third, we mainly consider regression in this work, where the square loss is a natural choice. An interesting extension is to study whether the results remain true when considering classification, which requires the study of other loss functions, such as cross-entropy loss and hinge loss.

% In both cases, the convergence rates are obtained, which indicates that adding augmentations can help with improving the convergence speed. 

% \ww{sdafkl;jasd;kljfds;kasdfj}

\bibliographystyle{apalike}
\bibliography{ref}

\newpage
\setcounter{page}{1}

\appendix

% \section{Reproducing kernel Hilbert space and Sobolev space}\label{subsec:introrkhs}

% In this section, we provide a brief introduction to reproducing kernel Hilbert spaces and Sobolev spaces. 

% Suppose $\Omega \subset \mathbb{R}^d$ has a positive Lebesgue measure and Lipschitz boundary. Assume that 

\section{Analysis of Gradient Update and Error Decomposition}\label{app:GD}
Let $\Xb = (\bx_1,...,\bx_n)$, $\alpha>0$ if there is weight decay, and $\alpha=0$ if there is no weight decay. By the gradient update rule, we have 
\begin{align*}
    f_{t+1}(\Xb) = & \Kb\bw_{t+1} = \sqrt{\Kb}\btheta_{t+1} \nonumber\\
    = & \sqrt{\Kb}\btheta_t - \beta\sqrt{\Kb} \left(\Kb\btheta_t-\sqrt{\Kb}\by\right) - \alpha\sqrt{\Kb}\btheta_t\nonumber\\
    = & ((1-\alpha)\Ib - \beta\Kb) f_t(\Xb) + \beta\Kb\by,
\end{align*}
which implies
\begin{align}\label{eq:gdupd1}
    & f_{t+1}(\Xb) - \beta(\alpha\Ib + \beta\Kb)^{-1}\Kb\by = ((1-\alpha)\Ib - \beta\Kb) (f_t(\Xb)-\beta(\alpha\Ib + \beta\Kb)^{-1}\Kb\by)\nonumber\\
    =&  \ldots = -((1-\alpha)\Ib - \beta\Kb)^{t+1}\beta(\alpha\Ib + \beta\Kb)^{-1}\Kb\by,
\end{align}
where we recall $f_0(\Xb) = \mathbf{0}$. If there is weight decay (i.e., $\alpha>0$), then it can be seen that
\begin{align}\label{eq:gdupdwd1}
    & f_{t+1}(\Xb) - \Kb(\alpha/\beta \Ib + \Kb)^{-1}\by = -((1-\alpha)\Ib - \beta\Kb)^{t+1}\beta(\alpha\Ib + \beta\Kb)^{-1}\Kb\by.
\end{align}
If there is no weight decay (i.e., $\alpha=0$), then by rearrangement of \eqref{eq:gdupd1}, we obtain
\begin{align}\label{eq:gdupdnwd1}
    f_{t+1}(\Xb) = & \left(\Ib-(\Ib - \beta\Kb)^{t+1}\right)\by.
    % \nonumber\\
    % = & \left(\Ib-((1-\alpha)\Ib - \beta\Kb)^{t+1}\right)\by - \alpha \left(\Ib-((1-\alpha)\Ib - \beta\Kb)^{t+1}\right)\by.
\end{align}
The estimator after $t$-th iteration can be obtained by
\begin{align}\label{eq:predictorapp}
     f_t(\bx) = & \bw_t^T \kb(\bx) = \kb(\bx)^T \Kb^{-1}  f_t(\Xb).
\end{align}
Note that the kernel matrix $\Kb$ is generated by the empirical kernel $K_S$ defined in \eqref{eq:empirical_kernel}. By taking the expectation with respect to $\bvarepsilon_{k_1}$ and $\bvarepsilon_{k_2}$, we define the expected smoothing kernel $\tilde K_S$ as 
\begin{align}\label{eq:Psit}
    \tilde K_S(\bx,\bx') = \int_{\RR^D}\int_{\RR^D}K(\bx+\bepsilon-(\bx'+\bepsilon'))p_\varepsilon(\bepsilon)p_\varepsilon(\bepsilon'){\rm d}\bepsilon{\rm d}\bepsilon'.
\end{align}
Since $\tilde K_S$ is close to the empirical version of the smoothing kernel $K_S$, we can consider the gradient flow with respect to the kernel function $\tilde K_S$. The error analysis between $\tilde K_S$ and $K_S$ is provided in Appendix \ref{app:PsistPsis}.

Let $g_{t}$ be the function obtained at $t$-th iteration by the gradient update rule with respect to the kernel function $\tilde K_S$. Analogous to \eqref{eq:gdupdwd1} and \eqref{eq:gdupdnwd1}, we have 
\begin{align}\label{eq:gdupdwdg1}
    & g_t(\Xb)  = \tilde\Kb(\alpha/\beta \Ib + \tilde\Kb)^{-1}\by -((1-\alpha)\Ib - \beta\tilde\Kb)^t\beta(\alpha\Ib + \beta\tilde\Kb)^{-1}\Kb\by,
\end{align}
if there is weight decay, and 
\begin{align}\label{eq:gdupdnwdg1}
    g_t(\Xb) = & \left(\Ib-(\Ib - \beta\tilde\Kb)^t\right)\by,
    % \nonumber\\
    % = & \left(\Ib-((1-\alpha)\Ib - \beta\Kb)^{t+1}\right)\by - \alpha \left(\Ib-((1-\alpha)\Ib - \beta\Kb)^{t+1}\right)\by.
\end{align}
if there is no weight decay, where 
% $\tilde \kb(\cdot) = (\tilde K_S(\cdot-\bx_1),\ldots,\tilde K_S(\cdot-\bx_n))^T$ and 
$\tilde\Kb = (\tilde K_S(\bx_j-\bx_k))_{jk}$. Similarly, the predictor of $f^*(\bx)$ using the kernel function $\tilde K$ can be obtained by 
\begin{align}\label{eq:predictorappg}
    g_t(\bx) = \tilde\kb(\bx)^T \tilde\Kb^{-1} g_t(\Xb).
\end{align}
Thus, the empirical error $\|f_t(\Xb) - f^*(\Xb)\|_2$ can be decomposed by
\begin{align}\label{eq:errdecompose1}
    \|f_t(\Xb) - f^*(\Xb)\|_2 \leq \|f_t(\Xb) - g_t(\Xb)\|_2 + \|g_t(\Xb) - f^*(\Xb)\|_2.
\end{align}
% The first term will be analyzed in Appendix \ref{app:PsistPsis}, and the second term with $\alpha = 0$ and $\alpha > 0$ will be analyzed in Appendix \ref{app:gderrorPsit} and Appendix ???, respectively.

% which implies,
% \begin{align}\label{eq:gdupdnowd1}
%     f_{t+1}(\Xb) - f^*(\Xb) = -(\Ib - \beta\Kb)^{t+1}f^*(\Xb) + \left(\Ib-(\Ib - \beta\Kb)^{t+1}\right)\bepsilon.
% \end{align}
% Next, we consider \eqref{eq:gdupdwd1} and \eqref{eq:gdupdnowd1} separately. In the following, let $\tilde \kb(\cdot) = (\tilde K_S(\cdot-\bx_1),\ldots,\tilde K_S(\cdot-\bx_n))^T$ and $\tilde\Kb = ((\tilde K_S(\bx_j-\bx_k))_{jk}$, with 

% {\bf We also need to deal with $\tilde g_{t+1}(\Xb)^T\tilde R^{-1}\tilde g_{t+1}(\Xb)$, which is the RKHS norm of $\tilde g_{t+1}$.}

% Use the basic inequality $2\ba^T\bb \geq -C_1\|\ba\|_2^2 - \frac{1}{C_1}\|\bb\|_2^2$, we have
% \begin{align*}
%     & \frac{2n(\alpha t)^{-1}}{n}\bepsilon^T(\tilde\Kb + n(\alpha t)^{-1}\Ib)^{-1}\tilde\Kb(\tilde\Kb + n(\alpha t)^{-1}\Ib)^{-1}f^*(\Xb)\nonumber\\
%     \geq & - C_1\frac{n^2(\alpha t)^{-2}}{n} (f^*(\Xb))^T (\tilde\Kb + n(\alpha t)^{-1}\Ib)^{-2}f^*(\Xb) -\frac{1}{C_1n}\bepsilon^T(\tilde\Kb + n(\alpha t)^{-1}\Ib)^{-1}\tilde\Kb^2(\tilde\Kb + n(\alpha t)^{-1}\Ib)^{-1}\bepsilon
% \end{align*}
% Thus, 
% \begin{align*}
%     \|\tilde g_t - f^*\|_n^2 \leq & J_{21} + J_{22}\nonumber\\
%     \leq & \frac{(2C+C_1)n^2(\alpha t)^{-2}}{n} (f^*(\Xb))^T (\tilde\Kb + n(\alpha t)^{-1}\Ib)^{-2}f^*(\Xb)
% \end{align*}

% Also, 

\section{Error of Data Augmentation}\label{app:PsistPsis}

We first consider bounding the difference between the empirical smoothing kernel function  
\begin{align*}
    K_S(\bx-\bx') = \frac{1}{N^2}\sum_{k=1}^N \sum_{j=1}^N K(\bx+\bvarepsilon_j-( \bx'+\bvarepsilon_k))
\end{align*}
and the expected smoothing kernel function
\begin{align*}
    \tilde K_S(\bx-\bx') = \EE_{\bvarepsilon,\bvarepsilon'}\big( K(\bx+\bvarepsilon-(\bx'+\bvarepsilon'))\big) = \int_{\RR^D}\int_{\RR^D}K(\bx+\bvarepsilon-(\bx'+\bvarepsilon'))p_\varepsilon(\bvarepsilon)p_\varepsilon(\bvarepsilon'){\rm d}\bvarepsilon{\rm d}\bvarepsilon'.
\end{align*}
Specifically, we have the following lemma.
\begin{lemma}\label{lem:converge_empirical_kernel}
If Assumption \ref{assum:PsiDecay} or \ref{assum:PsiDecay_tensor}, and Assumption \ref{assum:augnoise} are satisfied , then
\begin{align}
    &\sup_{\bx,\bx'\in \Omega}\left|\EE_{\bvarepsilon,\bvarepsilon'}\big( K(\bx+\bvarepsilon-(\bx'+\bvarepsilon'))\big)- \frac{1}{N^2}\sum_{k=1}^N \sum_{j=1}^N K(\bx+\bvarepsilon_j-(\bx'+\bvarepsilon_k))\right|=O_{\PP}\left(\sqrt{\frac{\log N}{N}}\right).\nonumber
\end{align}
\end{lemma}

Based on Lemma \ref{lem:converge_empirical_kernel}, we can obtain an upper bound of $\|f_t-g_t\|_{L_\infty(\Omega)}$ as follows. Recall that $\Kb = (K_S(\bx_j-\bx_k))_{j,k=1}^n$, $\tilde \Kb = (\tilde K_S(\bx_j-\bx_k))_{j,k=1}^n$. Let $\eta_1(\Kb)$ and $\eta_n(\Kb)$ be the largest and smallest eigenvalues of $\Kb$, respectively. Let $\eta_n(\tilde \Kb)$ be the smallest eigenvalue of $\tilde \Kb$.

% \begin{assumption}\label{assum_stepsize}
% Let $\eta_1$ be the largest eigenvalue of $\Kb$, where $\Kb$ is as in \eqref{eq_lossX}. The learning rate $\beta$ satisfies $\beta\eta_1 +\alpha < 1$, where $\alpha=0$ if there is no weight decay, and $\alpha>0$ if there is weight decay.

% % If there is no weight decay, the learning rate $\beta$ satisfies $\beta\lambda_1 < 1$ and $1 -\beta\lambda_n < C < 1$. If there is weight decay, the learning rate $\beta$ and the weight decay parameter $\alpha$ satisfies $\beta\lambda_1 + \alpha < 1$ and $1-\alpha -\beta\lambda_n < C < 1$.
% % We assume that $\beta\lambda_1 + \alpha < 1$ and $1-\alpha -\beta\lambda_n =s < 1$, where $s$ is some constant independent of $n$, $\lambda_1$ is the largest eigenvalue of $\Kb$, and $\lambda_n$ is the smallest eigenvalue of $\Kb$. 
% \end{assumption}

% and $\eta_n(\tilde \Kb), \eta_n(\Kb)$ are the smallest eigenvalue of $\tilde \Kb$ and $\Kb$, respectively.

\begin{lemma}\label{lem_gftinf}
Suppose Assumption \ref{assum:PsiDecay} or \ref{assum:PsiDecay_tensor}, and Assumption \ref{assum:augnoise} are satisfied. Furthermore, assume that
\begin{align}\label{eq:lem_gftinf}
    \frac{1}{2}\eta_n(\tilde \Kb)\geq n\sqrt{\frac{\log N}{N}},
\end{align}
and the learning rate $\beta$ satisfies $\beta\eta_1(\Kb) +\alpha < 1$, where $\alpha=0$ if there is no weight decay, and $\alpha>0$ if there is weight decay.
Then we have
\begin{align*}
    \sup_{t\geq 1}\|f_t-g_t\|_{L_\infty(\Omega)}=O_\PP\left(\frac{n^2\sqrt{\log N/N}}{\eta_{n}(\tilde \Kb)^2}\right),
\end{align*}
where the probability is with respect to the augmentation $\bvarepsilon$.
\end{lemma}

Since $\tilde \Kb$ and $\eta_n(\tilde \Kb)$ are determined by the data $(\bx_j,y_j)$, $j=1,...,n$, the left-hand side of \eqref{eq:lem_gftinf} is not depending on $N$. Therefore, the condition \eqref{eq:lem_gftinf} can be fulfilled if we add sufficient augmentations. In the next lemma, we provide a more explicit lower bound of $\eta_n(\tilde \Kb)$ in \eqref{eq:lem_gftinf} in terms of $\bx_j$'s.

\begin{lemma}\label{lem:minimum_eigvalue}
Let $q_{\Xb}$ be the separation distance defined as
\[q_{\Xb}=\frac{1}{2}\min_{j\neq k}\|\bx_j-\bx_k\|_2.\]
The minimum eigenvalue of $\tilde\Kb$, denoted by $\eta_n(\tilde \Kb)$, is lower bounded as follows.
\begin{enumerate}
    \item if Assumption \ref{assum:PsiDecay} and Assumption \ref{assum:augnoise} (C1) are satisfied, then
    \[\eta_n(\tilde \Kb)\geq  C_{1}\big(1+4M^2\big)^{-m_0}\big(1+4\sigma_n^2M^2\big)^{-m_\varepsilon}M^{D};\]
    \item if Assumption \ref{assum:PsiDecay_tensor} and Assumption \ref{assum:augnoise} (C2) are  satisfied, then
    \[\eta_n(\tilde \Kb)\geq  C_{2}\big(1+4M^2\big)^{-m_0D}\big(1+4\sigma_n^2M^2\big)^{-m_\varepsilon D}M^{D};\]
    \item if Assumption \ref{assum:PsiDecay} and Assumption \ref{assum:augnoise} (C3) are  satisfied, then
    \[\eta_n(\tilde \Kb)\geq  C_{3}\big(1+4M^2\big)^{-m_0}e^{-8\sigma_n^2M^2}M^{D}\]
\end{enumerate}
where $C_i$'s are constants only depending on $D$, $M=\frac{12}{q_\Xb}\big(\frac{\pi\Gamma^2(\frac{D}{2}+1)}{9}\big)^{\frac{1}{D+1}}$, 
and $\Gamma(\cdot)$ denotes the Gamma function.
\end{lemma}
The proofs of the above three lemmas are put in Appendix \ref{app:pffinB}.

% \section{Further Characterization of $\cH_{\tilde K_S}(\Omega)$}\label{app:smoothRKHS}

% {\bf did not polish this section}

% In this section, we consider the RKHS $\cH_{\tilde K_S}(\Omega)$ generated by $\tilde K_S$ defined in \eqref{eq:Psit}.

% The embedding in \eqref{eq:appGauEb1} and \eqref{eq:appGauEb2} establish a relationship between the RKHSs generated by the random smoothed kernel and the Gaussian kernel, which allows us to utilize the existing results in the Gaussian kernel RKHS, e.g., the entropy number.

% \subsection{Non-smooth Noise}
% If (C2) noise, then 
% \begin{align}\label{eq:NNfourier}
%     (1+\max(\sigma_n^2,1)\|\bomega\|_2^2)^{m_0+m_\varepsilon}\geq \cF(K)(\bomega)|\varphi_\varepsilon(\bomega)|^2 \geq (1+\min(\sigma_n^2,1)\|\bomega\|_2^2)^{m_0+m_\varepsilon}
% \end{align}
% Then reduce to Sobolev case. However, low intrinsic dimension case.

% % If there is weight decay, we need additionally $m_0+m_\varepsilon \geq m_f$.

% \subsection{(C3) Noise}

% Jiao gei Ding Liang

% which implies $\|f_n^*\|_{\cH_{\tilde K_S}(\Omega)}\geq C_2\sigma_n^{2m_0-d}\|f_n^*\|_{\cH_{\sigma_n/\sqrt{2}}(\Omega)}$.

\section{A Comparison Theorem}\label{sec:connecttoKRR}

In this section, we provide a byproduct, which is a generic comparison theorem between the early-stopping without weight decay and the kernel ridge regression estimator. Let $K_1$ be a positive definite kernel function. The kernel ridge regression is defined by
\begin{align}\label{eq:complemkrr}
    \tilde g = \argmin_{f\in \cH_{K_1}(\Omega)} \|f - \by\|_n^2 + \lambda \| f\|_{\cH_{K_1}(\Omega)}^2,
\end{align}
where $\by=(y_1,...,y_n)^T$, $y_j$'s are as in \eqref{eq:model}, and $\lambda>0$ is a regularization parameter. The main theorem in this subsection is as follows.

% This establishes a relationship between early-stopping method and kernel ridge regression.
% which shows that the mean squared prediction error of early-stopping (without weight decay) is at least as better as the kernel ridge regression estimator.

\begin{theorem}\label{lem:compLemma}
Let $(\beta t)^{-1} = n\lambda$. Suppose $\epsilon_j$'s are i.i.d. random noise with mean zero and finite variance $\sigma_\epsilon^2$. Let $\tilde g_t(\bx) = \bw_t^T\kb(\bx)$, which is similar to $\hat f_t(\bx)$ in \eqref{eq:predictor} but with $K_1$ instead of $K_S$ and with update rule \eqref{eq_GD_nowd}. Then there exists a constant $C>0$ such that
\begin{align}\label{eq:complemeq1}
    \EE\|\tilde g_t-f^*\|_n^2 \leq C\EE\|\tilde g-f^*\|_n^2,
\end{align}
and 
\begin{align}\label{eq:gtRKHSnorm}
    \EE\|\tilde g_t\|_{\cH_{K_1}(\Omega)}^2 \leq 2\EE\|\tilde g\|_{\cH_{K_1}(\Omega)}^2,
\end{align}
where the expectation is taken with respect to the noises $\epsilon_j$, $j=1,...,n$.
\end{theorem}

Theorem \ref{lem:compLemma} states that the mean squared prediction error of the early-stopping without weight decay is smaller than (at most the same as) that of the kernel ridge regression estimator, up to a multiplicative constant. This explains why the upper bounds on the early-stopping without weight decay and the kernel ridge regression estimator derived in \cite{raskutti2014early} are identical, in a more explicit way. Note that the conditions of Theorem \ref{lem:compLemma} are quite mild. We do not assume any relationship between $f^*$ and  $\cH_{K_1}(\Omega)$, and do not require any particular structure of the RKHS $\cH_{K_1}(\Omega)$. Furthermore, we do not impose any conditions on $\lambda$, and we only require that $\epsilon_j$'s are i.i.d. with finite variance (not necessarily sub-Gaussian and can be even heavy-tailed). 

It is worth noting that the complexity (i.e., the RKHS norm) of the early-stopping without weight decay is also bounded by the complexity of the kernel ridge regression estimator, up to a constant multiplier. Since the difference between the empirical norm $\|\cdot\|_n$ and the $L_2$ norm depends on the complexity of the estimator, it can be expected that \eqref{eq:complemeq1} still holds if we replace the empirical norm by the $L_2$ norm.

\section{Proof of Theorem \ref{thm:soboNN}}\label{app:pfthmsoboNN}

% We split the proof of Theorem \ref{thm:soboNN} into two parts, where the first part is for polynomial smoothing with fixed $m_\varepsilon$, and the second part is for polynomial smoothing with $m_\varepsilon\geq ??? \log n$.

% \subsection{Polynomial Smoothing with Fixed $m_\varepsilon$}

In this section, we show the proof of the following theorem. Note that the second statement in Theorem \ref{thm:soboNNapp} is Theorem \ref{thm:soboNN}.

\begin{theorem}[Polynomial smoothing]\label{thm:soboNNapp}
Suppose Assumptions \ref{assum:sub-G}, \ref{assum:PsiDecay}, \ref{assum:augnoise} (C1), and \ref{assum:intriD} are satisfied. Suppose there exists $\Omega_1$ with positive Lebesgue measure and a Lipschitz boundary such that $\Omega\subset \Omega_1$ and $f^*\in \cW^{m_f}(\Omega_1)$. Let $f_t(\bx)$ be as in \eqref{eq:predictor} and $\beta=n^{-1}C_1$ with $C_1\leq 2^{-1}\sup_{\bx\in \RR^D}K_S(\bx)$. Suppose the smoothing scale
% \yy{for polynomial smoothing, $\sigma_n$ appears in the scaling of characteristic functions. Is that the variance?} 
% of the augmentation noise 
$\sigma_n\asymp n^{\nu}$ with $\nu \leq 0$. Suppose one of the following holds:
\begin{enumerate}
    \item There is no weight decay in the gradient descent, and the iteration number $t$ satisfies
    $
        t \asymp  n^{\frac{2(m_0+m_\varepsilon)}{2m_f+d}}\sigma_n^{2m_\varepsilon}
    $
    \item There is weight decay in the gradient descent with $\alpha \asymp n^{-1-\frac{2(m_0+m_\varepsilon)}{2m_f+d}}\sigma_n^{-2m_\varepsilon}$, and the iteration number satisfies $t\geq C_2(\frac{m_f}{2m_f+d}+1/2)\log n/(\log (1-\alpha))$.
\end{enumerate}
Then the following statements are true with $N>N_0$, where $N$ is the number of augmentations, and $N_0$ depends on $n$ and the iteration number $t$. 
\begin{enumerate}
    \item For any $a>0$, there exists an $m_\varepsilon$ such that when 
    \begin{align*}
        \nu =\left\{\begin{array}{ll}
            -\frac{2(2m_0+2m_\varepsilon)D-(2m_0+2m_\varepsilon-D)d}{(2m_f+d)(4m_\varepsilon D-(2m_0+2(1-d^{-1}(2m_f+d)a)m_\varepsilon -D)d)}, & D>d, \\
            0, & D=d,
        \end{array}\right.        
    \end{align*}
    we have
    \begin{align*}
        \|f_t - f^*\|_{L_2(P_\Xb)}^2 = & O_{\PP}\left(n^{-\frac{2m_f}{2m_f + d}+a}\right).
    \end{align*}
    \item Set $m_\varepsilon = 2d^{-1}(2D \max(m_0,m_f)+ m_0d)\log n - m_0$. Then by choosing 
    \begin{align*}
        \nu =\left\{\begin{array}{ll}
             -\frac{2(2m_0+2m_\varepsilon)D-(2m_0+2m_\varepsilon-D)d}{(2m_f+d)(4m_\varepsilon D-(2m_0+2(1-(\log n)^{-1})m_\varepsilon -D)d)} < 0, & D>d, \\
            0, & D=d,
        \end{array}\right.                
    \end{align*}
    we have
    \begin{align*}
        \|f_t - f^*\|_{L_2(P_\Xb)}^2 = & O_{\PP}\left(n^{-\frac{2m_f}{2m_f + d}}(\log n)^{2m_f+1}\right).
    \end{align*}
\end{enumerate}
\end{theorem}

We first present several lemmas used in this proof. The proof of these lemmas can be found in Appendix \ref{app:pfappC}.

\begin{lemma}\label{lem:approx1NN}
Suppose the conditions of Theorem \ref{thm:soboNN} are fulfilled. Let $f_n^*$ be the solution to the optimization problem
\begin{align}\label{eq:krrestoraNN1}
    \min_{g\in \cH_{\tilde K_S}(\Omega)}\|f^*-g\|_{L_2(P_\Xb)}^2 + \lambda_n\|g\|_{\cH_{\tilde K_S}(\Omega)}^2.
\end{align}
Then if $m_0\leq m_f$, we have 
\begin{align}\label{eq:pflemapprox1NNgoal1}
    \|f^*-f_n^*\|_{L_2(P_\Xb)}^2 + \lambda_n\|f_n^*\|_{\cH_{\tilde K_S}(\Omega)}^2 \leq C_1\max\left((\lambda_n(m_\varepsilon + 1)^{m_\varepsilon}\sigma_n^{2m_\varepsilon})^{\frac{m_f}{m_0+m_\varepsilon}},\lambda_n\right).
\end{align}
and 
if $m_0> m_f$, we have 
\begin{align}\label{eq:pflemapprox1NNgoal2}
    \|f^*-f_n^*\|_{L_2(P_\Xb)}^2 + \lambda_n\|f_n^*\|_{\cH_{\tilde K_S}(\Omega)}^2 \leq C_2 \max\left((\lambda_n(m_\varepsilon + 1)^{m_\varepsilon}\sigma_n^{2m_\varepsilon})^{\frac{m_f}{m_0+m_\varepsilon}},\lambda_n^{\frac{m_f}{m_0}}\right).
\end{align}
Here the constants $C_1$ and $C_2$ are independent with $m_\varepsilon$.

% \begin{align}\label{eq:pflemapprox1NNgoal}
%     \|f^*-f_n^*\|_{L_2(P_\Xb)}^2 + \lambda_n\|f_n^*\|_{\cH_{\tilde K_S}(\Omega)}^2 \leq \max\left((\lambda_n\sigma_n^{2m_\varepsilon})^{\frac{m_f}{m_0+m_\varepsilon}},(\lambda_n\sigma_n^{-2m_0})^{\frac{m_f}{m_0+m_\varepsilon}}\right).
% \end{align}
% and 
% \begin{align*}
%     \|f_n^*\|_{\cH_{\tilde K_S}(\Omega)}^2 \leq \max\left(\lambda_n^{-1}(\lambda_n\sigma_n^{2m_\varepsilon})^{\frac{m_f}{m_0+m_\varepsilon}},\sigma_n^{-2m_0}\right).
% \end{align*}
\end{lemma}

\begin{lemma}\label{lem:approx2NN}
Suppose the conditions of Theorem \ref{thm:soboNN} are fulfilled. Let $f_n^*$ be as in Lemma \ref{lem:approx1NN}.
Suppose there exists $T>0$ (depending on $n$) such that
\begin{align*}
    \|f^*-f_n^*\|_{L_2(P_\Xb)}^2 + \lambda_n\|f_n^*\|_{\cH_{\tilde K_S}(\Omega)}^2\leq T.
\end{align*}
Let $\hat f_n$ be the solution to the optimization problem
\begin{align}\label{eq:lem2approx2NN}
    \min_{g\in \cH_{\tilde K_S}(\Omega)}\|\by-g\|_n^2 + \lambda_n\|g\|_{\cH_{\tilde K_S}(\Omega)}^2,
\end{align}
where $\by = (y_1,...,y_n)^T$. 
Suppose 
\begin{align*}
    \sigma_n^{-d/2}m^{\frac{mD}{2m-D}+\frac{1}{2}}\log p
\end{align*}
converges to zero as $n$ goes to infinity, where $p = \frac{4D}{2m-D}$, and $m = m_0+m_\varepsilon$.
Then we have
\begin{align*}
    M_1 = & \max\bigg((T + n^{-1/2}T^{1/2})^{1/2}, \lambda_n^{-\frac{p}{2(4-p)}}\left(\sigma_n^{-d/2}n^{-1/2}m^{\frac{mD}{2m-D}+\frac{1}{2}}(T + n^{-1/2}T^{1/2})^{\frac{1}{2}-\frac{p}{4}}\right)^{\frac{2}{4-p}}, \nonumber\\
    & \sigma_n^{-d/2}n^{-1/2}m^{\frac{mD}{2m-D}+\frac{1}{2}}\lambda_n^{-\frac{p}{4}},  \left(\sigma_n^{-d/2}n^{-1/2}m^{\frac{mD}{2m-D}+\frac{1}{2}}(\lambda_n^{-1}T)^{\frac{p}{2}}(T+n^{-1/2}T^{1/2})^{1-\frac{p}{2}}\right)^{1/2},\nonumber\\
    & (\sigma_n^{-d/2}n^{-1/2}m^{\frac{mD}{2m-D}+\frac{1}{2}})^{\frac{2}{2+p}}(\lambda_n^{-1}T)^{\frac{p}{2(2+p)}}\bigg),\nonumber\\
    M_2 = & \max\bigg((\lambda_n^{-1}(T + n^{-1/2}T^{1/2}))^{1/2}, \left(\lambda_n^{-1}\sigma_n^{-d/2}n^{-1/2}m^{\frac{mD}{2m-D}+\frac{1}{2}}(T + n^{-1/2}T^{1/2})^{\frac{1}{2}-\frac{p}{4}}\right)^{\frac{2}{4-p}}, \nonumber\\
    & \sigma_n^{-d/2}n^{-1/2}m^{\frac{mD}{2m-D}+\frac{1}{2}}\lambda_n^{-\frac{2+p}{4}}, \left(\lambda_n^{-1} \sigma_n^{-d/2}n^{-1/2}m^{\frac{mD}{2m-D}+\frac{1}{2}}(\lambda_n^{-1}T)^{\frac{p}{2}}(T+n^{-1/2}T^{1/2})^{1-\frac{p}{2}}\right)^{1/2},\nonumber\\
    & \lambda_n^{-1/2}(\sigma_n^{-d/2}n^{-1/2}m^{\frac{mD}{2m-D}+\frac{1}{2}})^{\frac{2}{2+p}}(\lambda_n^{-1}T)^{\frac{p}{2(2+p)}}\bigg),
\end{align*}
 Then we have 
\begin{align*}
    \|f^*-\hat f_n\|_n = O_{\PP}(M_1), \|\hat f_n\|_{\cH_{\tilde K_S}(\Omega)} = O_{\PP}(M_2).
\end{align*}
Furthermore, if  $\tilde f_n$ be the solution to the optimization problem
\begin{align}\label{eq:lem2approx22c}
    \min_{f\in \cH_{\tilde K_S}(\Omega)}\|f^*-f\|_n^2 + \lambda_n\|f\|_{\cH_{\tilde K_S}(\Omega)},
\end{align}
then 
\begin{align*}
    \|f^*-\tilde f_n\|_n = O_{\PP}((T+n^{-1/2}T^{1/2})^{1/2}), \|\tilde f_n\|_{\cH_{\tilde K_S}(\Omega)} = O_{\PP}((\lambda_n^{-1}(T + n^{-1/2}T^{1/2}))^{1/2}).
\end{align*}
\end{lemma}

\begin{lemma}[Lemma F.5 of \cite{wang2021inference}]\label{lemmaratio1}
Assume for class $\mathcal{G}$, $\sup_{g\in \mathcal{G}}\|g\|_{L_\infty(\Omega)}\leq c < 1$, and the bracket entropy $H_B(\delta_n,\mathcal{G},\|\cdot\|_{L_2(P_\Xb)})\leq \frac{n\delta_n^2}{1200c^2}$, and $n\delta_n^2\rightarrow \infty$, where $0 < \delta_n < 1$. Then we have
\begin{align*}
P\bigg(\inf_{\|g\|_{L_2(P_\Xb)} \geq 2\delta_n, g\in \mathcal{G}} \frac{\|g\|^2_n}{\|g\|_{L_2(P_\Xb)}^2}<C_3\bigg)\leq C_5\exp(-C_6n\delta_n^2/c^2),
\end{align*}
and
\begin{align*}
P\bigg(\sup_{\|g\|_{L_2(P_\Xb)} \geq 2\delta_n, g\in \mathcal{G}} \frac{\|g\|^2_n}{\|g\|_{L_2(P_\Xb)}^2}>C_4\bigg)\leq C_7\exp(-C_8n\delta_n^2/c^2),
\end{align*}
for some constants $C_3,C_4 > 0$ and $C_i$'s ($i=5,6,7,8$) are only depending on $\Omega$.
\end{lemma}

\begin{lemma}[Interpolation inequality for Polynomial RKHS]\label{lem:ineqPolyRKHS}
Let $g\in\cW^m(\RR^D)$. When $r = \frac{D}{2(m_0+m_\varepsilon)}$ and $D>1$, we have
\begin{align*}
    \|g\|_{L_\infty(\RR^D)}\leq C_9\|g\|_{L_2(\RR^D)}^{1-r}\|g\|_{\cW^m(\RR^D)}^r,
\end{align*}

where $C_9=\left(\int_{\RR^D}(1+\|\bomega\|_2^2)^{-\frac{D}{2}}d\bomega\right)^{\frac{1}{2}}<\infty$.
\end{lemma}

\subsection{Without weight decay}\label{app:pfthmsoboNNnwd}

By the triangle inequality, it can be seen that
\begin{align}\label{eq:thmfsnNNwd1X}
   \|f_t - f^*\|_{L_2(P_\Xb)} 
   \leq & \|f_t - g_t\|_{L_2(P_\Xb)} + \|g_t- f^*\|_{L_2(P_\Xb)},
\end{align}
where $g_t$ is as in \eqref{eq:gdupdnwdg1}. 

By Lemma \ref{lem_gftinf}, the first term $\|f_t - g_t\|_{L_2(P_\Xb)}$ in \eqref{eq:thmfsnNNwd1X} can be bounded by
\begin{align*}
    \|f_t - g_t\|_{L_2(P_\Xb)} \leq C_{10} \|f_t - g_t\|_{L_\infty(\Omega)} = O_\PP\left(\frac{n^2\sqrt{\log N/N}}{\eta_{n}(\tilde \Kb)^2}\right),
\end{align*}
as long as 
\begin{align}\label{eq:lem_gftinf11}
    \frac{1}{2}\eta_n(\tilde \Kb)\geq n\sqrt{\frac{\log N}{N}}.
\end{align}
Choose
\begin{align}\label{eq_N0require}
    N_0 = \frac{4n^2}{\eta_n(\tilde \Kb)^2}.
\end{align}
Then it holds that when $N\geq N_0$,
\begin{align}\label{eq_ftgtc2}
    \|f_t - g_t\|_{L_2(P_\Xb)} = O_\PP\left(n^{-1/2}\right).
\end{align}

It remains to consider $\|g_t- f^*\|_{L_2(P_\Xb)}$ in \eqref{eq:thmfsnNNwd1X}. In order to do so, we consider the empirical version of $\|g_t- f^*\|_{L_2(P_\Xb)}$, and let 
\begin{align}\label{eq:thmfsnNNwd1}
   J_2 = \|g_t- f^*\|_n^2 = \frac{1}{n}\| g_t(\Xb)- f^*(\Xb)\|_2^2.
\end{align}
Let $(\beta t)^{-1} = n\lambda_n$. Consider the kernel ridge regression 
\begin{align}\label{eq:sobogtNN}
    \tilde g = \argmin_{f\in \cH_{\tilde K_S}(\Omega)} \|f - \by\|_n^2 + \lambda_n \| f\|_{\cH_{\tilde K_S}(\Omega)}^2. 
\end{align}
By the representer theorem, $\tilde g(\bx) = \tilde\kb(\bx)^T(\tilde\Kb + n\lambda_n\Ib)^{-1}\by$ for all $\bx\in \Omega$, where $\tilde\kb(\bx) = (\tilde K_S(\bx - \bx_1),...,\tilde K_S(\bx - \bx_n))^T$. Then it can be seen that
\begin{align*}
    \tilde g(\Xb) - f^*(\Xb) =  n\lambda_n(\tilde\Kb + n\lambda_n\Ib)^{-1}f^*(\Xb) + \tilde\Kb(\tilde\Kb + n\lambda_n\Ib)^{-1}\bepsilon = \bq_1 + \bq_2
\end{align*}
Recall that (see \eqref{eq:gdupdnwdg1})
\begin{align*}
    g_t(\Xb) = & \left(\Ib-(\Ib - \beta\tilde\Kb)^t\right)\by,
\end{align*}
which implies 
\begin{align}\label{eq:nwdgerror12c}
    g_t(\Xb) - f^*(\Xb) = -(\Ib - \beta\tilde\Kb)^t f^*(\Xb) + \left(\Ib-(\Ib - \beta\tilde\Kb)^t\right)\bepsilon.
\end{align}

By the Cauchy-Schwarz inequality, \eqref{eq:thmfsnNNwd1}, and \eqref{eq:nwdgerror12c}, it can be seen that
\begin{align}\label{eq:nwdNNgerror1XX}
    nJ_2 \leq & 2(f^*(\Xb))^T(\Ib-\beta \tilde \Kb)^{2t} f^*(\Xb) + 2\bepsilon^T(\Ib - (\Ib-\beta \tilde \Kb)^t)^2\bepsilon\nonumber\\
    = & 2nJ_{21} + 2nJ_{22},
\end{align}
and 
\begin{align}\label{eq:krrNNgerror1Xx}
    n\|\tilde g - f^*\|_n^2 \leq & 2(n\lambda_n)^2(f^*(\Xb))^T(\tilde\Kb + n\lambda_n\Ib)^{-2}f^*(\Xb) + 2\bepsilon^T(\tilde\Kb + n\lambda_n\Ib)^{-1}\tilde\Kb^2(\tilde\Kb + n\lambda_n\Ib)^{-1}\bepsilon\nonumber\\
    = & 2\|\bq_1\|_2^2 + 2\|\bq_2\|_2^2.
\end{align}
Similar to \eqref{eq:complemj11j21}, it can be seen that
\begin{align}\label{eq:nwdNNgerror1J21}
    2nJ_{21}\leq C_{11}\|\bq_1\|_2^2, 
\end{align}
and similar to \eqref{eq:complemj12j22}, the term $2nJ_{22}$ can be further bounded by 
\begin{align}\label{eq:nwdNNgerror1J22}
    2nJ_{22} = & 2\sum_{j=1}^n (1-(1-\beta\eta_j)^t)^2(\bv_j^T\bepsilon)^2\leq 2\sum_{j=1}^n \frac{4(\beta t \eta_j)^{2}}{(1+\beta t \eta_j)^2}(\bv_j^T\bepsilon)^2\nonumber\\
    % = & 2\sum_{j=1}^n \frac{4(n \lambda_j)^{2}}{(n(\beta t)^{-1}+ n \lambda_j)^2}(\bv_j^T\bepsilon)^2\nonumber\\
    = & 8\bepsilon^T(\tilde\Kb + (\beta t)^{-1}\Ib)^{-1}\tilde\Kb^2(\tilde\Kb + (\beta t)^{-1}\Ib)^{-1}\bepsilon = 8\|\bq_2\|_2^2,
\end{align}
where $\eta_1\geq \ldots\geq \eta_n>0$ and $\bv_j$, $j=1,\ldots,n$ be the eigenvalues and corresponding eigenvectors of $\tilde\Kb$, respectively. In the last inequality of \eqref{eq:nwdNNgerror1J22}, we note $(\beta t)^{-1}=n\lambda_n$.

Plugging \eqref{eq:nwdNNgerror1J21} and \eqref{eq:nwdNNgerror1J22} into \eqref{eq:nwdNNgerror1XX}, we obtain
\begin{align}\label{eq:pfsobonowdNNJ2}
    J_2 \leq \frac{2C_{12}}{n}\left(\|\bq_1\|_2^2 + \|\bq_2\|_2^2\right).
\end{align}
The term $\|\bq_1\|_2^2$ and $\|\bq_2\|_2^2$ can be directly bounded by Lemma \ref{lem:approx2NN}. To see this, let $f_0(\bx) = 0$ for all $\bx\in \Omega$. Then it can be checked that
\begin{align*}
    \frac{1}{n}\|\bq_1\|_2^2 = & \|\tilde f_n - f\|_n^2
\end{align*}
and 
\begin{align*}
    \frac{1}{n}\|\bq_2\|_2^2 = & \|\hat f_{0,n} - f_0\|_n^2,
\end{align*}
where $\tilde f_n$ is as in \eqref{eq:lem2approx22c}, and $\hat f_{0,n}$ is the solution to the optimization problem
\begin{align*}%\label{eq:lem2approx2NN}
    \min_{g\in \cH_{\tilde K_S}(\Omega)}\|\bepsilon-g\|_n^2 + \lambda_n\|g\|_{\cH_{\tilde K_S}(\Omega)}^2.
\end{align*}

Let $\delta_0 \in (0,1)$ such that $ 4m_\varepsilon D-(2m_0+2(1-\delta_0)m_\varepsilon -D)d> 0$. Take 
\begin{align*}
    \lambda_n \asymp  n^{-\frac{2(m_0+m_\varepsilon)}{2m_f+d}}\sigma_n^{-2m_\varepsilon},
    \sigma_n \asymp  n^{-\frac{2(2m_0+2m_\varepsilon)D-(2m_0+2m_\varepsilon-D)d}{(2m_f+d)(4m_\varepsilon D-(2m_0+2(1-\delta_0)m_\varepsilon -D)d)}},
    n^{-1}(\beta t)^{-1} \asymp  \lambda_n,
    \beta \asymp n^{-1}.
\end{align*}
Therefore, if $m_\varepsilon = O((\log n)^C)$ for some constant $C$, and
\begin{align}\label{eq:mainthmnn1111}
    & \lambda_n \leq C_{13}(\lambda_n(m_\varepsilon + 1)^{m_\varepsilon}\sigma_n^{2m_\varepsilon})^{\frac{m_f}{m_0+m_\varepsilon}}\nonumber\\
    \Leftrightarrow & n^{-\frac{2(m_0+m_\varepsilon)}{2m_f+d}}n^{\frac{4m_\varepsilon(2m_0+2m_\varepsilon)D-2m_\varepsilon(2m_0+2m_\varepsilon-D)d}{(2m_f+d)(4m_\varepsilon D-(2m_0+2(1-\delta_0)m_\varepsilon -D)d)}} 
    \leq C_{14}n^{-\frac{2m_f}{2m_f+d}}(m_\varepsilon + 1)^{\frac{m_\varepsilon m_f}{m_0+m_\varepsilon}}\nonumber\\
    % \Leftrightarrow & -\frac{2(m_0+m_\varepsilon)}{2m_f+d}-\frac{2m_\varepsilon(m_0+m_\varepsilon)(D-d)}{(2m_f+d)(m_\varepsilon D-m_0 d -(1-\delta_0) m_\varepsilon d)}\leq -\frac{2(m_0+m_\varepsilon)}{2m_f+d}\nonumber\\
    \Leftarrow & m_\varepsilon^2 \delta_0d > m_\varepsilon(2m_f D + (m_0-m_f)(1-\delta_0)d) \nonumber\\
    \Leftarrow & m_\varepsilon > \frac{2m_fD+m_0d}{\delta_0d},
\end{align}
when $m_0\leq m_f$, or
\begin{align}\label{eq:mainthmnn1112}
    & \lambda_n^{\frac{m_f}{m_0}} \leq C_{15}(\lambda_n(m_\varepsilon + 1)^{m_\varepsilon}\sigma_n^{2m_\varepsilon})^{\frac{m_f}{m_0+m_\varepsilon}}\nonumber\\
    \Leftrightarrow & n^{-\frac{2(m_0+m_\varepsilon)}{2m_f+d}}n^{\frac{4m_\varepsilon(2m_0+2m_\varepsilon)D-2m_\varepsilon(2m_0+2m_\varepsilon-D)d}{(2m_f+d)(4m_\varepsilon D-(2m_0+2(1-\delta_0)m_\varepsilon -D)d)}} 
    \leq C_{16}n^{-\frac{2m_0}{2m_f+d}}(m_\varepsilon + 1)^{\frac{m_\varepsilon m_0}{m_0+m_\varepsilon}}\nonumber\\
    % \Leftrightarrow & -\frac{2(m_0+m_\varepsilon)}{2m_f+d}-\frac{2m_\varepsilon(m_0+m_\varepsilon)(D-d)}{(2m_f+d)(m_\varepsilon D-m_0 d -(1-\delta_0) m_\varepsilon d)}\leq -\frac{2(m_0+m_\varepsilon)}{2m_f+d}\nonumber\\
    \Leftarrow & m_\varepsilon^2 \delta_0d > 2m_0 m_\varepsilon D \nonumber\\
    \Leftarrow & m_\varepsilon > \frac{2m_0D+m_0d}{\delta_0d},
\end{align}
when $m_0 > m_f$,
we have
\begin{align*}
    T \leq C_{17}n^{-\frac{2m_f}{2m_f+d}}(m_\varepsilon + 1)^{\frac{m_\varepsilon m_f}{m_0+m_\varepsilon}}\leq C_{17}n^{-\frac{2m_f}{2m_f+d}}(m_\varepsilon + 1)^{m_f},
\end{align*}
where $T$ is as in Lemma \ref{lem:approx2NN}. Suppose $D>1$, long but tedious calculation shows that 
\begin{align*}
    M_1 \leq C_{18}(m_\varepsilon + m_0)^{m_f+\frac{1}{2}} n^{-\frac{m_f}{2m_f+d}+\delta'},
\end{align*}
where $M_1$ is as in Lemma \ref{lem:approx2NN}, and 
\begin{align*}
    % \delta' = & \frac{\delta_1}{2}\\
    \delta' = & \frac{((4m_0+4m_\varepsilon)D-(2m_0+2m_\varepsilon-D)d)m_\varepsilon d}{(2m_f+d)(2m_\varepsilon+2m_0-D)(4m_\varepsilon D-(2m_0+2(1-\delta_0)m_\varepsilon-D)d)}\delta_0 
    % \leq \frac{m_\varepsilon d}{2(m_\varepsilon (D-d) +m_0 D)}\delta_0\nonumber\\
    \leq \frac{d}{2(2m_f+d)}\delta_0,\\
    % \Leftarrow & m_\varepsilon > \frac{(2m_0-D)d+2D^2}{2\delta_0 d}
\end{align*}
where the inequality is because of \eqref{eq:mainthmnn1111} (if $m_0\leq m_f$) or \eqref{eq:mainthmnn1112} (if $m_0> m_f$).
Therefore, by taking $\delta_0 = d^{-1}(2m_f+d)a$ and $m_\varepsilon = (\delta_0 d)^{-1}(2D \max(m_0,m_f)+ m_0d) + 1$, we have 
\begin{align}\label{eq:pfsobonowdq1q2_a}
    \frac{1}{n}\|\bq_1\|_2^2 = & \|\tilde f_n - f\|_n^2 = O_{\PP}\left(n^{-\frac{2m_f}{2m_f + d}+a}\right),\nonumber\\
    \frac{1}{n}\|\bq_2\|_2^2 = & \|\hat f_{0,n} - f_0\|_n^2  = O_{\PP}\left(n^{-\frac{2m_f}{2m_f + d}+a}\right).
\end{align}
Then by \eqref{eq:pfsobonowdNNJ2} and \eqref{eq:pfsobonowdq1q2_a}, we obtain
\begin{align}\label{eq:pfsobonowdNNJ2Xa}
    J_2 = O_{\PP}\left(n^{-\frac{2m_f}{2m_f + d}+a}\right),
\end{align}
which corresponds to the first statement of Theorem \ref{thm:soboNN}.

Taking $\delta_0 = (\log n)^{-1}$, we obtain that 
\begin{align*}
     M_1 \leq C_{18} n^{-\frac{m_f}{2m_f+d}}e^{\frac{d}{2(2m_f+d)}}(m_\varepsilon + m_0)^{m_f+\frac{1}{2}}\leq C_{19} n^{-\frac{m_f}{2m_f+d}}(m_\varepsilon + m_0)^{m_f+\frac{1}{2}},
\end{align*}
where we require $m_\varepsilon > d^{-1}(2D \max(m_0,m_f)+ m_0d)\log n$. Thus, we can directly take $m_\varepsilon = 2d^{-1}(2D \max(m_0,m_f)+ m_0d)\log n - m_0$ such that
\begin{align}\label{eq:pfsobonowdq1q2}
    \frac{1}{n}\|\bq_1\|_2^2 = & \|\tilde f_n - f\|_n^2 = O_{\PP}\left(n^{-\frac{2m_f}{2m_f + d}}(\log n)^{2m_f+1}\right),\nonumber\\
    \frac{1}{n}\|\bq_2\|_2^2 = & \|\hat f_{0,n} - f_0\|_n^2  = O_{\PP}\left(n^{-\frac{2m_f}{2m_f + d}}(\log n)^{2m_f+1}\right).
\end{align}

Thus, by \eqref{eq:pfsobonowdNNJ2} and \eqref{eq:pfsobonowdq1q2}, we have
\begin{align}\label{eq:pfsobonowdNNJ2X}
    J_2 = O_{\PP}\left(n^{-\frac{2m_f}{2m_f + d}}(\log n)^{2m_f+1}\right),
\end{align}
which corresponds to the second statement of Theorem \ref{thm:soboNN}.

% \textbf{l;asjdfl;kasjkl;asdj;lkadsjfdklj}

% Plugging \eqref{eq:pfsobonowdNNJ1} and \eqref{eq:pfsobonowdNNJ2X} into \eqref{eq:thmfsnNNwd1}, we have 
% \begin{align}\label{eq:thmfsnwf1}
%   \|f_t - f^*\|_n^2 = O_{\PP}\left(n^{-\frac{2m_f}{2m_f + d}} + nN^{-2} + n^3\beta^2t^2N^{-4}\right) = O_{\PP}\left(n^{-\frac{2m_f}{2m_f + d}}\right),
% \end{align}
% as long as $nN^{-2} + n^{\frac{4m_0+6m_f}{2m_f+D}}N^{-4}\lesssim n^{-\frac{2m_f}{2m_f + d}}$, which is satisfied for $N\gtrsim n^{\frac{2m_f+m_0}{2m_f+d}}$.
It remains to bound $\|g_t - f^*\|_{L_2(P_\Xb)}$. Note that
\begin{align*}
    \|g_t - f^*\|_{L_2(P_\Xb)} \leq \|g_t - f_n^*\|_{L_2(P_\Xb)} + \|f_n - f^*\|_{L_2(P_\Xb)} \leq \|g_t - f_n^*\|_{L_2(P_\Xb)} + T^{1/2},
\end{align*}
and
\begin{align*}
    \|g_t - f_n^*\|_n \leq & \|g_t - f^*\|_n + \|f_n^* - f^*\|_n \leq \|g_t - f^*\|_n  + O_{\PP}\left(\left(T + n^{-1/2}T^{1/2}\right)^{1/2}\right)\nonumber\\
    \leq & O_{\PP}\left(n^{-\frac{2m_f}{2m_f + d}}(\log n)^{2m_f+1}\right),
\end{align*}
where the second inequality is because of \eqref{eq:lemapprx2NNfn1}. Therefore, it suffices to bound the difference between $\|g_t - f_n^*\|_{L_2(P_\Xb)}$ and $\|g_t - f_n^*\|_n$.

By \eqref{eq:gtRKHSnormpf} and Lemma \ref{lem:approx2NN}, we have
\begin{align}\label{eq:pfsobonowdNNgpsi}
   \|g_t\|_{\cN_{\sigma}(\Omega)}^2\leq \sigma_n^{-2m_0} \|g_t\|_{\cH_{\tilde K_S}(\Omega)}^2\leq C_{20}\sigma_n^{-2m_0}\|\tilde g\|_{\cH_{\tilde K_S}(\Omega)}^2= O_{\PP}\left(n^{\nu_1}(\log n)^{2m_f+1}\right),
\end{align}
where 
\begin{align}\label{eq:pfNNnu1cond}
    \nu_1 = & \frac{2(m_0+m_\varepsilon -m_f)}{2m_f+d} + 2(m_\varepsilon-m_0)\nu,\nonumber\\
    \mbox{and } \nu = & -\frac{2(2m_0+2m_\varepsilon)D-(2m_0+2m_\varepsilon-D)d}{(2m_f+d)(4m_\varepsilon D-(2m_0+2(1-\delta_0)m_\varepsilon -D)d)}.
\end{align}
% \ww{asdfl;jask;djajsfd;lkfadsjl;k}

Consider function class 
$\cG = \{h: h = (g_t-f_n^*)/(C_{21}n^{\nu_1/2}(\log n)^{m_f+1/2})\}$, where the constant $C_{21}$ is taken such that $\|h_1\|_{\cN_{\sigma}(\Omega)}<1$ for all $h_1\in \cG$. Then lemma \ref{lem:ineqPolyRKHS} leads to
\begin{align*}
    \|h_1\|_{L_\infty(\Omega)}\leq C_{22}\|h_1\|_{L_2(P_\Xb)}^{1-\frac{D}{2(m_0+m_\varepsilon)}}\|h_1\|_{\cN_{\sigma}(\Omega)}^{\frac{D}{2(m_0+m_\varepsilon)}},
\end{align*}
for all $h_1\in \cG$, which implies
\begin{align*}
    c_1:= \sup_{h_1\in \cG}\|h_1\|_{L_\infty(\Omega)} \leq C_{22}R_1^{1-\frac{D}{2(m_0+m_\varepsilon)}},
\end{align*}
where $R_1 = \sup_{h_1\in \cG}\|h_1\|_{L_2(P_\Xb)} \leq \sup_{h_1\in \cG}\|h_1\|_{L_\infty(\Omega)}\leq  \sup_{h_1\in \cG}\|h_1\|_{\cN_{\sigma}(\Omega)}<1$, because of the reproducing property. Let $m=m_0+m_\varepsilon$. Taking $c= C_{22} R_1^{1-\frac{D}{2m}}<1$, and $\delta_n= C_{23}(\sigma_n^{-d}n^{-1}c^2m^{\frac{2mD}{2m-D}})^{\frac{2m-D}{4m}}$ in Lemma \ref{lemmaratio1}, it can be checked that
\begin{align*}
    C_{24}n\delta_n^2c^{-2} \geq H(\delta,\cB_{\cH_\sigma(\Omega)},\|\cdot\|_{L_\infty(\Omega)}),
\end{align*}
which implies the conditions of Lemma \ref{lemmaratio1} are fulfilled. Applying Lemma \ref{lemmaratio1} to the case $\|g_t - f_n^*\|_{L_2(P_\Xb)}^2\geq \delta_n^2n^{\nu_1}$, together with \eqref{eq:pfsobonowdNNJ2X}, we have
\begin{align}\label{eq:thmfsnwNNf2}
   R_1 = O_{\PP}\left(\max\{n^{-\frac{m_f}{2m_f + d}-\nu_1/2}(\log n)^{m_f+1/2}, \delta_n\}\right).
\end{align}
If $\delta_n \geq n^{-\frac{m_f}{2m_f + d}-\nu_1/2}(\log n)^{m_f+1/2}$, we have $R_1\leq C_{25}\delta_n$,
which implies
\begin{align*}
    R_1 \leq C_{26} (\sigma_n^{-d}n^{-1}c^2m^{\frac{2mD}{2m-D}})^{\frac{2m-D}{4m}}.
\end{align*}
Therefore, we have
\begin{align*}
    \|g_t - f_n^*\|_{L_2(P_\Xb)}\leq C_{21} n^{\nu_1/2}R_1\leq C_{27}n^{\nu_2}(\log n)^{D/2},
\end{align*}
where
\begin{align*}
    \nu_2 = & \frac{(m_0+m_\varepsilon -m_f)}{2m_f+d} + (m_\varepsilon-m_0)\nu-\frac{2m-D}{4m}(d\nu+1) < -\frac{m_f}{2m_f+d}.
\end{align*}
% Therefore, if
% \begin{align*}
    
% \end{align*}
% we have
% \begin{align*}
%     \|g_t - f^*\|_{L_2(P_\Xb)}= O_{\PP}(n^{-\frac{m_f}{2m_f + d}}).
% \end{align*}
% where the inequality holds because of \eqref{eq:mainthmnn1111} (if $m_0\leq m_f$) or \eqref{eq:mainthmnn1112} (if $m_0> m_f$).

If $\delta_n < n^{-\frac{m_f}{2m_f + d}-\nu_1/2}(\log n)^{m_f+1/2}$, then $R_1=O_{\PP}(n^{-\frac{m_f}{2m_f + d}-\nu_1/2}(\log n)^{m_f+1/2})$, which implies 
$\|g_t - f_n^*\|_{L_2(P_\Xb)} = O_{\PP}(n^{-\frac{m_f}{2m_f + d}}(\log n)^{m_f+1/2})$. Here we note that the proof is still valid if we replace $g_t$ with $\tilde g$. Therefore, in both cases we have $\|g_t - f_n^*\|_{L_2(P_\Xb)} = O_{\PP}(n^{-\frac{m_f}{2m_f + d}}(\log n)^{m_f+1/2})$, which, together with \eqref{eq:nwdNNgerror1J22} and \eqref{eq_ftgtc2}, finishes the proof.
\hfill\BlackBox

\subsection{With weight decay}\label{app:pfthmsoboNNwd}

If $\alpha>0$, we decompose the error by
\begin{align}\label{eq:thmfsNNwd1X}
   \|f_t - f^*\|_{L_2(P_\Xb)} 
   \leq & \|f_t - g_t\|_{L_2(P_\Xb)} + \|\tilde \kb(\cdot)^T(\alpha/\beta \Ib + \tilde \Kb)^{-1}\by - f^* \|_{L_2(P_\Xb)} \nonumber\\
   & + \|\beta\kb(\cdot)^T((1-\alpha)\Ib - \beta\tilde\Kb)^t(\alpha\Ib + \beta\tilde\Kb)^{-1}\by\|_{L_2(P_\Xb)}\nonumber\\
   = & I_1 + I_2 + I_3.
\end{align}
As in \eqref{eq_ftgtc2}, there exists an $N_0$ (depending on $n$) such that when $N\geq N_0$,
\begin{align}\label{eq:pfsobowdNNI1}
    I_1 = O_\PP\left(n^{-1/2}\right).
\end{align}

The second term is the error $\|\tilde f_n - f^*\|_{L_2(P_\Xb)}$, where $\tilde f_n$ is as in \eqref{eq:lem2approx22c}. Lemma \ref{lem:approx2NN} gives us that 
\begin{align*}
    \|\tilde f_n - f^*\|_n= O_{\PP}(n^{-\frac{m_f}{2m_f+d}}).
\end{align*}
Following a similar approach in Appendix \ref{app:pfthmsoboNNnwd}, it can be further shown that 
\begin{align}\label{eq:pfsobowdNNI2}
    I_2 = O_{\PP}(n^{-\frac{m_f}{2m_f+d}}),
\end{align}
where we let $\alpha \asymp n^{-1-\frac{2(m_0+m_\varepsilon)}{2m_f+d}}\sigma_n^{-2m_\varepsilon}$, and $\beta$ and $\sigma_n$ are as in Theorem \ref{thm:soboNN}.

It remains to bound $I_3$ in \eqref{eq:thmfsNNwd1X}. By Cauchy-Schwarz inequality, 
\begin{align}\label{eq:pfsobowdNNI3}
    & \|\beta\kb(\cdot)^T((1-\alpha)\Ib - \beta\tilde\Kb)^t(\alpha\Ib + \beta\tilde\Kb)^{-1}\by\|_{L_2(P_\Xb)} \nonumber\\
    \leq & \left\|\left({\rm tr}\left(\left((\alpha/\beta\Ib + \tilde\Kb)^{-1}\by\kb(\cdot)^T\right)^2\right){\rm tr}\left(((1-\alpha)\Ib - \beta\tilde\Kb)^{2t}\right)\right)^{1/2}\right\|_{L_2(P_\Xb)}\nonumber\\
    \leq & \left\|\kb(\cdot)^T(\alpha/\beta\Ib + \tilde\Kb)^{-1}\by\right\|_{L_2(P_\Xb)}\left({\rm tr}\left(((1-\alpha)\Ib - \beta\tilde\Kb)^{2t})\right)\right)^{1/2}\nonumber\\
    \leq & \|(\kb(\cdot)^T\kb(\cdot))^{1/2}\|_{L_2(P_\Xb)}\|\by\|_2\beta/\alpha\nonumber\\
    = & O_{\PP}\left(n^{1+\frac{2(m_0+m_\varepsilon)}{2m_f+d}}\sigma_n^{2m_\varepsilon}(1-\alpha)^{t}\right).
\end{align}
Thus, there exists $t_0>0$ such that as long as $t>t_0$, $I_2$ dominates $I_3$. Combining \eqref{eq:pfsobowdNNI1}, \eqref{eq:pfsobowdNNI2}, and \eqref{eq:pfsobowdNNI3}, we finish the proof.
\hfill\BlackBox

% Is $g_t = k(x,X)K^{-1}((\Ib-\beta\tilde \Kb)^t)\by$?

% In this case, the error can be decomposed by
% \begin{align}
%     & \|f_{t+1}(\Xb) - f^*(\Xb)\|_2\leq \|f_{t+1}(\Xb) -  g_{t+1}(\Xb)\|_2 + \|g_{t+1}(\Xb) - f^*(\Xb)\|_2\nonumber\\
%     \leq & \|f_{t+1}(\Xb) - g_{t+1}(\Xb)\|_2 + \|\tilde \Kb(\alpha/\beta \Ib + \tilde \Kb)^{-1}\by - f^*(\Xb)\|_2 + \|((1-\alpha)\Ib - \beta\tilde \Kb)^{t+1}\beta(\alpha\Ib + \beta\tilde \Kb)^{-1}\tilde \Kb\by\|_2\nonumber\\
%     = & I_1 + I_2 + I_3,
% \end{align}
% where 
% \begin{align*}
%     g_{t+1}(\Xb) = \tilde \Kb(\alpha/\beta \Ib + \tilde \Kb)^{-1}\by -((1-\alpha)\Ib - \beta\tilde \Kb)^{t+1}\beta(\alpha\Ib + \beta\tilde \Kb)^{-1}\tilde \Kb\by.
% \end{align*}

% ...

\section{Proof of Theorem \ref{thm:soboGN}}\label{app:pfthmsoboGN}
We first present some lemmas, whose proofs can be found in Appendix \ref{app:pfofGaussianThm}.

\begin{lemma}\label{lem:GPembedding}
Let $k_\sigma(\bx-\bx')$ be a Gaussian kernel defined by
\begin{align}\label{eq:gaussK}
    k_\sigma(\bx-\bx') = \exp\left(-\frac{\|\bx-\bx'\|_2^2}{4\sigma^2}\right),
\end{align}
and $\cH_\sigma(\RR^D)$ be the RKHS generated by $k_\sigma(\bx-\bx')$. Then we have
\begin{align}\label{eq:appGauEb1}
   \|h_1\|_{\cH_{\sigma_n/\sqrt{2}}(\RR^D)} \leq C_1\sigma_n^{-D/2}\|h_1\|_{\cH_{\tilde K_S}(\RR^D)}
\end{align}
and
\begin{align}\label{eq:appGauEb2}
    \|h_2\|_{\cH_{\tilde K_S}(\RR^D)}\leq C_2\sigma_n^{-m_0-D/2}\|h_2\|_{\cH_{\sqrt{3}\sigma_n}(\RR^D)},
\end{align}
for $h_1\in \cH_{\tilde K_S}(\RR^D)$ and $h_2\in \cH_{\sqrt{3}\sigma_n}(\RR^D)$, where $C_1$ and $C_2$ does not depend on $\sigma_n$. 
\end{lemma}

\begin{lemma}\label{lem:approx1}
Let $f_n^*$ be the solution to the optimization problem
\begin{align}
    \min_{g\in \cH_{\tilde K_S}(\Omega)}\|f^*-g\|_{L_2(P_\Xb)}^2 + \lambda_n\|g\|_{\cH_{\tilde K_S}(\Omega)}^2.
\end{align}
Then 
\begin{align*}
    \|f^*-f_n^*\|_{L_2(P_\Xb)}^2 \leq C_3\max(\lambda_n\sigma_n^{-2m_0},\sigma_n^{2m_f}),
\end{align*}
and 
\begin{align*}
    \|f_n^*\|_{\cH_{\tilde K_S}(\Omega)}^2 \leq C_3\lambda_n^{-1}\max(\lambda_n\sigma_n^{-2m_0},\sigma_n^{2m_f}).
\end{align*}
\end{lemma}

\begin{lemma}\label{lem:approx2}
Let $f_n^*$ be the solution to the optimization problem
\begin{align}
    \min_{g\in \cH_{\tilde K_S}(\Omega)}\|f^*-g\|_{L_2(P_\Xb)}^2 + \lambda_n\|g\|_{\cH_{\tilde K_S}(\Omega)}^2.
\end{align}
Suppose there exists $T>0$ (depending on $n$) such that
\begin{align*}
    \|f^*-f_n^*\|_{L_2(P_\Xb)}^2 + \lambda_n\|f_n^*\|_{\cH_{\tilde K_S}(\Omega)}^2\leq T.
\end{align*}
Let $\hat f_n$ be the solution to the optimization problem
\begin{align}\label{eq:lem2approx2}
    \min_{g\in \cH_{\tilde K_S}(\Omega)}\|\by-g\|_n^2 + \lambda_n\|g\|_{\cH_{\tilde K_S}(\Omega)}^2.
\end{align}
Let $p=(\log n)^{-1}$,
\begin{align*}
    M_1 = & \max\bigg((T + n^{-1/2}T^{1/2})^{1/2}, \sigma_n^{-d/2 - \frac{pD}{4}}p^{-(D+1)/2}n^{-1/2}\lambda_n^{-\frac{p}{4}}, \nonumber\\
    & \lambda_n^{-\frac{p}{2(4-p)}}\left(\sigma_n^{-d/2 - \frac{pD}{4}}p^{-(D+1)/2}n^{-1/2}(T + n^{-1/2}T^{1/2})^{\frac{1}{2}-\frac{p}{4}}\right)^{\frac{2}{4-p}}, \nonumber\\
    & \left(\sigma_n^{-d/2 - \frac{pD}{4}}p^{-(D+1)/2}n^{-1/2}(\lambda_n^{-1}T)^{\frac{p}{2}}(T+n^{-1/2}T^{1/2})^{1-\frac{p}{2}}\right)^{1/2},\nonumber\\
    & (\sigma_n^{-d/2 - \frac{pD}{4}}p^{-(D+1)/2}n^{-1/2})^{\frac{2}{2+p}}(\lambda_n^{-1}T)^{\frac{p}{2+p}}\bigg),\nonumber\\
    M_2 = & \max\bigg((\lambda_n^{-1}(T + n^{-1/2}T^{1/2}))^{1/2}, \sigma_n^{-d/2 - \frac{pD}{4}}p^{-(D+1)/2}n^{-1/2}\lambda_n^{-\frac{2+p}{4}}, \nonumber\\
    & \left(\lambda_n^{-1} \sigma_n^{-d/2 - \frac{pD}{4}}p^{-(D+1)/2}n^{-1/2}(T + n^{-1/2}T^{1/2})^{\frac{1}{2}-\frac{p}{4}}\right)^{\frac{2}{4-p}}, \nonumber\\
    & \left(\lambda_n^{-1} \sigma_n^{-d/2 - \frac{pD}{4}}p^{-(D+1)/2}n^{-1/2}(\lambda_n^{-1}T)^{\frac{p}{2}}(T+n^{-1/2}T^{1/2})^{1-\frac{p}{2}}\right)^{1/2},\nonumber\\
    & \lambda_n^{-1/2}(\sigma_n^{-d/2 - \frac{pD}{4}}p^{-(D+1)/2}n^{-1/2})^{\frac{2}{2+p}}(\lambda_n^{-1}T)^{\frac{p}{2+p}}\bigg).
\end{align*}
Then we have 
\begin{align*}
    \|f^*-\hat f_n\|_n = O_{\PP}(M_1), \|\hat f_n\|_{\cH_{\tilde K_S}(\Omega)} = O_{\PP}(M_2).
\end{align*}
Furthermore, if  $\tilde f_n$ be the solution to the optimization problem
\begin{align}\label{eq:lem2approx22}
    \min_{f\in \cH_{\tilde K_S}(\Omega)}\|f^*-f\|_n^2 + \lambda_n\|f\|_{\cH_{\tilde K_S}(\Omega)},
\end{align}
then 
\begin{align*}
    \|f^*-\tilde f_n\|_n = O_{\PP}((T+n^{-1/2}T^{1/2})^{1/2}), \|\tilde f_n\|_{\cH_{\tilde K_S}(\Omega)} = O_{\PP}((\lambda_n^{-1}(T + n^{-1/2}T^{1/2}))^{1/2}).
\end{align*}
% Let $\lambda_n \asymp n^{-\frac{2m_f-2m_0}{2m_f+d}}$ and $\sigma_n\asymp n^{-\frac{1}{2m_f + d}}$, we have 
% \begin{align}\label{eq:lemapp1fubXX}
%     \|f-\hat f\|_{L_2(P_\Xb)}^2 = O_{\PP}\left(n^{-\frac{2m_f}{2m_f+d}}\right),\quad\|\hat f\|_{\cH_{\tilde K_S}(\Omega)}^2 = O_{\PP}\left(n^{-\frac{2m_0}{2m_f+d}}\right).
% \end{align}
% It remains to bound the difference between $\|f-\hat f\|_n^2$ and $\|f-\hat f\|_{L_2(P_\Xb)}^2$.
% Then 
% \begin{align*}
%     \|f-\hat f_n\|_{L_2(\RR^D)}^2 =O_{\PP}(\max(\lambda_n\sigma_n^{-2m_0},\sigma_n^{2m_f}))
% \end{align*}
% and 
% \begin{align*}
%     \|\hat f_n\|_{\cH_{\tilde K_S}(\RR^D)}^2=O_{\PP}( \lambda_n^{-1}\max(\lambda_n\sigma_n^{-2m_0},\sigma_n^{2m_f})\|f^*\|_{\cW^{m_f}(\Omega)}^2).
% \end{align*}
\end{lemma}

\begin{lemma}[Interpolation inequality for Gaussian RKHS]\label{lem:ineqGRKHS}
Let $g\in\cH_\sigma(\RR^D)$. For any $1>r>0$, we have
\begin{align*}
    \|g\|_{L_\infty(\RR^D)}\leq C_4r^{-\frac{D}{4}}\sigma^{\frac{D(r-1)}{2}}\|g\|_{L_2(\RR^D)}^{1-r}\|g\|_{\cH_\sigma(\RR^D)}^r,
\end{align*}
where $C_4$ is a constant not related to $r,\sigma$ and $g$.

\end{lemma}

\subsection{Without weight decay}\label{app:pfthmsoboGNnwd}

We first decompose the error as
\begin{align}\label{eq:thmfsnGNwd1X}
   \|f_t - f^*\|_{L_2(P_\Xb)} 
   \leq & \|f_t - g_t\|_{L_2(P_\Xb)} + \|g_t- f^*\|_{L_2(P_\Xb)},
\end{align}
where $g_t$ is as in \eqref{eq:gdupdnwdg1}. 

By Lemma \ref{lem_gftinf}, the first term $\|f_t - g_t\|_{L_2(P_\Xb)}$ in \eqref{eq:thmfsnGNwd1X} can be bounded by
\begin{align*}
    \|f_t - g_t\|_{L_2(P_\Xb)} \leq C_5 \|f_t - g_t\|_{L_\infty(\Omega)} = O_\PP\left(\frac{n^2\sqrt{\log N/N}}{\eta_{n}(\tilde \Kb)^2}\right),
\end{align*}
as long as 
\begin{align}\label{eq:lem_gftinf11Gnn}
    \frac{1}{2}\eta_n(\tilde \Kb)\geq n\sqrt{\frac{\log N}{N}}.
\end{align}
Choose
\begin{align}\label{eq_N0requireG}
    N_0 = \frac{4n^2}{\eta_n(\tilde \Kb)^2}.
\end{align}
Then it holds that when $N\geq N_0$,
\begin{align}\label{eq_ftgtc2Gnn}
    \|f_t - g_t\|_{L_2(P_\Xb)} = O_\PP\left(n^{-1/2}\right).
\end{align}

It remains to consider $\|g_t- f^*\|_{L_2(P_\Xb)}$. We consider the empirical version of $\|g_t- f^*\|_{L_2(P_\Xb)}$, and let 
\begin{align}\label{eq:thmfsnGNwd1}
   J_2 = \|g_t- f^*\|_n^2 = \frac{1}{n}\| g_t(\Xb)- f^*(\Xb)\|_2^2.
\end{align}
Let $(\beta t)^{-1} = n\lambda_n$. Consider the kernel ridge regression 
\begin{align*}
    \tilde g = \argmin_{f\in \cH_{\tilde K_S}(\Omega)} \|f - \by\|_n^2 + \lambda_n \| f\|_{\cH_{\tilde K_S}(\Omega)}^2. 
\end{align*}
By the representer theorem, $\tilde g(\bx) = \tilde\kb(\bx)^T(\tilde\Kb + n\lambda_n\Ib)^{-1}\by$ for all $\bx\in \Omega$. Then it can be seen that
\begin{align*}
    \tilde g(\Xb) - f^*(\Xb) =  n\lambda_n(\tilde\Kb + n\lambda_n\Ib)^{-1}f^*(\Xb) + \tilde\Kb(\tilde\Kb + n\lambda_n\Ib)^{-1}\bepsilon = \bq_1 + \bq_2
\end{align*}
Following the arguments in Appendix \ref{app:pfthmsoboNNnwd}, the term $J_2$ can be bounded by
\begin{align}\label{eq:pfsobonowdGNJ2}
    J_2 \leq \frac{2}{n}\left( 2C_6 \|\bq_1\|_2^2 + 8\|\bq_2\|_2^2\right),
\end{align}
and 
\begin{align*}
    \frac{1}{n}\|\bq_1\|_2^2 = & \|\tilde f_n - f^*\|_n^2
\end{align*}
and 
\begin{align*}
    \frac{1}{n}\|\bq_2\|_2^2 = & \|\hat f_{0,n} - f_0\|_n^2,
\end{align*}
where $f_0(\bx) = 0$ for all $\bx\in \Omega$, $\tilde f_n$ is as in \eqref{eq:lem2approx22}, and $\hat f_{0,n}$ is the solution to the optimization problem
\begin{align*}%\label{eq:lem2approx2NN}
    \min_{g\in \cH_{\tilde K_S}(\Omega)}\|\bepsilon-g\|_n^2 + \lambda_n\|g\|_{\cH_{\tilde K_S}(\Omega)}^2.
\end{align*}
By setting $\beta t \asymp n^{\frac{2m_0-d}{2m_f+d}}$ (which implies $\lambda_n \asymp n^{-\frac{2m_0+2m_f}{2m_f+d}}$), $\sigma_n\asymp n^{-\frac{1}{2m_f + d}}$, Lemma \ref{lem:approx1} implies that $T\asymp n^{-\frac{2m_f}{2m_f+d}}$, which, together with Lemma \ref{lem:approx2}, implies
\begin{align}\label{eq:pfsobonowdGNq1q2}
    \frac{1}{n}\|\bq_1\|_2^2 = & \|\tilde f_n - f^*\|_n^2 = O_{\PP}\left(n^{-\frac{2m_f}{2m_f + d}}(\log n)^{D+1}\right),\nonumber\\
    \frac{1}{n}\|\bq_2\|_2^2 = & \|\hat f_{0,n} - f_0\|_n^2  = O_{\PP}\left(n^{-\frac{2m_f}{2m_f + d}}(\log n)^{D+1}\right).
    % \frac{1}{n}\|\bq_2\|_2^2 = & \|\hat f_{0,n} - f_0\|_n^2  = O_{\PP}\left(n^{\frac{m_0}{2m_f + d}}(\log n)^{\frac{d+1}{2}}\right),\nonumber\\
\end{align}
By \eqref{eq:pfsobonowdGNq1q2} and \eqref{eq:pfsobonowdGNJ2}, we obtain
\begin{align}\label{eq:pfsobonowdGNJX}
    J_2 = O_{\PP}\left(n^{-\frac{2m_f}{2m_f + d}}(\log n)^{D+1}\right).
\end{align}
Next, we consider bounding $\|g_t - f^*\|_{L_2(P_\Xb)}$. Similar to the proof in Appendix \ref{app:pfthmsoboNNnwd}, it suffices to consider bounding the difference between $\|g_t - f_n^*\|_{L_2(P_\Xb)}$ and $\|g_t - f_n^*\|_n$. Lemma \ref{lem:GPembedding} implies that 
\begin{align}\label{eq:pfsobonowdgpsi}
    \|\tilde g\|_{\cH_{\sigma_n/\sqrt{2}}(\Omega)}^2\leq C_7\sigma_n^{-D}\|\tilde g\|_{\cH_{\tilde K_S}(\Omega)}^2 = O_{\PP}\left(n^{\frac{2m_0+D}{2m_f + d}}(\log n)^{D+1}\right).
\end{align}
Consider function class 
$\cG = \{h: h = (g_t-f_n^*)/(2C_8n^{\frac{m_0+D/2}{2m_f+D}}(\log n)^{(D+1)/2})\}$, where the constant $C_8$ is taken such that $\|h_1\|_{\cH_{\sigma_n/\sqrt{2}}(\Omega)} < 1$ for all $h_1\in \cG$. Taking $r=(\log n)^{-1}$ in Lemma \ref{lem:ineqGRKHS}, together with the extension theorem leads to
\begin{align*}
    \|h_1\|_{L_\infty(\Omega)}\leq C_9r^{-\frac{D}{4}}\sigma_n^{\frac{D(r-1)}{2}}\|h_1\|_{L_2(P_\Xb)}^{1-r}\|h_1\|_{\cH_{\sigma_n/\sqrt{2}}(\Omega)}^r,
\end{align*}
for all $h_1\in \cG$. Therefore, we have
\begin{align*}
    c_1:= \sup_{h_1\in \cG}\|h_1\|_{L_\infty(\Omega)} \leq C_{9}r^{-\frac{D}{4}}\sigma_n^{\frac{D(r-1)}{2}}R_1^{1-r},
\end{align*}
where $R_1 = \sup_{h_1\in \cG}\|h_1\|_{L_2(P_\Xb)} \leq \sup_{h_1\in \cG}\|h_1\|_{L_\infty(\Omega)}\leq  \sup_{h_1\in \cG}\|h_1\|_{\cH_{\sigma_n/\sqrt{2}}(\Omega)}<1$, because of the reproducing property. Taking $c=C_{9}r^{-\frac{D}{4}}\sigma_n^{\frac{D(r-1)}{2}}R_1^{1-r}$ and $\delta_n =C_{10}(\sigma_n^dr^{-D-1}c^{-2})^{\frac{1}{r+2}}$ in Lemma \ref{lemmaratio1}, it can be checked that
\begin{align*}
    C_{11}n\delta_n^2c^{-2} \geq H(\delta_n,\cB_{\cH_{\sigma_n/\sqrt{2}}(\Omega)},\|\cdot\|_{L_\infty(\Omega)}).
\end{align*}
By repeating the proof in Appendix \ref{app:pfthmsoboNNnwd}, we obtain that
\begin{align*}
    \|g_t-f^*\|_{L_2(P_\Xb)} = O_{\PP}\left(n^{-\frac{2m_f}{2m_f + D}}(\log n)^{D+1}\right),
\end{align*}
which, together with \eqref{eq:thmfsnGNwd1X} and \eqref{eq_ftgtc2Gnn}, implies 
\begin{align*}
    \|f_t-f^*\|_{L_2(P_\Xb)} = O_{\PP}\left(n^{-\frac{2m_f}{2m_f + D}}(\log n)^{D+1}\right).
\end{align*}
% for sufficiently large $N$ such that $nN^{-2} + n^3\beta^2t^2N^{-4}\leq n^{\frac{2m_f}{2m_f + D}}(\log n)^{D+1}$. 

This finishes the proof.\hfill\BlackBox

\subsection{With weight decay}

The results can be obtained by merely repeating the proof in Appendix \ref{app:pfthmsoboNNwd}, where the only difference is that the corresponding convergence rate for $I_2$ (in \eqref{eq:thmfsNNwd1X} of Appendix \ref{app:pfthmsoboNNwd}) is obtained via the proof in Appendix \ref{app:pfthmsoboGNnwd}. Thus we omit it here.

{
\section{Proof of Theorem \ref{thm:soboNN_tensor}}\label{app:pfthmsoboNN_tensor}

We first present several lemmas used in this proof.

%\begin{lemma}\label{lem:approx1NN_tensor}
%Suppose the conditions of Theorem \ref{thm:soboNN_tensor} are fulfilled and $f^*\in \cW^{m_f}(\Omega_1)$. Let $f_n^*$ be the solution to the optimization problem
%\begin{align*}
%    \min_{g\in \cH_{\tilde K_S}(\Omega)}\|f^*-g\|_{L_2(P_\Xb)}^2 + \lambda_n\|g\|_{\cH_{\tilde K_S}(\Omega)}^2.
%\end{align*}
%Then %if $m_0\leq m_f$, we have 
%\begin{align*}
%    \|f^*-f_n^*\|_{L_2(P_\Xb)}^2 + \lambda_n\|f_n^*\|_{\cH_{\tilde K_S}(\Omega)}^2 \lesssim \max\left((\lambda_n\sigma_n^{2m_\varepsilon D})^{\frac{m_f}{(m_0+m_\varepsilon)D}},\lambda_n^{\frac{\min(m_0,m_f)}{m_0}}\right).
%\end{align*}
%and 
%if $m_0> m_f$, we have 
%\begin{align}\label{eq:pflemapprox1NNgoal2}
%    \|f^*-f_n^*\|_{L_2(P_\Xb)}^2 + \lambda_n\|f_n^*\|_{\cH_{\tilde K_S}(\Omega)}^2 \lesssim \max\left((\lambda_n\sigma_n^{2m_\varepsilon D})^{\frac{m_f}{(m_0+m_\varepsilon)D}},\lambda_n^{\frac{m_f}{m_0D}}\right).
%\end{align}
%\end{lemma}

\begin{lemma}\label{lem:approx1NN_tensor_1}
Suppose the conditions of Theorem \ref{thm:soboNN_tensor} are fulfilled and $f^*\in \cMW^{m_f}(\Omega_1)$. Let $f_n^*$ be the solution to the optimization problem
\begin{align*}
    \min_{g\in \cH_{\tilde K_S}(\Omega)}\|f^*-g\|_{L_2(P_\Xb)}^2 + \lambda_n\|g\|_{\cH_{\tilde K_S}(\Omega)}^2.
\end{align*}
Then %if $m_0\leq m_f$, we have 
\begin{align*}
    &\quad\|f^*-f_n^*\|_{L_2(P_\Xb)}^2 + \lambda_n\|f_n^*\|_{\cH_{\tilde K_S}(\Omega)}^2\lesssim \sum_{\bl\in\{0,1\}^D:|\bl|\geq 1}(\lambda_n\sigma_n^{2m_\varepsilon|\bl|})^{\frac{m_f}{m_0+m_\varepsilon}}
\end{align*}
\end{lemma}

\begin{lemma}\label{lem:approx2NN_tensor}
Suppose the conditions of Theorem \ref{thm:soboNN_tensor} are fulfilled. Let $f_n^*$ be as in Lemma \ref{lem:approx1NN_tensor_1}.
Suppose there exists $T>0$ (depending on $n$) such that
\begin{align*}
    \|f^*-f_n^*\|_{L_2(P_\Xb)}^2 + \lambda_n\|f_n^*\|_{\cH_{\tilde K_S}(\Omega)}^2\leq T.
\end{align*}
Let $\hat f_n$ be the solution to the optimization problem
\begin{align}\label{eq:lem2approx2NNt}
    \|\by-\hat f_n\|_n^2 + \lambda_n\|\hat f_n\|_{\cH_{\tilde K_S}(\Omega)}^2.
\end{align}
Let $p=\frac{1}{m_0+m_\varepsilon}$, $q=\frac{D-1}{2}+\frac{p}{4}$
\begin{align*}
    M_1 = & \max\bigg( \lambda_n^{-\frac{p}{2(4-p)}}\left(\sigma_n^{-d/2}n^{-1/2}(T + n^{-1/2}T^{1/2})^{\frac{1}{2}-\frac{p}{4}}\big|\log (T + n^{-1/2}T^{1/2})\big|^{q}\right)^{\frac{2}{4-p}}, \nonumber\\
    & (T + n^{-1/2}T^{1/2})^{1/2},\sigma_n^{-d/2}n^{-1/2}\lambda_n^{-\frac{p}{4}}\big|\log (\sigma_n^{-d/2}n^{-1/2}\lambda_n^{-\frac{p}{4}})\big|^{q},  \nonumber\\
    &  \left(\sigma_n^{-d/2}n^{-1/2}(\lambda_n^{-1}T)^{\frac{p}{2}}(T+n^{-1/2}T^{1/2})^{1-\frac{p}{2}}\big|\log (T+n^{-1/2}T^{1/2})\big|^{q}\right)^{1/2}, \nonumber\\
    & (\sigma_n^{-d/2}n^{-1/2})^{\frac{2}{2+p}}(\lambda_n^{-1}T)^{\frac{p}{2(2+p)}}\big|\log \big((\sigma_n^{-d/2}n^{-1/2})^{\frac{2}{2+p}}(\lambda_n^{-1}T)^{\frac{p}{2(2+p)}}\big)\big|^{q\frac{2}{2+p}}\bigg),\nonumber\\
    M_2 = & \max\bigg( \left(\lambda_n^{-1}\sigma_n^{-d/2}n^{-1/2}(T + n^{-1/2}T^{1/2})^{\frac{1}{2}-\frac{p}{4}}\big|\log (T + n^{-1/2}T^{1/2})\big|^q\right)^{\frac{2}{4-p}}, \nonumber\\
    & (\lambda_n^{-1}(T + n^{-1/2}T^{1/2}))^{1/2}, \sigma_n^{-d/2}n^{-1/2}{\lambda_n^{-\frac{2+p}{4}}}\big|\log (\sigma_n^{-d/2}n^{-1/2}\lambda_n^{-\frac{p}{4}})\big|^{q}, \nonumber\\
    & \left(\lambda_n^{-1} \sigma_n^{-d/2}n^{-1/2}(\lambda_n^{-1}T)^{\frac{p}{2}}(T+n^{-1/2}T^{1/2})^{1-\frac{p}{2}}\big|\log (T + n^{-1/2}T^{1/2})) \big|^q\right)^{1/2},\nonumber\\
    & \lambda_n^{-1/2}(\sigma_n^{-d/2}n^{-1/2})^{\frac{2}{2+p}}(\lambda_n^{-1}T)^{\frac{p}{2(2+p)}}\left|\log \left((\sigma_n^{-d/2}n^{-1/2})^{\frac{2}{2+p}}(\lambda_n^{-1}T)^{\frac{p}{2(2+p)}}\right)\right|^{\frac{2q}{2+p}}\bigg).
\end{align*}
Then we have 
\begin{align*}
    \|f^*-\hat f_n\|_n = O_{\PP}(M_1), \|\hat f_n\|_{\cH_{\tilde K_S}(\Omega)} = O_{\PP}(M_2).
\end{align*}
Furthermore, if  $\tilde f_n$ is the solution to the optimization problem
\begin{align}\label{eq:lem2approx22t}
    \|f^*-\tilde f_n\|_n^2 + \lambda_n\|\tilde f_n\|_{\cH_{\tilde K_S}(\Omega)},
\end{align}
then 
\begin{align*}
    \|f^*-\tilde f_n\|_n = O_{\PP}((T+n^{-1/2}T^{1/2})^{1/2}), \|\tilde f_n\|_{\cH_{\tilde K_S}(\Omega)} = O_{\PP}((\lambda_n^{-1}(T + n^{-1/2}T^{1/2}))^{1/2}).
\end{align*}
\end{lemma}

\begin{lemma}[Interpolation inequality for tensored RKHS]\label{lem:ineqTensorRKHS}
Let $g\in\cMW^m(\RR^D)$. For any $1\geq  r>m^{-1}/2$, we have
\begin{align*}
    \|g\|_{L_\infty(\RR^D)}\leq C_r\|g\|_{L_2(\RR^D)}^{1-r}\|g\|_{\cMW^m(\RR^D)}^r,
\end{align*}
where $C_r$ is a constant that only depends on $r$.

\end{lemma}
% and 
% \begin{align*}
%     \|f_n^*\|_{\cH_{\tilde K_S}(\Omega)}^2 \leq \max\left(\lambda_n^{-1}(\lambda_n\sigma_n^{2m_\varepsilon})^{\frac{m_f}{m_0+m_\varepsilon}},\sigma_n^{-2m_0}\right).
% \end{align*}
}
\subsection{Without weight decay}
\label{app:pfthmsoboNNnwd_tensor}
The result can be obtained by merely repeating the proof in Appendix \ref{app:pfthmsoboNNnwd}. We let $\lambda_n \asymp n^{-\frac{2(m_0+m_\varepsilon)}{2m_f+1}}(\log n)^{\frac{2(D-1)(m_0+m_\varepsilon)+1}{2m_f+1}}$, $\sigma_n\asymp 1$, then by Lemma \ref{lem:approx1NN_tensor_1} and Lemma \ref{lem:approx2NN_tensor}, the term $J_2$ in \eqref{eq:pfsobonowdNNJ2X} becomes
\begin{align}\label{eq:pfsobonowdNNJ2X_tensor}
    J_2 = O_{\PP}\left(n^{-\frac{2m_f}{2m_f + 1}}(\log n)^{\frac{2m_f}{2m_f+1}(D-1+\frac{1}{2(m_0+m_\varepsilon)})}\right).
\end{align}
Similar to the proof in Appendix \ref{app:pfthmsoboNNnwd}, we can choose
\begin{align}\label{eq_N0requirete}
    N_0 = \frac{4n^2}{\eta_n(\tilde \Kb)^2},
\end{align}
and obtain that when $N\geq N_0$,
\begin{align}\label{eq_ftgtc2ggg}
    \|f_t - g_t\|_{L_2(P_\Xb)} = O_\PP\left(n^{-1/2}\right).
\end{align}

To bound the difference between the empirical norm $\|g_t - f^*\|_n$ and $\|g_t - f^*\|_{L_2(P_\Xb)}$. By \eqref{eq:gtRKHSnormpf} and Lemma \ref{lem:approx2NN_tensor}, we have
\begin{align}\label{eq:pfsobonowdNNgpsi_tensor}
   \|g_t\|_{\cN_{\sigma}(\Omega)}^2\leq \sigma_n^{-2m_0} \|g_t\|_{\cH_{\tilde K_S}(\Omega)}^2\leq C_{17}\sigma_n^{-2m_0}\|\tilde g\|_{\cH_{\tilde K_S}(\Omega)}^2= O_{\PP}\left(n^{\nu_1}(\log n)^{\nu_2}\right),
\end{align}
where 
\begin{align*}
\sigma_n \asymp & 1,\\
    \nu_1 = & \frac{2(m_0+m_\varepsilon -m_f)}{2m_f+1},\\
    \nu_2= & 2(m_f-m_0-m_\varepsilon)+\frac{1}{2m_f+1}\left(\frac{m_f}{2(m_0+m_\varepsilon)}-1\right).
\end{align*}

Consider function class 
$\cG = \{h: h = (g_t-f^*)/(Cn^{\nu_1/2}(\log n)^{\nu_2/2}\}$, where the constant $C$ is taken such that $\|h_1\|_{\cN_{\sigma}(\Omega)}<1$ for all $h_1\in \cG$. Select $r=\frac{1}{2}\frac{2m_f+1}{m_0+m_\varepsilon}>\frac{1}{2}\frac{1}{m_0+m_\varepsilon}$, then Lemma \ref{lem:ineqTensorRKHS} leads to
\begin{align*}
    \|h_1\|_{L_\infty(\Omega)}\leq C_{1}\|h_1\|_{L_2(P_\Xb)}^{1-\frac{2m_f+1}{2(m_0+m_\varepsilon)}}\|h_1\|_{\cN_{\sigma}(\Omega)}^{\frac{2m_f+1}{2(m_0+m_\varepsilon)}},
\end{align*}
for all $h_1\in \cG$, which implies
\begin{align*}
    c_1:= \sup_{h_1\in \cG}\|h_1\|_{L_\infty(\Omega)} \leq C_{2}R_1^{1-\frac{2m_f+1}{2(m_0+m_\varepsilon)}},
\end{align*}
where $R_1 = \sup_{h_1\in \cG}\|h_1\|_{L_2(P_\Xb)} \leq \sup_{h_1\in \cG}\|h_1\|_{L_\infty(\Omega)}\leq  \sup_{h_1\in \cG}\|h_1\|_{\cN_{\sigma}(\Omega)}<1$, because of the reproducing property. Taking $c= C_{2} R_1^{1-\frac{1}{2(m_0+m_\varepsilon)}}<1$, and we also let $\delta_n= C_{3}(n^{-1}c^2)^{\frac{m_0+m_\varepsilon}{2(m_0+m_\varepsilon)+1}}(\log n)^{\frac{D-1}{2}+\frac{1}{4(m_0+m_\varepsilon)}}$ in Lemma \ref{lemmaratio1}, it can be checked that
\begin{align*}
    C_{4}n\delta_n^2c^{-2} \geq H(\delta,\cB_{\cH_\sigma(\Omega)},\|\cdot\|_{L_\infty(\Omega)}),
\end{align*}
which implies the conditions of Lemma \ref{lemmaratio1} are fulfilled. Applying Lemma \ref{lemmaratio1} to the case $\|g_t - f^*\|_{L_2(P_\Xb)}^2\geq \delta_n^2n^{\nu_1}(\log n)^{\nu_2}$, together with \eqref{eq:pfsobonowdNNJ2X_tensor}, calculations similar to the proof in section \ref{app:pfthmsoboNNnwd} shows
\[\|g_t - f^*\|_{L_2(P_\Xb)}=O_{\PP}\left(n^{-\frac{m_f}{2m_f + 1}}(\log n)^{\frac{m_f}{2m_f+1}(D-1+\frac{1}{2(m_0+m_\varepsilon)})}\right).\]
This finishes the proof.\hfill\BlackBox
\subsection{With weight decay}
The results can be obtained by merely repeating the proof in Appendix \ref{app:pfthmsoboNNwd}, where the only difference is that the corresponding convergence rate for $I_2$ (in \eqref{eq:thmfsNNwd1X} of Appendix \ref{app:pfthmsoboNNwd}) is obtained via the proof in Appendix \ref{app:pfthmsoboNNnwd_tensor}. Thus we omit it here.

\section{Proof of Theorem \ref{lem:compLemma}}\label{app:gderrorPsit}

% \yy{The treatment below is the key difference to traditional way of early stopping by passing to ridge formulation.}

Similar to \eqref{eq:gdupdnwdg1}, we have
\begin{align*}
    \tilde g_t(\Xb) = & \left(\Ib-(\Ib - \beta\Kb_1)^t\right)\by,
\end{align*}
thus
\begin{align}\label{eq:nwdgerror12}
    \tilde g_t(\Xb) - f^*(\Xb) = -(\Ib - \beta\Kb_1)^t f^*(\Xb) + \left(\Ib-(\Ib - \beta\Kb_1)^t\right)\bepsilon,
\end{align}
where $\Kb_1=(K_1(\bx_j-\bx_k))_{jk}$, and $f^*(\Xb) = (f^*(\bx_1),...,f^*(\bx_n))^T$. Taking expectation with respect to $\bepsilon$, the mean squared prediction error of $\tilde g_t$ with respect to the empirical norm is given by
\begin{align}\label{eq:nwdgerror1}
    \EE\|\tilde g_t - f^*\|_n^2 =  &\frac{1}{n}\left( (f^*(\Xb))^T(\Ib - \beta \Kb_1)^{2t}f^*(\Xb) + \sigma_\epsilon^2{\rm tr}\left(\Ib-(\Ib - \beta \Kb_1)^{t}\right)^2\right)\nonumber\\
    = & \frac{1}{n}J_{11} + \frac{1}{n}J_{12}.
\end{align}
By the representer theorem, the solution to \eqref{eq:complemkrr} is given by
\begin{align}\label{eq:complemkrrsolu} 
    \tilde g(\bx) = \kb_1(\bx)^T(\Kb_1 + n\lambda\Ib)^{-1}\by,
\end{align}
where $\kb_1(\cdot) = (K_1(\cdot-\bx_1),\ldots, K_1(\cdot-\bx_n))^T$. Thus, the mean squared prediction error with respect to the empirical norm of $\tilde g$ can be computed by 
\begin{align}\label{eq:krrgerror1}
    & \EE\|\tilde g - f^*\|_n^2\nonumber\\
    = &\frac{1}{n}\left( (n\lambda)^2(f^*(\Xb))^T(\Kb_1 + n\lambda\Ib)^{-2}f^*(\Xb) + \sigma_\epsilon^2{\rm tr}\left((\Kb_1 + n\lambda\Ib)^{-1}\Kb_1^2(\Kb_1 + n\lambda\Ib)^{-1}\right)^2\right)\nonumber\\
    = & J_{21} + J_{22}.
\end{align}
Let $\eta_1\geq \ldots\geq \eta_n>0$ and $\bv_j$, $j=1,\ldots,n$ be the eigenvalues and corresponding eigenvectors of $\Kb_1$, respectively. By the basic inequalities $1-u\leq \exp(-u)\leq 2e(1+u)^{-2}$ for any $u>0$, the term $J_{11}$ can be bounded by
\begin{align}\label{eq:complemj11j21}
    J_{11} = & \sum_{j=1}^n(1-\beta\eta_j)^{2t}(\bv_j^Tf^*(\Xb))^2\leq \sum_{j=1}^n(1-\beta\eta_j)^t(\bv_j^Tf^*(\Xb))^2\nonumber\\
    \leq & \sum_{j=1}^n\exp(-\beta t\eta_j)(\bv_j^Tf^*(\Xb))^2 \leq 2e\sum_{j=1}^n\frac{(\beta t)^{-2}}{((\beta t)^{-1}+\eta_j)^2}(\bv_j^Tf^*(\Xb))^2\nonumber\\
    = & 2e(\beta t)^{-2} (f^*(\Xb))^T (\Kb_1 + (\beta t)^{-1}\Ib)^{-2}f^*(\Xb)\nonumber\\
    = & 2eJ_{21}
\end{align}
where the last equality is because we choose $n\lambda = (\beta t)^{-1}$.

Next, we consider $J_{12}$. Let $r$ be the smallest integer such that $\beta t \eta_r \leq 1$. Then for $j = 1,...,r-1$, we have
\begin{align}\label{eq:thm44eib1}
    1-(1-\beta\eta_j)^t \leq 1 \leq \frac{2\beta t \eta_j}{1+\beta t \eta_j},
\end{align}
and for $j=r,...,n$, we have
\begin{align}\label{eq:thm44eib2}
    1-(1-\beta\eta_j)^t \leq \beta t\eta_j\leq \frac{2\beta t \eta_j}{1+\beta t \eta_j},
\end{align}
where the first inequality is by Bernoulli's inequality. Combining \eqref{eq:thm44eib1} and \eqref{eq:thm44eib2}, we have
\begin{align}\label{eq:basiceqbtlambda}
    1-(1-\beta\eta_j)^t \leq \frac{2\beta t \eta_j}{1+\beta t \eta_j}
\end{align}
for all $j=1,\ldots,n$. By \eqref{eq:basiceqbtlambda}, the second term $J_{12}$ in \eqref{eq:nwdgerror1} can be bounded by
\begin{align}\label{eq:complemj12j22}
    J_{12} = & \sigma_\epsilon^2 \sum_{j=1}^n (1-(1-\beta\eta_j)^t)^2 \leq \sigma_\epsilon^2\sum_{j=1}^n \frac{4(\beta t \eta_j)^{2}}{(1+\beta t \eta_j)^2}\nonumber\\
    = & 4\sigma_\epsilon^2{\rm tr}\left((\Kb_1 + n\lambda\Ib)^{-1}\Kb_1^2(\Kb_1 + n\lambda\Ib)^{-1}\right)^2 = 4J_{22},
\end{align}
where in the second equality, we use $n\lambda = (\beta t)^{-1}$ again.
By \eqref{eq:nwdgerror1}, \eqref{eq:krrgerror1} \eqref{eq:complemj11j21} and \eqref{eq:complemj12j22}, and $2e>4$, we have  
\begin{align*}
    \EE\|g_t - f^*\|_n^2 \leq 2e \EE\|\tilde g - f^*\|_n^2,
\end{align*}
which finishes the proof of \eqref{eq:complemeq1}.

Next, we consider the RKHS norm of $\tilde g_t$ and show that \eqref{eq:gtRKHSnorm} holds. Direct computation shows that
\begin{align}\label{eq:gtRKHSnormpf}
    & \|g_t\|_{\cH_{K_1}(\Omega)}^2 =  g_t(\Xb)^T \Kb_1^{-1}g_t(\Xb) = \sum_{j=1}^n \frac{(1-(1-\beta\eta_j)^t)^2}{\eta_j}(\bv_j^T\by)^2\nonumber\\
    \leq & \sum_{j=1}^n \frac{4(\beta t )^{2}\eta_j}{(1+\beta t \eta_j)^2}(\bv_j^T\by)^2 = 4\by^T(\Kb_1 + (\beta t)^{-1}\Ib)^{-1}\Kb_1(\Kb_1 + (\beta t)^{-1}\Ib)^{-1}\by\nonumber\\
    = & 4\|\tilde g\|_{\cH_{K_1}(\Omega)}^2,
\end{align}
where the inequality is by \eqref{eq:basiceqbtlambda}, and the last equality is because $n\lambda = (\beta t)^{-1}$. This finishes the proof of \eqref{eq:gtRKHSnorm}.
\hfill\BlackBox

\section{Proof of Lemmas in Appendix \ref{app:PsistPsis}}\label{app:pffinB}

\subsection{Proof of Lemma \ref{lem:converge_empirical_kernel}}
From Assumption \ref{assum:PsiDecay} or Assumption \ref{assum:PsiDecay_tensor}, for any $\bx,\bx'\in\Omega$, the Fourier inversion theorem yields
\begin{align}
    &   \left|\EE_{\bvarepsilon,\bvarepsilon'}\big( K(\bx+\bvarepsilon-(\bx'+\bvarepsilon'))\big)- \frac{1}{N^2}\sum_{k=1}^N \sum_{j=1}^N K(\bx+\bvarepsilon_j-(\bx'+\bvarepsilon_k))\right|\nonumber \\
    = & \left|\int_{\RR^D}\EE_{\bvarepsilon,\bvarepsilon'}\big( e^{i\bomega^T(\bx+\bvarepsilon-\bx'-\bvarepsilon')}\big)\mathcal{F}(K)(\bomega)- \frac{1}{N^2}\sum_{k=1}^N \sum_{j=1}^N e^{i\bomega^T(\bx+\bvarepsilon_k-\bx'-\bvarepsilon'_j)}\mathcal{F}(K)(\bomega){\rm d}\bomega\right|\nonumber\\
    \leq  &  \int_{\RR^D}\left|\big|\EE_{\bvarepsilon}\big( e^{i\bomega^T\bvarepsilon}\big)\big|^2 -\left| \frac{1}{N}\sum_{k=1}^N e^{i\bomega^T\bvarepsilon_k}\right|^2\right|\mathcal{F}(K)(\bomega){\rm d}\bomega\nonumber\\
    \leq & \int_{\RR^D}\bigg|\EE_{\bvarepsilon}\big( e^{i\bomega^T\bvarepsilon}\big) -\frac{1}{N}\sum_{k=1}^N e^{i\bomega^T\bvarepsilon_k}\bigg|\bigg(\big|\EE_{\bvarepsilon}\big( e^{-i\bomega^T\bvarepsilon}\big)\big| +\left|\frac{1}{N}\sum_{k=1}^N e^{-i\bomega^T\bvarepsilon_k}\right|\bigg)\mathcal{F}(K)(\bomega){\rm d}\bomega\nonumber\\
    \leq& 2\int_{\RR^D}\left|\EE_{\bvarepsilon}\big( e^{i\bomega^T\bvarepsilon}\big) -\frac{1}{N}\sum_{k=1}^N e^{i\bomega^T\bvarepsilon_k}\right| \mathcal{F}(K)(\bomega){\rm d}\bomega\label{eq:empirical_ker}.
\end{align}
According to Assumption \ref{assum:augnoise}, $\bvarepsilon$ is sub-Gaussian. 
% $\PP(|\bvarepsilon|>t)\leq L_1e^{-L_2t^a}$ for some $a>0$. 
From \cite{csorgHo1985rates},  we can have the following error estimate for the empirical characteristic function $\frac{1}{N}\sum_{k=1}^N e^{-i\bomega^T\bvarepsilon_k}$ almost surely. Specifically, for any $A>0$, we have
\begin{align}\label{eq_limsupgft}
    \limsup_{N\to\infty} \sqrt{\frac{N}{\log N}}\sup_{\|\omega\|\leq N^A}\bigg|\EE_{\bvarepsilon}\big( e^{i\bomega^T\bvarepsilon}\big) -\frac{1}{N}\sum_{k=1}^N e^{i\bomega^T\bvarepsilon_k}\bigg|\leq 2+\sqrt{2\min(A,1)}+4\sqrt{1+(A+\frac{1}{2})D}.
\end{align}
By \eqref{eq_limsupgft}, \eqref{eq:empirical_ker} can be further bounded by
\begin{align*}
    & 2\int_{\RR^D}\bigg|\EE_{\bvarepsilon}\big( e^{i\bomega^T\bvarepsilon}\big) -\frac{1}{N}\sum_{k=1}^N e^{i\bomega^T\bvarepsilon_k}\bigg| \mathcal{F}(K)(\bomega){\rm d}\bomega\\
    = & 2\int_{\|\bomega\|\leq N^A}\bigg|\EE_{\bvarepsilon}\big( e^{i\bomega^T\bvarepsilon}\big) -\frac{1}{N}\sum_{k=1}^N e^{i\bomega^T\bvarepsilon_k}\bigg| \mathcal{F}(K)(\bomega){\rm d}\bomega + \int_{\|\bomega\| > N^A}\bigg|\EE_{\bvarepsilon}\big( e^{i\bomega^T\bvarepsilon}\big) -\frac{1}{N}\sum_{k=1}^N e^{i\bomega^T\bvarepsilon_k}\bigg| \mathcal{F}(K)(\bomega){\rm d}\bomega\\
    = &  O_{\PP}\bigg(\int_{\|\bomega\|\leq N^A}\sqrt{\frac{\log N}{N}}\mathcal{F}(K)(\bomega){\rm d}\bomega+2\int_{\|\bomega\|>N^A}\mathcal{F}(K)(\bomega){\rm d}\bomega\bigg)
    %\leq & C\bigg\{\sqrt{\frac{\log N}{N}}+\int_{\|\bomega\|>N^A}\mathcal{F}(K)(\omega){\rm d}\bomega\bigg\}
\end{align*}
 If  Assumption \ref{assum:PsiDecay} is satisfied, then we can set $A=(2m_0-d)^{-1}$ and obtain
\begin{align*}
    &\quad \int_{\|\bomega\|\leq N^A}\sqrt{\frac{\log N}{N}}\mathcal{F}(K)(\bomega){\rm d}\bomega+2\int_{\|\bomega\|>N^A}\mathcal{F}(K)(\bomega){\rm d}\bomega\\
    \leq & C_1 \bigg(\int_{\|\bomega\|\leq N^A}\sqrt{\frac{\log N}{N}}(1+\|\bomega\|_2^2)^{-m_0}{\rm d}\bomega+2\int_{\|\bomega\|>N^A}(1+\|\bomega\|_2^2)^{-m_0}{\rm d}\bomega\bigg)\\
    \lesssim &\sqrt{\frac{\log N}{N}}
\end{align*}
where the last inequality is because $m_0>D/2$. Similarly, if Assumption \ref{assum:PsiDecay_tensor} is satisfied, then we set $A=(2m_0-1)^{-1}$ and get
\begin{align*}
    &\quad \int_{\|\bomega\|\leq N^A}\sqrt{\frac{\log N}{N}}\mathcal{F}(K)(\bomega){\rm d}\bomega+2\int_{\|\bomega\|>N^A}\mathcal{F}(K)(\bomega){\rm d}\bomega\\
    \leq & C_2 \bigg(\int_{\|\bomega\|\leq N^A}\sqrt{\frac{\log N}{N}}\prod_{j=1}^D(1+\omega_j^2)^{-m_0}{\rm d}\bomega+2\int_{\|\bomega\|>N^A}\prod_{j=1}^D(1+\omega_j^2)^{-m_0}{\rm d}\bomega\bigg)\\
    \lesssim &\sqrt{\frac{\log N}{N}}+\int_{\max_j|\omega_j|\geq N^A/\sqrt{D}}\prod_{j=1}^D(1+\omega_j^2)^{-m_0}{\rm d}\bomega\\
    \lesssim & \sqrt{\frac{\log N}{N}}.
\end{align*}
This finishes the proof. \hfill\BlackBox

\subsection{Proof of Lemma \ref{lem_gftinf}}

For any $\bx\in\Omega$, by \eqref{eq:predictorapp} and \eqref{eq:predictorappg}, we have 
% the following identities for $f_t(\bx)$ and $g_t(\bx)$, respectively
\begin{align}
    & f_{t}(\bx)=\kb(\bx)^T(\alpha/\beta \Ib + \Kb)^{-1}\by  -\beta\kb(\bx)^T((1-\alpha)\Ib - \beta\Kb)^{t}(\alpha \Ib + \beta\Kb)^{-1}\by,\nonumber\\
    &g_{t}(\bx)=\tilde{\kb}(\bx)^T(\alpha/\beta \Ib + \tilde \Kb)^{-1}\by  -\beta\tilde{\kb}(\bx)^T((1-\alpha)\Ib - \beta\tilde \Kb)^{t}(\alpha\Ib + \beta\tilde \Kb)^{-1}\by.\nonumber
\end{align}
Applying the triangle inequality yields
\begin{align}
    & \|f_t-g_t\|_{L_\infty(\Omega)}\nonumber \\
    \leq & \|\kb(\cdot)^T(\alpha/\beta \Ib + \Kb)^{-1}\by-\tilde{\kb}(\cdot)^T(\alpha/\beta \Ib + \tilde \Kb)^{-1}\by\|_{L_\infty(\Omega)}\nonumber\\
    & + \|\kb(\cdot)^T((1-\alpha)\Ib - \beta\Kb)^{t}(\alpha/\beta \Ib + \Kb)^{-1}\by-\tilde{\kb}(\cdot)^T((1-\alpha)\Ib - \tilde \Kb)^{t}(\alpha/\beta\Ib + \tilde \Kb)^{-1}\by\|_{L_\infty(\Omega)}\nonumber\\
    \leq &\|\big(\kb(\cdot)-\tilde{\kb}(\cdot)\big)^T(\alpha/\beta \Ib +  \Kb)^{-1}\by\|_{L_\infty(\Omega)}\label{eq:gt_ft_Linf_part1}\\
    & + \|\tilde{\kb}(\cdot)^T\big((\alpha/\beta \Ib + \Kb)^{-1}-(\alpha/\beta \Ib + \tilde \Kb)^{-1}\big)\by\|_{L_\infty(\Omega)} \label{eq:gt_ft_Linf_part2}\\
    &  + \|\big(\kb(\cdot)-\tilde{\kb}(\cdot)\big)^T((1-\alpha)\Ib - \beta\Kb)^{t}(\alpha/\beta \Ib + \Kb)^{-1}\by\|_{L_\infty(\Omega)} \label{eq:gt_ft_Linf_part3}\\
     &  + \|\tilde{\kb}(\cdot)^T((1-\alpha)\Ib - \beta\Kb)^{t}\big((\alpha/\beta \Ib +\tilde  \Kb)^{-1}-(\alpha/\beta \Ib + \Kb)^{-1}\big)\by\|_{L_\infty(\Omega)} \label{eq:gt_ft_Linf_part4}\\
    & + \|\tilde{\kb}(\cdot)^T\big(((1-\alpha)\Ib - \beta\tilde\Kb)^{t}-((1-\alpha)\Ib - \beta\Kb)^{t}\big)(\alpha/\beta \Ib +\tilde  \Kb)^{-1}\by\|_{L_\infty(\Omega)}. \label{eq:gt_ft_Linf_part5}
\end{align}
For \eqref{eq:gt_ft_Linf_part1}, we have
\begin{align}
    & \|\big(\kb(\cdot)-\tilde{\kb}(\cdot)\big)^T(\alpha/\beta \Ib +  \Kb)^{-1}\by\|_{L_\infty(\Omega)}\nonumber\\
    \leq & \eta_{n}(\alpha/\beta \Ib + \Kb)^{-1}\|\by\|_2\left(\sup_{\bx\in\Omega}\|\kb(\bx)-\tilde{\kb}(\bx)\|_2\right)\nonumber\\
    \leq & \eta_{n}(\Kb)^{-1}\sqrt{\sum_{j=1}^ny_j^2}
    \left(\sup_{\bx\in\Omega}\sqrt{\sum_{j=1}^n\big(K_s(\bx_i,\bx)-\tilde K_s(\bx_i,\bx)\big)^2}\right)\nonumber\\
    = &  \eta_{n}(\Kb)^{-1}\sqrt{\sum_{j=1}^ny_j^2}\  O_{\PP}\bigg(\sqrt{\frac{n\log N}{N}}\bigg)\nonumber \\
    \leq  &  \eta_{n}(\Kb)^{-1}\sqrt{3n}\big(\max_{j=1,\ldots,n}|f^*(\bx_j)|+\sqrt{\frac{1}{n}\sum_{j=1}^n|\epsilon_j|^2}\big)  O_{\PP}\bigg(\sqrt{\frac{n\log N}{N}}\bigg)\nonumber\\
    = & O_{\PP}\bigg( \eta_{n}(\Kb)^{-1}n\sqrt{\frac{\log N}{N}}\bigg)\nonumber\\
    = & O_{\PP}\bigg( \eta_{n}(\tilde\Kb)^{-1}n\sqrt{\frac{\log N}{N}}\bigg),\label{eq:ft_gt_part1_estimate}
\end{align}
where the fourth line is by Lemma \ref{lem:converge_empirical_kernel}, the sixth line is because $\max_{j=1,\ldots,n}|f^*(\bx_j)|\lesssim \|f^*\|_{\cW^{m_f}(\Omega_1)}$ and $\epsilon_j$'s are sub-Gaussian variables, and the last line is because 
\begin{align}\label{eq:gt_ft_Linf_lambb}
\eta_n(\Kb)= & \eta_n(\tilde \Kb+(\Kb-\tilde{\Kb)})
\geq \eta_n(\tilde \Kb) - n\max_{j,k}|K_S(\bx_j,\bx_k)-\tilde K_S(\bx_j,\bx_k)|\nonumber\\
\geq &\eta_n(\tilde \Kb)- n\sqrt{\frac{\log N}{N}}\geq \frac{1}{2}\eta_n(\tilde\Kb). 
\end{align}

By Gershgorin's theorem \citep{varga}, we have
\begin{align}\label{eq:gt_ft_Linf_maxe}
    \|\Kb-\tilde\Kb\|_2 \leq n\max_{j,k}|K_S(\bx_j,\bx_k)-\tilde K_S(\bx_j,\bx_k)| = O_{\PP}\left(n\sqrt{\frac{\log N}{N}}\right).
\end{align}
Therefore, it can be checked that
\begin{align}\label{eq:gt_ft_Linf_part21}
    & \|(\alpha/\beta \Ib + \Kb)^{-1}-(\alpha/\beta \Ib + \tilde \Kb)^{-1}\|_2 = \|(\alpha/\beta \Ib + \Kb)^{-1}(\alpha/\beta \Ib +\tilde  \Kb)^{-1}(\Kb-\tilde\Kb)\|_2\nonumber \\
    \leq& \frac{n\max_{j,k}|K_S(\bx_j,\bx_k)-\tilde K_S(\bx_j,\bx_k)|}{\eta_{n}(\Kb)\eta_{n}(\tilde \Kb)} =  O_\PP\bigg(\frac{n\sqrt{\log N/N}}{\eta_{n}(\Kb)\eta_{n}(\tilde \Kb)}\bigg)\nonumber\\
    = & O_\PP\bigg(\frac{n\sqrt{\log N/N}}{\eta_{n}(\tilde \Kb)^2}\bigg),
\end{align}
where second line is because of Gershgorin's theorem \citep{varga}, the third line is from Lemma \ref{lem:converge_empirical_kernel}, and the last line is from \eqref{eq:gt_ft_Linf_lambb}. Therefore, plugging \eqref{eq:gt_ft_Linf_part21} into \eqref{eq:gt_ft_Linf_part2} gives us
\begin{align}
     & \|\tilde{\kb}(\cdot)^T\big((\alpha/\beta \Ib + \Kb)^{-1}-(\alpha/\beta \Ib + \tilde \Kb)^{-1}\big)\by\|_{L_\infty(\Omega)}\nonumber\\
    \leq & \sup_{\bx\in\Omega}\|\tilde{\kb}(\bx)\|_2\|\yb\|_2 \|(\alpha/\beta \Ib + \Kb)^{-1}-(\alpha/\beta \Ib + \tilde \Kb)^{-1}\|_2\nonumber\\
    \leq & n\bigg(\sup_{\bx\in\Omega}\max_{j=1,\ldots, n}\tilde{K}_s(\bx_j,\bx)\bigg)\bigg(\sqrt{3}\max_{j=1,\ldots, n}|f^*(\bx_j)|+\sqrt{3}\sqrt{\frac{1}{n}\sum_{j=1}^n|\varepsilon_j|^2}\bigg)  O_\PP\bigg(\frac{n\sqrt{\log N/N}}{\eta_{n}(\tilde \Kb)^2}\bigg)\nonumber\\
    = & O_\PP\bigg(\frac{n^2\sqrt{\log N/N}}{\eta_{n}(\tilde \Kb)^2}\bigg).\label{eq:ft_gt_part2_estimate}
\end{align}
For \eqref{eq:gt_ft_Linf_part3}, because $0<1-\alpha-\beta\eta_1(\Kb)<1$, we have
\begin{align}
     &\|\big(\kb(\cdot)-\tilde{\kb}(\cdot)\big)^T((1-\alpha)\Ib - \beta\Kb)^{t}(\alpha/\beta \Ib + \Kb)^{-1}\by\|_{L_\infty(\Omega)} \nonumber\\
    \leq & \eta_{n}(\Kb)^{-1}\|\by\|_2\left(\sup_{\bx\in\Omega}\|\kb(\bx)-\tilde{\kb}(\bx)\|_2\right)\nonumber \\
    = &  O_{\PP}\bigg( \eta_{n}(\Kb)^{-1}n\sqrt{\frac{\log N}{N}}\bigg) = O_{\PP}\bigg( \eta_{n}(\tilde\Kb)^{-1}n\sqrt{\frac{\log N}{N}}\bigg),\label{eq:ft_gt_part3_estimate}
\end{align}
where the last line is from \eqref{eq:ft_gt_part1_estimate} and \eqref{eq:gt_ft_Linf_lambb}.

Similarly, for \eqref{eq:gt_ft_Linf_part4}, we have
\begin{align}
     & \|\tilde{\kb}(\cdot)^T((1-\alpha)\Ib - \beta\Kb)^{t}\big((\alpha/\beta \Ib +\tilde  \Kb)^{-1}-(\alpha/\beta \Ib + \Kb)^{-1}\big)\by\|_{L_\infty(\Omega)}\nonumber\\
    \leq & \sup_{\bx\in\Omega}\|\tilde{\kb}(\bx)\|_2\|\yb\|_2 \|(\alpha/\beta \Ib + \Kb)^{-1}-(\alpha/\beta \Ib + \tilde \Kb)^{-1}\|_2\nonumber\\
    = &  O_\PP\bigg(\frac{n^2\sqrt{\log N/N}}{\eta_{n}(\tilde \Kb)^2}\bigg)\label{eq:ft_gt_part4_estimate}.
\end{align}

For \eqref{eq:gt_ft_Linf_part5}, we have
\begin{align}
     & \|\tilde{\kb}(\cdot)^T\big(((1-\alpha)\Ib - \beta\tilde\Kb)^{t}-((1-\alpha)\Ib - \beta\Kb)^{t}\big)(\alpha/\beta \Ib +\tilde  \Kb)^{-1}\by\|_{L_\infty(\Omega)}\nonumber\\
    \leq & \sup_{\bx\in\Omega}\|\tilde{\kb}(\bx)\|_2\|(\alpha/\beta \Ib +\tilde  \Kb)^{-1}\|_2\|\by\|_2\|((1-\alpha)\Ib - \beta\tilde\Kb)^{t}-((1-\alpha)\Ib - \beta\Kb)^{t}\|_2\nonumber\\
     \leq & \frac{n}{\eta_n(\tilde \Kb)} \|((1-\alpha)\Ib - \beta\tilde\Kb)^{t}-((1-\alpha)\Ib - \beta\Kb)^{t}\|_2. \label{eq:ft_gt_part5_estimate_1}
    % \leq & \frac{n}{\lambda_n(\tilde \Kb)} \left\|\sum_{j=0}^{t-1}\big((1-\alpha)\Ib-\beta\tilde{\Kb}\big)^j\big((1-\alpha)\Ib-\beta{\Kb}\big)^{t-1-j}\right\|_2. \label{eq:ft_gt_part5_estimate_1}
    \end{align}
    %= & O_\PP\bigg( \|\tilde{\kb}(\cdot)^T(\alpha/\beta \Ib +\tilde  \Kb)^{-1}\by\|_{L_\infty(\Omega)}\|\sum_{j=0}^{t-1}\big((1-\alpha)\Ib-\beta\tilde{\Kb}\big)^j\big((1-\alpha)\Ib-\beta{\Kb}\big)^{t-1-j}\|\bigg)\nonumber\\
The term $\|((1-\alpha)\Ib - \beta\tilde\Kb)^{t}-((1-\alpha)\Ib - \beta\Kb)^{t}\|_2$ can be further bounded by
% For the second term of \eqref{eq:ft_gt_part5_estimate_1}, we have
\begin{align}
& \|((1-\alpha)\Ib - \beta\tilde\Kb)^{t}-((1-\alpha)\Ib - \beta\Kb)^{t}\|_2\nonumber\\
  \leq  &\|\beta\tilde\Kb - \beta\Kb\|_2 \left\|\sum_{j=0}^{t-1}\big((1-\alpha)\Ib-\beta\tilde{\Kb}\big)^j\big((1-\alpha)\Ib-\beta{\Kb}\big)^{t-1-j}\right\|_2\nonumber\\
  = & O_\PP\bigg(\beta n\sqrt{\frac{\log N}{N}}\bigg)\left( \sum_{j=0}^{t-1}\big|(1-\alpha)-\beta \eta_n(\tilde{\Kb})\big|^j\big|(1-\alpha)-\beta\eta_n({\Kb})\big|^{t-1-j}\right)\nonumber\\
  \leq & O_\PP\bigg(\sqrt{\frac{\log N}{N}}\bigg)\left( \sum_{j=0}^{t-1}\big|(1-\alpha)-\beta(\eta_n({\Kb})- \eta_1(\tilde{\Kb}-\Kb))\big|^j\big|(1-\alpha)-\beta\eta_n({\Kb})\big|^{t-1-j}\right)\nonumber\\ 
  \leq & O_\PP\bigg(\sqrt{\frac{\log N}{N}}\bigg)\left(\sum_{j=0}^{t-1}\left|(1-\alpha)-\beta\eta_n( {\Kb})+ O_\PP\bigg(n\sqrt{\frac{\log N}{N}}\bigg)\right|^j\big|(1-\alpha)-\beta\eta_n({  \Kb})\big|^{t-1-j}\right)\nonumber\\ 
  \leq & O_\PP\bigg(t\sqrt{\frac{\log N}{N}}\left|1-\alpha-\beta\eta_n( {\Kb})+ n\sqrt{\frac{\log N}{N}}\right|^t\bigg), \label{eq:ft_gt_part5_estimate}
%   \leq & \sum_{j=0}^{t-1}\big|(1-\alpha)-\beta(\lambda_n({\Kb})- \lambda_1(\tilde{\Kb}-\Kb))\big|^j\big|(1-\alpha)-\beta\lambda_n({\Kb})\big|^{t-1-j}\nonumber\\
%     = &   \sum_{j=0}^{t-1}\left|(1-\alpha)-\beta\lambda_n( {\Kb})+ O_\PP\bigg(n\sqrt{\frac{\log N}{N}}\bigg)\right|^j\big|(1-\alpha)-\beta\lambda_n({  \Kb})\big|^{t-1-j}\nonumber\\
%     \leq & \sum_{j=0}^{t-1}\left|(1-\alpha)-\beta\lambda_n( {\Kb})\big|^j\right|(1-\alpha)-\beta\lambda_n({  \Kb})\big|^{t-1-j}+ n\sqrt{\frac{\log N}{N}}\big|(1-\alpha)-\beta\lambda_n( {\Kb})\big|^j\bigg)\nonumber\\
%     \leq & ts^{t-1} +\frac{s}{1-s} n\sqrt{\frac{\log N}{N}}
\end{align}
where the second line is because of the basic identity $a^t-b^t = (a-b)(\sum_{j=0}^{t-1}a^jb^{t-1-j})$, the third line is because of \eqref{eq:gt_ft_Linf_maxe}, and the fifth line is by the second inequality in \eqref{eq:gt_ft_Linf_lambb}.

Since $\alpha,\beta,$ and $\eta_n( {\Kb})$ are not depending on $N$, we can let $N_0$ satisfy $ n\sqrt{\frac{\log N_0}{N_0}} \leq (\alpha+ \beta\eta_n( {\Kb}))/2$ such that for all $N>N_0+3$
\begin{align*}
    \left|1-\alpha-\beta\eta_n( {\Kb})+ n\sqrt{\frac{\log N}{N}}\right| \leq \left|1-\frac{\alpha+\beta\eta_n( {\Kb})}{2}\right|.
\end{align*}
Let $t_0 = 2/(\alpha+\beta\eta_n( {\Kb}))$, and $h(t) = t(1 - (\alpha+ \beta\eta_n( {\Kb}))/2)^t$. Basic calculation shows that if $t>t_0$, $h(t)$ is a decreasing function. Thus, $h(t)\leq h(t_0)$. By the basic inequality $(1-x)^x\leq e^{-1}$, we obtain that if $t>t_0$, \eqref{eq:ft_gt_part5_estimate} can be further bounded by
\begin{align}\label{eq:ft_gt_part5_estimate11}
     & t\sqrt{\frac{\log N}{N}}\left|1-\alpha-\beta\eta_n( {\Kb})+ n\sqrt{\frac{\log N}{N}}\right|^t\nonumber\\
     \leq & \sqrt{\frac{\log N}{N}} t_0 e^{-1} \leq \sqrt{\frac{\log N}{N}} t_0 \nonumber\\
     = & \sqrt{\frac{\log N}{N}} \frac{2}{\alpha+ \beta\eta_n( {\Kb})} \leq \sqrt{\frac{\log N}{N}} \frac{2n}{n\beta\eta_n( {\Kb})}\nonumber\\
     \leq & C_1\frac{n\sqrt{\log N/N}}{\eta_{n}(\tilde \Kb)},
\end{align}
where we use $n\beta$ is a constant. If $t\leq t_0$, then 
\begin{align}\label{eq:ft_gt_part5_estimate12}
     & t\sqrt{\frac{\log N}{N}}\left|1-\alpha-\beta\eta_n( {\Kb})+ n\sqrt{\frac{\log N}{N}}\right|^t\nonumber\\
     \leq & \sqrt{\frac{\log N}{N}} t_0 \leq C_1\frac{n\sqrt{\log N/N}}{\eta_{n}(\tilde \Kb)},
\end{align}
since $1-\alpha-\beta\eta_n( {\Kb})+ n\sqrt{\frac{\log N}{N}}<1$. Therefore, as long as $N>N_0+3$, by plugging \eqref{eq:ft_gt_part5_estimate}, \eqref{eq:ft_gt_part5_estimate11}, and \eqref{eq:ft_gt_part5_estimate12} in \eqref{eq:ft_gt_part5_estimate_1}, we have
\begin{align}
    &\|\tilde{\kb}(\cdot)^T(\alpha/\beta \Ib +\tilde  \Kb)^{-1}\by\|_{L_\infty(\Omega)}\|\sum_{j=0}^{t-1}\big((1-\alpha)\Ib-\beta\tilde{\Kb}\big)^j\big((1-\alpha)\Ib-\beta{\Kb}\big)^{t-1-j}\|\nonumber \\
    =& O_\PP\bigg(\frac{n^2\sqrt{\log N/N}}{\eta_{n}(\tilde \Kb)^2}\bigg). \label{eq:ft_gt_part5_estimate_2}
\end{align}
Putting together \eqref{eq:ft_gt_part1_estimate}, \eqref{eq:ft_gt_part2_estimate}, \eqref{eq:ft_gt_part3_estimate}, \eqref{eq:ft_gt_part4_estimate}, and \eqref{eq:ft_gt_part5_estimate_2}, we obtain the final result.
\hfill\BlackBox

\subsection{Proof of Lemma \ref{lem:minimum_eigvalue}}
If Assumption \ref{assum:PsiDecay} is satisfied, the Fourier inversion theorem implies that 
% the following identities holds 
for any $\bx\in\RR^D$, it holds that
\begin{align*}
    \tilde K_S(\bx) = & \int_{\RR^D}\int_{\RR^D}K(\bx+\bvarepsilon-\bvarepsilon')p_\varepsilon(\bvarepsilon)p_\varepsilon(\bvarepsilon'){\rm d}\bvarepsilon{\rm d}\bvarepsilon'\nonumber\\
    = & (2\pi)^{-D/2}\int_{\RR^D}\int_{\RR^D}\int_{\RR^D} e^{-i(\bx+\bvarepsilon-\bvarepsilon')^T\bomega}\cF(K)(\bomega){\rm d}\bomega p_\varepsilon(\bvarepsilon)p_\varepsilon(\bvarepsilon'){\rm d}\bvarepsilon{\rm d}\bvarepsilon'\nonumber\\
    = & (2\pi)^{-D/2}\int_{\RR^D}e^{-i\bx^T\bomega}\cF(K)(\bomega)|\varphi_\varepsilon(\bomega)|^2{\rm d}\bomega,
\end{align*}
where $\varphi_\varepsilon$ is the characteristic function of $p_\varepsilon$.
Thus, by the Fourier theorem, 
\begin{align*}
    \cF(\tilde K_S)(\bomega) = \cF(K)(\bomega)|\varphi_\varepsilon(\bomega)|^2.
\end{align*}
Therefore, for any $\boldsymbol{a}\in\RR^n$, we have
\begin{align}
    \boldsymbol{a}^T \tilde \Kb\boldsymbol{a}=&\sum_{j=1}^n\sum_{k=1}^na_j\tilde K_S(\bx_j-\bx_k)a_k\nonumber \\
    =& (2\pi)^{-D/2}\int_{\RR^D}\sum_{j,k=1}^na_je^{-i(\bx_j-\bx_k)^T\bomega}a_k\cF(K)(\bomega)|\varphi_\varepsilon(\bomega)|^2{\rm d}\bomega\nonumber \\
    \geq & C_1\int_{\RR^D}\left|\sum_{k=1}a_ke^{i\bomega^T\bx_k}\right|^2\big(1+\|\bomega\|^2_2\big)^{-m_0}|\varphi_\varepsilon(\bomega)|^2{\rm d}\bomega.\label{eq:Kernel_mat_1}
\end{align}
where $C_1$ is a constant only depending on $D$.
Similarly, if Assumption \ref{assum:PsiDecay_tensor}  and Assumption \ref{assum:augnoise} (C2) are  satisfied, the Fourier inversion theorem implies that for any $\bx\in\RR^D$,
\begin{align*}
    \tilde K_S(\bx,\bx') = & (2\pi)^{-D/2}\int_{\RR^D}e^{-i(\bx-\bx')^T\bomega}\prod_{j=1}^D\cF(K_j)(\omega_j)|\varphi_\varepsilon(\bomega)|^2{\rm d}\bomega.
    % \geq & C_D\int_{\RR^D}e^{-i(\bx-\bx')^T\bomega}\prod_{j=1}^D|1+\omega_j^2|^{-m_0}|1+\sigma_n^2\omega_j^2|^{-m_\varepsilon}{\rm d}\bomega
\end{align*}
Thus, for any $\{a_i\}_{i=1}^n\subset\RR$, we have
\begin{align}
\ba^T \tilde \Kb\ba = & \sum_{k,j=1}^na_k\tilde K_S(\bx_k,\bx_j)a_j\nonumber\\
    \geq  & C_2\int_{\RR^D}\left|\sum_{k=1}a_ke^{i\bomega^T\bx_k}\right|^2\prod_{j=1}^D|1+\omega_j^2|^{-m_0}|1+\sigma_n^2\omega_j^2|^{-m_\varepsilon}{\rm d}\bomega\nonumber \\
    \geq &  C_2\int_{\RR^D}\left|\sum_{k=1}a_ke^{i\bomega^T\bx_k}\right|^2(1+\|\bomega\|^2_2)^{-m_0D}(1+\sigma_n^2\|\bomega\|^2_2)^{-m_\varepsilon D}{\rm d}\bomega \label{eq:Kernel_mat_2}
\end{align}
where $C_2$ is only depending on $D$.

We then apply Theorem 12.3 of \cite{wendland2004scattered} on \eqref{eq:Kernel_mat_1} and \eqref{eq:Kernel_mat_2}, respectively, and the final results can be straightforwardly derived.
\hfill\BlackBox

% Then the Fourier transform of $k_\sigma(\cdot)$ is
% \begin{align}\label{eq:appFourGaussian}
%     \cF(k_\sigma)(\bomega) = (2\sigma)^d e^{-\sigma^2\|\bomega\|_2^2}.
% \end{align}
% Let $\cH_\sigma(\RR^D)$ be the RKHS generated by $k_\sigma(\bx-\bx')$.

\section{Proof of Lemmas in Appendix \ref{app:pfthmsoboNN}}\label{app:pfappC}
In this section, we present the proof of lemmas in Appendix \ref{app:pfthmsoboNN}.

\subsection{Proof of Lemma \ref{lem:approx1NN}}\label{app:pfapprox1NN}

% Note that it suffices to show
% \begin{align}\label{eq:pflemapprox1NNgoal}
%     \|f^*-f_n^*\|_{L_2(P_\Xb)}^2 + \lambda_n\|f_n^*\|_{\cH_{\tilde K_S}(\Omega)}^2 \leq \max\left((\lambda_n\sigma_n^{2m_\varepsilon})^{\frac{m_f}{m_0+m_\varepsilon}},\lambda_n\sigma_n^{-2m_0}\right).
% \end{align}
Let $\tilde f_n^*$ be the solution to the optimization problem
\begin{align}\label{eq:pflemapprox1NNe1}
    \min_{g\in \cH_{\tilde K_S}(\RR^D)}\|f^*-g\|_{L_2(\RR^D)}^2 + \lambda_n\|g\|_{\cH_{\tilde K_S}(\RR^D)}^2.
\end{align}
Since $f_n^*$ is the solution to \eqref{eq:krrestoraNN1}, we have 
\begin{align}\label{eq:pflemapprox1NNe2}
    \|f^*-f_n^*\|_{L_2(P_\Xb)}^2 + \lambda_n\|f_n^*\|_{\cH_{\tilde K_S}(\Omega)}^2 \leq \|f^*-\tilde f_n^*\|_{L_2(P_\Xb)}^2 + \lambda_n\|\tilde f_n^*\|_{\cH_{\tilde K_S}(\Omega)}^2.
\end{align}
Let $f_1=f^*-\tilde f_n^*$. Then $f_1$ is well-defined in $\RR^D$ and the Fourier inversion theorem implies that
\begin{align}\label{eq:pflemapprox1NNe3}
    \|f_1\|_{L_2(P_\Xb)}^2 = & \left(\int_\Omega \left|\int_{\RR^D}e^{i\bx^T\bomega}(\cF(f_1)(\bomega)){\rm d}\bomega\right|^2{\rm d}P_\Xb\right)\nonumber\\
    \leq & \left(\int_{\RR^D}\left(\int_\Omega\left|e^{i\bx^T\bomega}(\cF(f_1)(\bomega))\right|^2{\rm d}P_\Xb\right)^{1/2}{\rm d}\bomega\right)^2\nonumber\\
    \leq & C_1 \left(\int_{\RR^D}\left|(\cF(f_1)(\bomega))\right|{\rm d}\bomega\right)^2\nonumber\\
    \leq & C_1\int_{\RR^D}\left|(\cF(f_1)(\bomega))\right|^2{\rm d}\bomega\nonumber\\
    = & C_1\|f_1\|_{L_2(\RR^D)},
\end{align}
where the first inequality is by Minkowski's integral inequality, the second inequality is by the finiteness of $P_\Xb$, the third inequality is by Jensen's inequality, and the last equality is because of Parseval's identity.

Combining \eqref{eq:pflemapprox1NNe2} and \eqref{eq:pflemapprox1NNe3}, we have
\begin{align}\label{eq:pflemapprox1NNb1}
    \|f^*-f_n^*\|_{L_2(P_\Xb)}^2 + \lambda_n\|f_n^*\|_{\cH_{\tilde K_S}(\Omega)}^2 \leq & C_1\|f^*-\tilde f_n^*\|_{L_2(\RR^D)}^2 + \lambda_n\|\tilde f_n^*\|_{\cH_{\tilde K_S}(\RR^D)}^2\nonumber\\
    \leq & \max(C_1,1)\left(\|f^*-\tilde f_n^*\|_{L_2(\RR^D)}^2 + \lambda_n\|\tilde f_n^*\|_{\cH_{\tilde K_S}(\RR^D)}^2\right).
\end{align}
It remains to bound 
$$\|f^*-\tilde f_n^*\|_{L_2(\RR^D)}^2 + \lambda_n\|\tilde f_n^*\|_{\cH_{\tilde K_S}(\RR^D)}^2.$$

The Fourier inversion theorem implies that
\begin{align}\label{eq:pfapx1Sobonoise1}
    & \|f^*-\tilde f_n^*\|_{L_2(\RR^D)}^2 + \lambda_n\|\tilde f_n^*\|_{\cH_{\tilde K_S}(\RR^D)}^2 = \int_{\RR^D} |\mathcal{F}(f^*)(\bomega)-\mathcal{F}(\tilde f_n^*)(\bomega)|^2 + \lambda_n \frac{|\mathcal{F}(\tilde f_n^*)(\bomega)|^2}{\cF(\tilde K_S)(\bomega)}{\rm d}\bomega\nonumber\\
    \leq & \int_{\RR^D} |\mathcal{F}(f^*)(\bomega)-\mathcal{F}(\tilde g_n^*)(\bomega)|^2 + \lambda_n \frac{|\mathcal{F}(\tilde g_n^*)(\bomega)|^2}{\cF(\tilde K_S)(\bomega)}{\rm d}\bomega\nonumber\\
    \leq & \int_{\RR^D} |\mathcal{F}(f^*)(\bomega)-\mathcal{F}(\tilde g_n^*)(\bomega)|^2 + C_2\lambda_n |\mathcal{F}(\tilde g_n^*)(\bomega)|^2(1+\|\bomega\|_2^2)^{m_0}(1+\sigma_n^2\|\bomega\|_2^2)^{m_\varepsilon}{\rm d}\bomega\nonumber\\
    = & \int_{\RR^D} \frac{C_2\lambda_n(1+\|\bomega\|_2^2)^{m_0}(1+\sigma_n^2\|\bomega\|_2^2)^{m_\varepsilon}}{1+C_2\lambda_n(1+\|\bomega\|_2^2)^{m_0}(1+\sigma_n^2\|\bomega\|_2^2)^{m_\varepsilon}}|\mathcal{F}(f^*)(\bomega)|^2{\rm d}\bomega\nonumber\\
    \leq & \int_{\Omega_1} C_2\lambda_n(1+\|\bomega\|_2^2)^{m_0}(1+\sigma_n^2\|\bomega\|_2^2)^{m_\varepsilon}|\mathcal{F}(f^*)(\bomega)|^2{\rm d}\bomega \nonumber\\
    & + \int_{\Omega_2} C_2\lambda_n(1+\|\bomega\|_2^2)^{m_0}(1+\sigma_n^2\|\bomega\|_2^2)^{m_\varepsilon}|\mathcal{F}(f^*)(\bomega)|^2{\rm d}\bomega + \int_{\Omega_3} |\mathcal{F}(f^*)(\bomega)|^2{\rm d}\bomega\nonumber\\
    = & I_1 + I_2 + I_3,
\end{align}
where $\tilde g_n^*$ minimizes
\begin{align*}
    \int_{\RR^D} |\mathcal{F}(f^*)(\bomega)-\mathcal{F}(\tilde g_n^*)(\bomega)|^2 + C_2\lambda_n |\mathcal{F}(\tilde g_n^*)(\bomega)|^2(1+\|\bomega\|_2^2)^{m_0}(1+\sigma_n^2\|\bomega\|_2^2)^{m_\varepsilon}{\rm d}\bomega,
\end{align*}
$\Omega_1 = \{\bomega: C_2\lambda_n(1+\|\bomega\|_2^2)^{m_0}(1+\sigma_n^2\|\bomega\|_2^2)^{m_\varepsilon}\leq 1, \sigma_n^2\|\bomega\|_2^2\leq m_\varepsilon^{-1}\}$, $\Omega_2 = \{\bomega: C_2\lambda_n(1+\|\bomega\|_2^2)^{m_0}(1+\sigma_n^2\|\bomega\|_2^2)^{m_\varepsilon} \leq 1, \sigma_n^2\|\bomega\|_2^2\geq m_\varepsilon^{-1}\}$, and $\Omega_3 = \{\bomega: C_2\lambda_n(1+\|\bomega\|_2^2)^{m_0}(1+\sigma_n^2\|\bomega\|_2^2)^{m_\varepsilon} > 1\}$.
% here we also use $\lambda_n\sigma_n^{-2m_0}=o(1)$ to ensure that $\RR^D = \Omega_1\cup \Omega_2\cup\Omega_3$. 
In \eqref{eq:pfapx1Sobonoise1}, the first inequality is because $\tilde f_n^*$ is the solution to the optimization problem \eqref{eq:pflemapprox1NNe1}, and the second inequality is by Assumption \ref{assum:augnoise} (C1).
% Here we note that the constants $C_1$ and $C_2$ are not depending on $m_\varepsilon$.

Since $\sigma_n^2\|\bomega\|_2^2\leq m_\varepsilon^{-1}$ for $\bomega\in \Omega_1$, the first term $I_1$ in \eqref{eq:pfapx1Sobonoise1} can be bounded by
\begin{align}\label{eq:pfapx1SobonoiseI1}
   I_1 \leq & \int_{\Omega_1} C_2\lambda_n(1+\|\bomega\|_2^2)^{m_0}(1+m_\varepsilon^{-1})^{m_\varepsilon}|\mathcal{F}(f^*)(\bomega)|^2{\rm d}\bomega\nonumber\\
   \leq & C_2e \int_{\Omega_1}\lambda_n(1+\|\bomega\|_2^2)^{m_0}|\mathcal{F}(f^*)(\bomega)|^2{\rm d}\bomega.
\end{align}
If $m_0 \leq m_f$, then we directly have
\begin{align}\label{eq:pfapx1SobonoiseI11}
   I_1 \leq & C_2 e\lambda_n\int_{\Omega_1}(1+\|\bomega\|_2^2)^{m_f} |\mathcal{F}(f^*)(\bomega)|^2d\bomega.
\end{align}
If $m_0 > m_f$, then for $\bomega\in \Omega_1$, we have 
\begin{align*}
    C_2\lambda_n(1+\|\bomega\|_2^2)^{m_0}\leq C_2\lambda_n(1+\|\bomega\|_2^2)^{m_0}(1+\sigma_n^2\|\bomega\|_2^2)^{m_\varepsilon}\leq 1,
\end{align*}
which implies
\begin{align*}
    C_2\lambda_n(1+\|\bomega\|_2^2)^{m_0} \leq \left(C_2\lambda_n(1+\|\bomega\|_2^2)^{m_0}\right)^{\frac{m_f}{m_0}} = C_3 \lambda_n^{\frac{m_f}{m_0}}(1+\|\bomega\|_2^2)^{m_f}.
\end{align*}
therefore, by \eqref{eq:pfapx1SobonoiseI1}, we have
\begin{align}\label{eq:pfapx1SobonoiseI12}
   I_1 \leq & C_4e \lambda_n^{\frac{m_f}{m_0}}\int_{\Omega_1}(1+\|\bomega\|_2^2)^{m_f} |\mathcal{F}(f^*)(\bomega)|^2d\bomega.
\end{align}
% where the constants $C_3$ and $C_4$ are not depending on $m_\varepsilon$.

The second term $I_2$ in \eqref{eq:pfapx1Sobonoise1} can be bounded by
\begin{align}\label{eq:pfapx1SobonoiseI2}
   I_2\leq & \int_{\Omega_2} \left(C_2\lambda_n(1+\|\bomega\|_2^2)^{m_0}(1+\sigma_n^2\|\bomega\|_2^2)^{m_\varepsilon}\right)^{\frac{m_f}{m_0+m_\varepsilon}}|\mathcal{F}(f^*)(\bomega)|^2{\rm d}\bomega\nonumber\\
   \leq & \int_{\Omega_2} \left(C_2\lambda_n(1+\|\bomega\|_2^2)^{m_0}(m_\varepsilon + 1)^{m_\varepsilon}\sigma_n^{2m_\varepsilon}\|\bomega\|_2^{2m_\varepsilon}\right)^{\frac{m_f}{m_0+m_\varepsilon}}|\mathcal{F}(f^*)(\bomega)|^2{\rm d}\bomega\nonumber\\
   \leq & \int_{\Omega_2} \left(C_2\lambda_n(m_\varepsilon + 1)^{m_\varepsilon}\sigma_n^{2m_\varepsilon}(1+\|\bomega\|_2^2)^{m_0}(1+\|\bomega\|_2^2)^{m_\varepsilon}\right)^{\frac{m_f}{m_0+m_\varepsilon}}|\mathcal{F}(f^*)(\bomega)|^2{\rm d}\bomega\nonumber\\
   \leq & (C_2\lambda_n(m_\varepsilon + 1)^{m_\varepsilon}\sigma_n^{2m_\varepsilon})^{\frac{m_f}{m_0+m_\varepsilon}}\int_{\Omega_2}(1+\|\bomega\|_2^2)^{m_f} |\mathcal{F}(f^*)(\bomega)|^2d\bomega,
\end{align}
where the first inequality is because on $\Omega_2$,
\begin{align*}
     C_2\lambda_n(1+\|\bomega\|_2^2)^{m_0}(1+\sigma_n^2\|\bomega\|_2^2)^{m_\varepsilon} \leq 1
\end{align*}
implies
\begin{align*}
    & C_2\lambda_n(1+\|\bomega\|_2^2)^{m_0}(1+\sigma_n^2\|\bomega\|_2^2)^{m_\varepsilon}\leq \left(C_2\lambda_n(1+\|\bomega\|_2^2)^{m_0}(1+\sigma_n^2\|\bomega\|_2^2)^{m_\varepsilon}\right)^{\frac{m_f}{m_0+m_\varepsilon}},
    % \leq & C_3 (\lambda_n\sigma_n^{2m_\varepsilon})^{\frac{m_f}{m_0+m_\varepsilon}}(1+\|\bomega\|_2^2)^{m_f},
\end{align*}
provided $m_0+m_\varepsilon \geq m_f$.

The third term $I_3$ in \eqref{eq:pfapx1Sobonoise1} can be bounded by
\begin{align}\label{eq:pfapx1SobonoiseI3}
   I_3 \leq & \int_{\Omega_3} \left(C_2\lambda_n(1+\|\bomega\|_2^2)^{m_0}(1+\sigma_n^2\|\bomega\|_2^2)^{m_\varepsilon}\right)^{\frac{m_f}{m_0+m_\varepsilon}} |\mathcal{F}(f^*)(\bomega)|^2{\rm d}\bomega\nonumber\\
    \leq & (C_2\lambda_n(m_\varepsilon + 1)^{m_\varepsilon}\sigma_n^{2m_\varepsilon})^{\frac{m_f}{m_0+m_\varepsilon}}\int_{\Omega_3}(1+\|\bomega\|_2^2)^{m_f} |\mathcal{F}(f^*)(\bomega)|^2d\bomega,
\end{align}
where the first inequality is because on $\Omega_3$,
\begin{align*}
    1 \leq & C_2\lambda_n(1+\|\bomega\|_2^2)^{m_0}(1+\sigma_n^2\|\bomega\|_2^2)^{m_\varepsilon}
\end{align*}
implies
\begin{align*}
    1\leq \left(C_2\lambda_n(1+\|\bomega\|_2^2)^{m_0}(1+\sigma_n^2\|\bomega\|_2^2)^{m_\varepsilon}\right)^{\frac{m_f}{m_0+m_\varepsilon}}.
\end{align*}
Note that all constants $C_j$, $j=1,...,4$ are not depending on $m_\varepsilon$. Furthermore, we have
\begin{align}\label{eq:c2sobX}
    C_2^{\frac{m_f}{m_0+m_\varepsilon}} \leq (\max(C_2,1))^{\frac{m_f}{m_0+m_\varepsilon}}\leq \max(C_2,1),
\end{align}
since $m_0+m_\varepsilon \geq m_f$. By \eqref{eq:c2sobX}, plugging \eqref{eq:pfapx1SobonoiseI11} (if $m_0\leq m_f$) or \eqref{eq:pfapx1SobonoiseI12} (if $m_0 > m_f$), \eqref{eq:pfapx1SobonoiseI2}, and \eqref{eq:pfapx1SobonoiseI3} into \eqref{eq:pfapx1Sobonoise1}, together with \eqref{eq:pflemapprox1NNb1}, finishes the proof.
\hfill\BlackBox

\subsection{Proof of Lemma \ref{lem:approx2NN}}\label{app:pfapprox2NN}

For $\bx\in \Omega$, the Fourier inversion theorem implies
\begin{align*}
    \tilde K_S(\bx) = & \int_{\RR^D}\int_{\RR^D}K(\bx+\bvarepsilon-\bvarepsilon')p_\varepsilon(\bvarepsilon)p_\varepsilon(\bvarepsilon'){\rm d}\bvarepsilon{\rm d}\bvarepsilon'\nonumber\\
    = & (2\pi)^{-D/2}\int_{\RR^D}\int_{\RR^D}\int_{\RR^D} e^{-i(\bx+\bvarepsilon-\bvarepsilon')^T\bomega}\cF(K)(\bomega){\rm d}\bomega p_\varepsilon(\bvarepsilon)p_\varepsilon(\bvarepsilon'){\rm d}\bvarepsilon{\rm d}\bvarepsilon'\nonumber\\
    = & (2\pi)^{-D/2}\int_{\RR^D}e^{-i\bx^T\bomega}\cF(K)(\bomega)|\varphi_\varepsilon(\bomega)|^2{\rm d}\bomega,
\end{align*}
where $\varphi_\varepsilon$ is the characteristic function of $p_\varepsilon$.
Thus, by the Fourier theorem, 
\begin{align}\label{eq:relatKsKV}
    \cF(\tilde K_S(\bx))(\bomega) = \cF(K)(\bomega)|\varphi_\varepsilon(\bomega)|^2.
\end{align}

Let $\Psi_{\sigma}$ be a positive definite function satisfying 
\begin{align*}
    c_1\left(1+\frac{\sigma^2}{m_0+m_\varepsilon}\|\bomega\|_2^2\right)^{-(m_0+m_\varepsilon)} \leq  \cF(\Psi_{\sigma})\leq c_2\left(1+\frac{\sigma^2}{m_0+m_\varepsilon}\|\bomega\|_2^2\right)^{-(m_0+m_\varepsilon)}, \forall \bomega\in\mathbb{R}^D,
\end{align*}
and $\cN_{\sigma}(\Omega)$ be the RKHS generated by $\Psi_{\sigma}$, where the constants $c_1$ and $c_2$ are not depending on $m_\varepsilon$. Therefore, for any $f\in \cH_{\tilde K_S}(\Omega)$, we have that
\begin{align*}%\label{eq:approx2nnnb}
    \|f\|_{\cN_{\sigma_n}(\Omega)}^2 = & \int_{\RR^D} \frac{|\mathcal{F}(f)(\bomega)|^2}{\cF(\Psi_{\sigma})(\bomega)}{\rm d}\bomega \nonumber\\
    \leq & C_1\int_{\RR^D} \left(1+\frac{\sigma^2}{m_0+m_\varepsilon}\|\bomega\|_2^2\right)^{m_0+m_\varepsilon}|\mathcal{F}(f)(\bomega)|^2{\rm d}\bomega\nonumber\\
    \leq & C_1 \int_{\RR^D}(1+\|\bomega\|_2^2)^{m_0} (1+\sigma^2\|\bomega\|_2^2)^{m_\varepsilon}|\mathcal{F}(f)(\bomega)|^2{\rm d}\bomega \nonumber\\
    \leq & C_2 \int_{\RR^D}\frac{|\mathcal{F}(f)(\bomega)|^2}{\cF(K)(\bomega)|\varphi_\varepsilon(\bomega)|^2}{\rm d}\bomega\nonumber\\
    = & C_2 \int_{\RR^D}\frac{|\mathcal{F}(f)(\bomega)|^2}{\cF(\tilde K_S)(\bomega)}{\rm d}\bomega,
\end{align*}
provided $\sigma\leq 1$, where the last inequality is because of Assumptions \ref{assum:PsiDecay} and \ref{assum:augnoise} (C1). Thus, we have if $\sigma\leq 1$,
\begin{align}\label{eq:approx2nnnb}
    \|f\|_{\cN_{\sigma}(\Omega)}\leq C_3\|f\|_{\cH_{\tilde K_S}(\Omega)}.
\end{align}

In order to prove Lemma \ref{lem:approx2NN}, we need the following lemmas. Although we can directly apply Corollary A.8 of \cite{hamm2021adaptive} and the entropy number of Sobolev spaces to obtain an upper bound on $H(\delta,\cB_{\cH_\sigma(\Omega)},\|\cdot\|_{L_\infty(\Omega)})$, which is 
\begin{align}\label{eq:EnofSovolev}
    H(\delta,\cB_{\cH_\sigma(\Omega)},\|\cdot\|_{L_\infty(\Omega)})\leq C\sigma^{-d}\delta^{-\frac{D}{m_0+m_\varepsilon}},
\end{align}
where $C$ is a constant \textit{depending on} $m_\varepsilon$. However, the dependency between $C$ and $m_\varepsilon$ is not clear as far as we know, and thus cannot meet our needs when $m_\varepsilon$ is dependent on the sample size $n$. Therefore, we develop Lemma \ref{lem:EnofSovolev_m}, providing a new upper bound on $H(\delta,\cB_{\cH_m([0,1]^D)},\|\cdot\|_{L_\infty([0,1]^D)})$, where the dependency between the upper bound and $m_\varepsilon$ is clearly described. Based on Lemma \ref{lem:EnofSovolev_m}, we provide Lemma \ref{lem:ENofSobonoise}, where the constant is independent with $m_\varepsilon$.

Lemma \ref{lem:Berforsingleg} is a Bernstein-type inequality for a single $g$. See, for example, \cite{massart2007concentration}.

\begin{lemma}\label{lem:ENofSobonoise}
Suppose the conditions of Lemma \ref{lem:approx2NN} are fulfilled. Let $\cB_{\cN_\sigma(\Omega)}$ be a unit ball in $\cN_{\sigma}(\Omega)$. Then for all $\delta>0$, we have
\begin{align*}
    H(\delta,\cB_{\cH_\sigma(\Omega)},\|\cdot\|_{L_\infty(\Omega)})\leq & C\sigma^{-d}(2m-D)^{-\frac{2D}{2m-D}}m^{\frac{2mD}{2m-D}}\delta^{-\frac{2D}{2m-D}}\log(1+\delta^{-1}),
\end{align*}
where the constant $C$ is independent with $m_\varepsilon$, and $m=m_\varepsilon + m_0$.
\end{lemma}

\begin{lemma}\label{lem:Berforsingleg}
Suppose $X_i\sim Unif(\Omega)$ for $i=1, \ldots, n$. Let $g$ be a fixed function. We have for all $t>0$, 
\begin{align*}
    P\left(\left|\|g\|_n^2 - \|g\|_{L_2(P_\Xb)}^2\right|\geq t\right)\leq 2\exp\left(-\frac{nt^2}{8(t+\|g\|_{L_2(P_\Xb)}^2)}\right).
\end{align*}
\end{lemma}

\begin{lemma}\label{lem:lem84geerPS}
Suppose conditions of Theorem \ref{thm:soboGN} are fulfilled. Then for some constant $C_2 > 0$ only related to Assumption \ref{assum:sub-G} and for $\delta > 0$ with
\begin{align*}
    \sqrt{n}\delta > 2C_2\max\left(\int_0^1 H(u,\cB_{\cH_\sigma(\Omega)},\|\cdot\|_{L_\infty(\Omega)})^{1/2}{\rm d}u,1\right),
\end{align*}
we have for $p = \frac{4D}{2(m_0+m_\varepsilon)-D}$, $m=m_0+m_\varepsilon$, and $\sqrt{n}\delta \geq C\sigma^{-d/2}m^{\frac{mD}{2m-D}+\frac{1}{2}}$,
\begin{align*}
    \PP\left(\sup_{g\in \cB_{\cH_\sigma(\Omega)}}\frac{\langle g, \bepsilon\rangle_n}{\|g\|_n^{1-\frac{p}{2}}}\geq \delta\right)\leq C_3p^{-1}\exp\left(-\frac{n\delta^2}{C_3^2}\right),
\end{align*}
where the constants $C$, $C_2$ and $C_3$ are independent with $m_\varepsilon$.
\end{lemma}

\noindent\textit{Proof of Lemma \ref{lem:lem84geerPS}.} The proof can be obtained by applying the peeling-off argument in Lemma 8.4 of \cite{geer2000empirical}. Let $m=m_0+m_\varepsilon$. Note that 
\begin{align*}
    & \int_0^\delta H(u,\cB_{\cH_\sigma(\Omega)},\|\cdot\|_{L_\infty(\Omega)})^{1/2}{\rm d}u\nonumber\\
    \leq & C\sigma^{-d/2}(2m-D)^{-\frac{D}{2m-D}}m^{\frac{mD}{2m-D}}\int_0^\delta u^{-\frac{D}{2m-D}}\sqrt{\log(1+u^{-1})}{\rm d}u\nonumber\\
    \leq & C\sigma^{-d/2}(2m-D)^{-\frac{D}{2m-D}}m^{\frac{mD}{2m-D}}\int_0^\delta u^{-\frac{D}{2m-D}}\sqrt{\frac{2m-D}{2D}\left(1+\frac{1}{u}\right)^{\frac{2D}{2m-D}}}{\rm d}u\nonumber\\
    \leq & C_1\sigma^{-d/2}(2m-D)^{-\frac{D}{2m-D}}m^{\frac{mD}{2m-D}+\frac{1}{2}}\int_0^\delta u^{-\frac{2D}{2m-D}}{\rm d}u\nonumber\\
    = & C_1\sigma^{-d/2}(2m-D)^{-\frac{D}{2m-D}}m^{\frac{mD}{2m-D}+\frac{1}{2}}\left(1-\frac{2D}{2m-D}\right)^{-1}\delta^{1-\frac{2D}{2m-D}}\nonumber\\
    \leq & C_1 \sigma^{-d/2}(2m_f + 2D)^{-\frac{D}{2m_f + 2D}}m^{\frac{mD}{2m-D}+\frac{1}{2}}\left(1-\frac{D}{m_f + 1}\right)^{-1}\delta^{1-\frac{2D}{2m-D}}\nonumber\\
    = & C_2\sigma^{-d/2}m^{\frac{mD}{2m-D}+\frac{1}{2}}\delta^{1-\frac{2D}{2m-D}},
\end{align*}
where the first inequality is by Lemma \ref{lem:ENofSobonoise}, the second inequality is by the basic inequality $\log(1+1/u)\leq a(1+1/u)^{1/a}$ for any $u,a>0$, and the fourth inequality holds as long as $m_\varepsilon \geq m_f+D$. Here the constant $C_2$ is independent of $m$.

Let $p = \frac{4D}{2m-D}$ and $\sqrt{n}\delta \geq 4CC_2\sigma^{-d/2}m^{\frac{mD}{2m-D}+\frac{1}{2}}$, where $C$ is only depending on Assumption \ref{assum:sub-G}. The proof then follows the proof of Lemma 8.4 of \cite{geer2000empirical}, while the last step becomes
\begin{align*}
   & \PP\left(\sup_{g\in \cB_{\cH_\sigma(\Omega)}}\frac{\langle g, \bepsilon\rangle_n}{\|g\|_n^{1-\frac{p}{2}}}\geq \delta\right)\leq \sum_{s=1}^\infty C_3\exp\left(-\frac{n\delta^2}{16C_3^2}2^{sp}\right)\leq C_4p^{-1}\exp\left(-\frac{n\delta^2}{C_4^2}\right)
\end{align*}
where we use a similar approach in the proof of Lemma \ref{lem:lem84geer}. \hfill\BlackBox

\textit{Proof of Lemma \ref{lem:approx2NN}.}
Since $\hat f$ is the solution to the optimization problem \eqref{eq:lem2approx2NN}, it can be seen that
\begin{align}\label{eq:lemapprx2NNbi1}
    \|\hat f_n - \by\|_n^2 + \lambda_n\|\hat f_n\|_{\cH_{\tilde K_S}(\Omega)}^2 \leq \|f_n^* - \by\|_n^2 + \lambda_n\|f_n^*\|_{\cH_{\tilde K_S}(\Omega)}^2,
\end{align}
where $f_n^*$ is as in Lemma \ref{lem:approx1NN}. By rearrangement, \eqref{eq:lemapprx2NNbi1} implies
\begin{align}\label{eq:lemapprx2Sobonoisebi1}
    \|f^*-\hat f_n\|_n^2 + C_5\lambda_n\|\hat f_n\|_{\cH_{\tilde K_S}(\Omega)}^2 \leq  \|f^*-f_n^*\|_n^2 + C_6\lambda_n\|f_n^*\|_{\cH_{\tilde K_S}(\Omega)}^2 + 2\langle \bepsilon, \hat f_n - f_n^*\rangle_n.
\end{align}
% Theorem 10.46 of \cite{wendland2004scattered} states that every RKHS defined on $\Omega$ possesses a natural extension to $\RR^D$ with equivalent norms. Applying this natural extension to $\cH_{\tilde K_S}(\Omega)$, we obtain that 
% \begin{align}\label{eq:lemapprx2bi4}
%     \|f^*-\hat f_n\|_n^2 + C_1\lambda_n\|\hat f_n\|_{\cH_{\tilde K_S}(\Omega)}^2 \leq  \|f^*-f_n^*\|_n^2 + C_2\lambda_n\|f_n^*\|_{\cH_{\tilde K_S}(\RR^D)}^2 + 2\langle \bepsilon, \hat f_n - f_n^*\rangle_n.
% \end{align}
Take 
\begin{align*}
    \delta_n = 4CC_2n^{-1/2}\sigma_n^{-d/2}m^{\frac{mD}{2m-D}+\frac{1}{2}},
\end{align*}
and let $p = \frac{4D}{2m-D}$, where $m = m_0+m_\varepsilon$.
Applying Lemma \ref{lem:lem84geerPS}, with probability at least
\begin{align*}
    C_6p^{-1}\exp\left(-C_7\sigma^{-d}m^{\frac{2mD}{2m-D}+1}\right),
\end{align*}
which converges to zero by our assumption, we have 
\begin{align*}
    2\langle \bepsilon, \hat f_n - f_n^*\rangle_n \leq C_8n^{-1/2}\sigma^{-d/2}m^{\frac{mD}{2m-D}+\frac{1}{2}}\|\hat f_n - f_n^*\|^{1-\frac{p}{2}}(\|\hat f_n\|_{\cN_{\sigma_n}(\Omega)} + \|f_n^*\|_{\cN_{\sigma_n}(\Omega)})^{\frac{p}{2}},
\end{align*}
which, together with \eqref{eq:lemapprx2Sobonoisebi1}, implies
\begin{align}\label{eq:lemapprx2Sobonoisebi2}
    & \|f^*-\hat f_n\|_n^2 + \lambda_n \|\hat f_n\|_{\cH_{\tilde K_S}(\Omega)}^2\nonumber\\
    \leq &  \|f^*-f_n^*\|_n^2 + \lambda_n \|f_n^*\|_{\cH_{\tilde K_S}(\Omega)}^2\nonumber\\
    & + C_8n^{-1/2}\sigma^{-d/2}m^{\frac{mD}{2m-D}+\frac{1}{2}}\|\hat f_n - f_n^*\|^{1-\frac{p}{2}}(\|\hat f_n\|_{\cN_{\sigma_n}(\Omega)} + \|f_n^*\|_{\cN_{\sigma_n}(\Omega)})^{\frac{p}{2}}.
\end{align}

By assumption of Lemma \ref{lem:approx2NN}, we have
\begin{align}\label{eq:apppfeq1}
    \|f^*-f_n^*\|_{L_2(P_\Xb)}^2 + \lambda_n\|f_n^*\|_{\cH_{\tilde K_S}(\Omega)}^2\leq T,
\end{align}
which implies $\|f_n^*\|_{\cH_{\tilde K_S}(\Omega)}^2 = O(\lambda_n^{-1}T)$.
% , where we apply the natural extension again. 

Now we consider bounding the difference between $\|f^*-f_n^*\|_n$ and $\|f^*-f_n^*\|_{L_2(P_\Xb)}$. Since $f_n^*$ does not depend on $\bx_j$'s and $\bepsilon$, we can directly apply Lemma \ref{lem:Berforsingleg} to $\|f^*-f_n^*\|_n$ and obtain that
\begin{align*}
    \left|\|f^*-f_n^*\|_n^2 - \|f^*-f_n^*\|_{L_2(P_\Xb)}^2\right| = O_{\PP}(n^{-1/2})\|f^*-f_n^*\|_{L_2(P_\Xb)},
\end{align*}
which, together with \eqref{eq:apppfeq1}, yields
\begin{align}\label{eq:lemapprx2NNfn1}
    \|f^*-f_n^*\|_n^2 = O_{\PP}\left(T + n^{-1/2}T^{1/2}\right).
\end{align}
Plugging \eqref{eq:lemapprx2NNfn1} into \eqref{eq:lemapprx2Sobonoisebi2}, together with \eqref{eq:apppfeq1}, gives us
\begin{align}\label{eq:lemapprx2NNbi6}
    & \|f^*-\hat f_n\|_n^2 + \lambda_n \|\hat f_n\|_{\cH_{\tilde K_S}(\Omega)}^2\nonumber\\
    = & O_{\PP}\left(T + n^{-1/2}T^{1/2}\right) + O_{\PP}\left( n^{-1/2}\sigma^{-d/2}m^{\frac{mD}{2m-D}+\frac{1}{2}}\|\hat f_n - f_n^*\|^{1-\frac{p}{2}}(\|\hat f_n\|_{\cN_{\sigma_n}(\Omega)} + \|f_n^*\|_{\cN_{\sigma_n}(\Omega)})^{\frac{p}{2}}\right),
    % \nonumber\\
    % & C_2 \sigma^{-\rho/2}p^{-(d+1)/2}n^{-1/2}\|\hat f - f_n^*\|_n^{1-\frac{p}{2}}(\|\hat f\|_{\cH_{\sigma_n/\sqrt{2}}(\Omega)} + \|f_n^*\|_{\cH_{\sigma_n/\sqrt{2}}(\Omega)})^{\frac{p}{2}}.
\end{align}
where we also use $\|f\|_{\cN_{\sigma_n}(\Omega)}\leq C_3\|f\|_{\cH_{\tilde K_S}(\Omega)}$ for all $f\in \cH_{\tilde K_S}(\Omega)$ (see \eqref{eq:approx2nnnb}). Then \eqref{eq:lemapprx2NNbi6} implies either
\begin{align}\label{eq:lemapp1NNfub1}
    \|f^*-\hat f_n\|_n^2 + \lambda_n \|\hat f_n\|_{\cH_{\tilde K_S}(\Omega)}^2 = O_{\PP}\left(T + n^{-1/2}T^{1/2}\right)
\end{align}
or
\begin{align}\label{eq:lemapp1NNfub2}
    & \|f^*-\hat f_n\|_n^2 + \lambda_n \|\hat f_n\|_{\cH_{\tilde K_S}(\Omega)}^2\nonumber\\
    = & O_{\PP}\left( n^{-1/2}\sigma^{-d/2}m^{\frac{mD}{2m-D}+\frac{1}{2}}\|\hat f_n - f_n^*\|^{1-\frac{p}{2}}(\|\hat f_n\|_{\cN_{\sigma_n}(\Omega)} + \|f_n^*\|_{\cN_{\sigma_n}(\Omega)})^{\frac{p}{2}}\right)
\end{align}
In order to solve \eqref{eq:lemapp1NNfub2}, we consider two cases. 

Case 1: $\|\hat f_n\|_{\cH_{\tilde K_S}(\Omega)} \geq \|f_n^*\|_{\cH_{\tilde K_S}(\Omega)}$. In this case, we have  
\begin{align}\label{eq:lemapp1NNfub21}
    & \|f^*-\hat f_n\|_n^2 + \lambda_n \|\hat f_n\|_{\cH_{\tilde K_S}(\Omega)}^2 = O_{\PP}\left( \sigma_n^{-d/2}n^{-1/2}m^{\frac{mD}{2m-D}+\frac{1}{2}}\|\hat f_n - f_n^*\|_n^{1-\frac{p}{2}}\|\hat f_n\|_{\cH_{\tilde K_S}(\Omega)}^{\frac{p}{2}}\right)\nonumber\\
    = & O_{\PP}\left( \sigma_n^{-d/2}n^{-1/2}m^{\frac{mD}{2m-D}+\frac{1}{2}}\|f^* - f_n^*\|_n^{1-\frac{p}{2}}\|\hat f_n\|_{\cH_{\tilde K_S}(\Omega)}^{\frac{p}{2}}\right)\nonumber\\
    & + O_{\PP}\left( \sigma_n^{-d/2}n^{-1/2}m^{\frac{mD}{2m-D}+\frac{1}{2}}\|f^* - \hat f_n \|_n^{1-\frac{p}{2}}\|\hat f_n\|_{\cH_{\tilde K_S}(\Omega)}^{\frac{p}{2}}\right),
\end{align}
where the second equality (with $O_{\PP}$ notation) is because of the triangle inequality and the basic inequality $(a+b)^q \leq a^q + b^q$ for $q\in (0,1)$.

It can be seen that \eqref{eq:lemapp1NNfub21} further implies 
\begin{align}\label{eq:lemapp1NNfub211}
    \|f^*-\hat f_n\|_n^2 + \lambda_n \|\hat f_n\|_{\cH_{\tilde K_S}(\Omega)}^2 =  O_{\PP}\left( \sigma_n^{-d/2}n^{-1/2}m^{\frac{mD}{2m-D}+\frac{1}{2}}\|f^* - f_n^*\|_n^{1-\frac{p}{2}}\|\hat f_n\|_{\cH_{\tilde K_S}(\Omega)}^{\frac{p}{2}}\right),
\end{align}
or
\begin{align}\label{eq:lemapp1NNfub212}
    \|f^*-\hat f_n\|_n^2 + \lambda_n \|\hat f_n\|_{\cH_{\tilde K_S}(\Omega)}^2 =  O_{\PP}\left( \sigma_n^{-d/2}n^{-1/2}m^{\frac{mD}{2m-D}+\frac{1}{2}}\|f^* - \hat f_n\|_n^{1-\frac{p}{2}}\|\hat f_n\|_{\cH_{\tilde K_S}(\Omega)}^{\frac{p}{2}}\right).
\end{align}

Plugging \eqref{eq:lemapprx2NNfn1} into \eqref{eq:lemapp1NNfub211}, we have
\begin{align}\label{eq:lemapp1NNfub2122}
    \|f^*-\hat f_n\|_n^2 + \lambda_n \|\hat f_n\|_{\cH_{\tilde K_S}(\Omega)}^2
    = O_{\PP}\left( \sigma_n^{-d/2}n^{-1/2}m^{\frac{mD}{2m-D}+\frac{1}{2}}(T + n^{-1/2}T^{1/2})^{\frac{1}{2}-\frac{p}{4}}\|\hat f_n\|_{\cH_{\tilde K_S}(\Omega)}^{\frac{p}{2}}\right).
\end{align}
Solving \eqref{eq:lemapp1NNfub2122} yields
\begin{align}\label{eq:lemapp1NNfub2122f}
    \|f^*-\hat f_n\|_n = & O_{\PP}\left(\lambda_n^{-\frac{p}{2(4-p)}}\left(\sigma_n^{-d/2}n^{-1/2}m^{\frac{mD}{2m-D}+\frac{1}{2}}(T + n^{-1/2}T^{1/2})^{\frac{1}{2}-\frac{p}{4}}\right)^{\frac{2}{4-p}}\right),\nonumber\\
    \|\hat f_n\|_{\cH_{\tilde K_S}(\Omega)} = &  O_{\PP}\left(\left(\lambda_n^{-1}\sigma_n^{-d/2}m^{\frac{mD}{2m-D}+\frac{1}{2}}n^{-1/2}(T + n^{-1/2}T^{1/2})^{\frac{1}{2}-\frac{p}{4}}\right)^{\frac{2}{4-p}}\right).
\end{align}

Solving \eqref{eq:lemapp1NNfub212} yields
\begin{align}\label{eq:lemapp1NNfub2121}
    \|f^*-\hat f_n\|_n = & O_{\PP}\left( \sigma_n^{-d/2}n^{-1/2}m^{\frac{mD}{2m-D}+\frac{1}{2}}\lambda_n^{-\frac{p}{4}}\right), \nonumber\\
    \|\hat f_n\|_{\cH_{\tilde K_S}(\Omega)} = & O_{\PP}\left( \sigma_n^{-d/2}n^{-1/2}m^{\frac{mD}{2m-D}+\frac{1}{2}}\lambda_n^{-\frac{2+p}{4}}\right).
\end{align}

Case 2: $\|\hat f_n\|_{\cH_{\tilde K_S}(\Omega)} < \|f_n^*\|_{\cH_{\tilde K_S}(\Omega)}$. In this case, \eqref{eq:lemapp1NNfub2} implies that
\begin{align}\label{eq:lemapp1NNfubX2}
     & \|f^*-\hat f_n\|_n^2 + \lambda_n \|\hat f_n\|_{\cH_{\tilde K_S}(\Omega)}^2 = O_{\PP}\left( \sigma_n^{-d/2}n^{-1/2}m^{\frac{mD}{2m-D}+\frac{1}{2}}\|\hat f_n - f_n^*\|_n^{1-\frac{p}{2}}\|f_n^*\|_{\cH_{\tilde K_S}(\Omega)}^{\frac{p}{2}}\right)\nonumber\\
    = & O_{\PP}\left( \sigma_n^{-d/2}n^{-1/2}m^{\frac{mD}{2m-D}+\frac{1}{2}}\|f^* - f_n^*\|_n^{1-\frac{p}{2}}\|f_n^*\|_{\cH_{\tilde K_S}(\Omega)}^{\frac{p}{2}}\right) \nonumber\\
    & + O_{\PP}\left( \sigma_n^{-d/2}n^{-1/2}m^{\frac{mD}{2m-D}+\frac{1}{2}}\|f^* - \hat f_n \|_n^{1-\frac{p}{2}}\|f_n^*\|_{\cH_{\tilde K_S}(\Omega)}^{\frac{p}{2}}\right),
\end{align}
where the second equality is because of the triangle inequality and the basic inequality $(a+b)^q \leq a^q + b^q$ for $q\in (0,1)$ again.

By \eqref{eq:lemapp1NNfubX2}, we have either
\begin{align}\label{eq:lemapp1NNfubX21}
    \|f^*-\hat f_n\|_n^2 + \lambda_n \|\hat f_n\|_{\cH_{\tilde K_S}(\Omega)}^2 = O_{\PP}\left( \sigma_n^{-d/2}n^{-1/2}m^{\frac{mD}{2m-D}+\frac{1}{2}}\|f^* - f_n^*\|_n^{1-\frac{p}{2}}\|f_n^*\|_{\cH_{\tilde K_S}(\Omega)}^{\frac{p}{2}}\right),
\end{align}
or
\begin{align}\label{eq:lemapp1NNfubX22}
    \|f^*-\hat f_n\|_n^2 + \lambda_n \|\hat f_n\|_{\cH_{\tilde K_S}(\Omega)}^2 =O_{\PP}\left( \sigma_n^{-d/2}n^{-1/2}m^{\frac{mD}{2m-D}+\frac{1}{2}}\|f^* - \hat f_n \|_n^{1-\frac{p}{2}}\|f_n^*\|_{\cH_{\tilde K_S}(\Omega)}^{\frac{p}{2}}\right).
\end{align}

Combining \eqref{eq:lemapp1NNfubX21} and \eqref{eq:lemapprx2NNfn1}, we have
\begin{align}\label{eq:lemapp1NNfubX211}
    % \|f-\hat f\|_n^2 = O_{\PP}\left((\lambda_n^{-1}T)^{\frac{p}{(2+p)}}n^{-\frac{2}{2+p}}\right),\quad \|\hat f\|_{\cH_{\tilde K_S}(\Omega)}^2= O_{\PP}\left(\lambda_n^{-1}\left((\lambda_n^{-1}T)^{\frac{p}{(2+p)}}n^{-\frac{2}{2+p}}\right)^{\frac{1}{2}-\frac{p}{4}}(\lambda_n^{-1}T)^{\frac{p}{4}}n^{-1/2}\right),
    \|f^*-\hat f\|_n^2 = & O_{\PP}\left(\sigma_n^{-d/2}n^{-1/2}m^{\frac{mD}{2m-D}+\frac{1}{2}}(\lambda_n^{-1}T)^{\frac{p}{2}}(T+n^{-1/2}T^{1/2})^{1-\frac{p}{2}}\right),\nonumber\\ 
    \|\hat f\|_{\cH_{\tilde K_S}(\Omega)}^2= &  O_{\PP}\left(\lambda_n^{-1} \sigma_n^{-d/2}n^{-1/2}m^{\frac{mD}{2m-D}+\frac{1}{2}}(\lambda_n^{-1}T)^{\frac{p}{2}}(T+n^{-1/2}T^{1/2})^{1-\frac{p}{2}}\right).
\end{align}
Combining \eqref{eq:lemapp1NNfubX22} and \eqref{eq:lemapprx2NNfn1}, we have
\begin{align}\label{eq:lemapp1NNfubX221}
    \|f^*-\hat f_n\|_n = & O_{\PP}\left(( \sigma_n^{-d/2}n^{-1/2}m^{\frac{mD}{2m-D}+\frac{1}{2}})^{\frac{2}{2+p}}(\lambda_n^{-1}T)^{\frac{p}{2(2+p)}}\right)\nonumber\\
   \|\hat f_n\|_{\cH_{\tilde K_S}(\Omega)}= & O_{\PP}\left(\lambda_n^{-1/2}(\sigma_n^{-d/2}n^{-1/2}m^{\frac{mD}{2m-D}+\frac{1}{2}})^{\frac{2}{2+p}}(\lambda_n^{-1}T)^{\frac{p}{2(2+p)}}\right).
\end{align}
By \eqref{eq:lemapp1NNfub1}, \eqref{eq:lemapp1NNfub2121}, \eqref{eq:lemapp1NNfub2122f}, \eqref{eq:lemapp1NNfubX211}, and \eqref{eq:lemapp1NNfubX221}, we finish the proof.\hfill\BlackBox

% By repeating the procedure in the proof of Lemma \ref{lem:approx2}, we obtain the desired results.

% We assume $p_\varepsilon(\bvarepsilon)$ is symmetric.

% \subsection{Proof of Lemma \ref{lem:rkhseq}}\label{app:pflemrkhseq}

% For any function $f\in \cH_{\tilde K_S}(\RR^D)$, it can be shown that
% \begin{align*}
%     \|f\|_{\cH_{\tilde K_S}(\RR^D)}^2 = 
% \end{align*}

\subsection{Proof of Lemma \ref{lem:ineqPolyRKHS}}\label{app:pflemineqPolyRKHS}
For any function $g\in \cW^m(\RR^D)$ where $m=m_0+m_\varepsilon$, the Fourier inversion theorem implies
\begin{align}\label{eq:pflemem1pk}
    |g(\bx)| = & \left|\int_{\RR^D}e^{i\bx^T\bomega}\cF(g)(\bomega) d\bomega\right| \leq \int_{\RR^D}\left|\cF(g)(\bomega) \right| d\bomega\nonumber\\
    = & \int_{\RR^D}\left|\cF(g)(\bomega) \right|^{1-r}(\cF(k_\sigma)(\bomega))^{r/2}\left|\cF(g)(\bomega) \right|^r(\cF(k_\sigma)(\bomega))^{-r/2} d\bomega\nonumber\\
    \leq & \left(\int_{\RR^D}\left|\cF(g)(\bomega) \right|^{\frac{2(1-r)}{2-r}}(\cF(k_\sigma)(\bomega))^{\frac{r}{2-r}}d\bomega\right)^{\frac{2-r}{2}}\left(\int_{\RR^D}\left|\cF(g)(\bomega) \right|^2(\cF(k_\sigma)(\bomega))^{-1} d\bomega\right)^{\frac{r}{2}}\nonumber\\
    \leq & \left(\int_{\RR^D}\left|\cF(g)(\bomega) \right|^{\frac{2(1-r)}{2-r}}\left|(1+\|\bomega\|_2^2)^{-m}\right|^{\frac{r}{2-r}}d\bomega\right)^{\frac{2-r}{2}}\|g\|_{\cH_\sigma(\RR^D)}^r\nonumber\\
    \leq & \left(\int_{\RR^D}\left|\cF(g)(\bomega) \right|^2 d\bomega \right)^{\frac{1-r}{2}}\left(\int_{\RR^D}(1+\|\bomega\|_2^2)^{-mr}d\bomega\right)^{\frac{1}{2}}\|g\|_{\cH_\sigma(\RR^D)}^r\nonumber\\
    = & \left(\int_{\RR^D}(1+\|\bomega\|_2^2)^{-mr}d\bomega\right)^{\frac{1}{2}} \|g\|_{L_2(\RR^D)}^{1-r}\|g\|_{\cH_\sigma(\RR^D)}^r,
\end{align}
where the second and fourth inequalities are by H\"older's inequality, and the third equality is by Parseval's identity. Taking $r = \frac{D}{2(m_0+m_\varepsilon)}$ in \eqref{eq:pflemem1pk}, we have
\begin{align*}%\label{eq:pflemem2pk}
    |g(\bx)| \leq & \left(\int_{\RR^D}(1+\|\bomega\|_2^2)^{-mr}d\bomega\right)^{\frac{1}{2}} \|g\|_{L_2(\RR^D)}^{1-r}\|g\|_{\cH_\sigma(\RR^D)}^r\nonumber\\
    = & \left(\int_{\RR^D}(1+\|\bomega\|_2^2)^{-\frac{D}{2}}d\bomega\right)^{\frac{1}{2}}\|g\|_{L_2(\RR^D)}^{1-r}\|g\|_{\cH_\sigma(\RR^D)}^r\nonumber\\
    = & C_4\|g\|_{L_2(\RR^D)}^{1-r}\|g\|_{\cH_\sigma(\RR^D)}^r.
\end{align*}
This finishes the proof.
\hfill\BlackBox

\section{Proof of Lemmas in Appendix \ref{app:pfthmsoboGN}}\label{app:pfofGaussianThm}

\subsection{Proof of Lemma \ref{lem:GPembedding}}
By Theorem 10.46 of \cite{wendland2004scattered}, there exists a nature extension of $f\in \cH_{\tilde K_S}(\Omega)$ on $\RR^D$, such that the RKHS norm is preserved. Thus, we can focus on the RKHS $\cH_{\tilde K_S}(\RR^D)$. 
 
By \eqref{eq:relatKsKV}, we have that for any $f\in \cH_{\tilde K_S}(\RR^D)$, 
\begin{align*}%\label{eq:approx2nnnb}
    \|f\|_{\cH_{\tilde K_S}(\RR^D)}^2 = \int_{\RR^D}\frac{|\mathcal{F}(f)(\bomega)|^2}{\cF(K)(\bomega)|\varphi_\varepsilon(\bomega)|^2}{\rm d}\bomega.
\end{align*}
% \subsection{Gaussian Noise}

For normal distribution, the characteristic function satisfies $\varphi_\varepsilon(\bomega) = e^{-\frac{1}{2}\sigma_n^2\|\bomega\|_2^2}$. Let $g_1(u) = \sigma_n^2 u-m_0\log(1+u)$. Taking the derivative, we obtain
\begin{align*}
    g_1'(u) = \sigma_n^2 - \frac{m_0}{1+u},
\end{align*}
which is smaller than zero when $u\in [0,\frac{m_0}{\sigma_n^2}-1)$, and larger than zero when $u\in (\frac{m_0}{\sigma_n^2}-1,\infty)$. Therefore, 
\begin{align*}
    g_1(u) \geq & g_1\left(\frac{m_0}{\sigma_n^2}-1\right) = m_0 - \sigma_n^2 - m_0\log m_0 + 2m_0\log \sigma_n\nonumber\\
    \geq & m_0 - 1 - m_0\log m_0 + 2m_0\log \sigma_n, \forall u\in [0,\infty),
\end{align*}
which implies
\begin{align*}
    (1+u)^{-m_0}e^{\sigma_n^2u} \geq e^{C_1}\sigma_n^{2m_0},
\end{align*}
where $C_1 = m_0 - 1 - m_0\log m_0$. By taking $u=\|\bomega\|_2^2$, Assumption \ref{assum:PsiDecay_tensor} implies
\begin{align}\label{eq:lemapp1lbFour}
    \cF(K)(\bomega)|\varphi_\varepsilon(\bomega)|^2 \geq c_1(1+\|\bomega\|_2^2)^{-m_0}e^{-2\sigma_n^2\|\bomega\|_2^2} \geq C_2\sigma_n^{2m_0}e^{-3\sigma_n^2\|\bomega\|_2^2},
\end{align}
As for an upper bound of $\cF(K)(\bomega)|\varphi_\varepsilon(\bomega)|^2$, direct computation shows that
\begin{align}\label{eq:lemapp1ubFour}
    \cF(K)(\bomega)|\varphi_\varepsilon(\bomega)|^2 \leq c_2(1+\|\bomega\|_2^2)^{-m_0}e^{-\sigma_n^2\|\bomega\|_2^2}\leq c_2e^{-\frac{1}{2}\sigma_n^2\|\bomega\|_2^2}.
\end{align}

% By similar approach as obtaining \eqref{eq:lemapp1lbFour}, it can be shown that an upper bound of $\cF(\tilde K_S(\bx))(\bomega)$ can be obtained as
% \begin{align}\label{eq:lemapp1ubFour}
%     \cF(\tilde K_S(\bx))(\bomega) = \cF(K)(\bomega)|\varphi_\varepsilon(\bomega)|^2 \leq c_1 e^{C_2}\sigma_n^{-2m_0}e^{-\frac{1}{2}\sigma_n^2\|\bomega\|_2^2}.
% \end{align}
% Let $k_\sigma(\bx-\bx')$ be a Gaussian kernel defined by
% \begin{align*}
%     k_\sigma(\bx-\bx') = \exp\left(-\frac{\|\bx-\bx'\|_2^2}{4\sigma^2}\right).
% \end{align*}
By \eqref{eq:gaussK}, the Fourier transform of $k_\sigma(\cdot)$ is
\begin{align}\label{eq:appFourGaussian}
    \cF(k_\sigma)(\bomega) = (2\sigma)^D e^{-\sigma^2\|\bomega\|_2^2}.
\end{align}
Let $\cH_\sigma(\RR^D)$ be the RKHS generated by $k_\sigma(\bx-\bx')$. From \eqref{eq:lemapp1lbFour}, \eqref{eq:lemapp1ubFour}, and \eqref{eq:appFourGaussian}, it can be seen that 
\begin{align}\label{eq:appGauEb1}
\|h_1\|_{\cH_{\tilde K_S}(\RR^D)}^2 = & \int_{\RR^D}\frac{|\mathcal{F}(f)(\bomega)|^2}{\cF(K)(\bomega)|\varphi_\varepsilon(\bomega)|^2}{\rm d}\bomega\nonumber\\
\geq & C_3\int_{\RR^D}|\mathcal{F}(f)(\bomega)|^2e^{\frac{1}{2}\sigma_n^2\|\bomega\|_2^2}{\rm d}\bomega\nonumber\\
\geq & C_4\sigma_n^{D}\|h_1\|_{\cH_{\sigma_n/\sqrt{2}}(\RR^D)}^2,
%   \sigma_n^{2m_0-D}\|h_1\|_{\cH_{\sigma_n/\sqrt{2}}(\RR^D)} \leq C_1\|h_1\|_{\cH_{\tilde K_S}(\RR^D)}
\end{align}
and
\begin{align}\label{eq:appGauEb2}
    \|h_2\|_{\cH_{\tilde K_S}(\RR^D)}^2 = & \int_{\RR^D}\frac{|\mathcal{F}(f)(\bomega)|^2}{\cF(K)(\bomega)|\varphi_\varepsilon(\bomega)|^2}{\rm d}\bomega\nonumber\\
\leq & C_5\int_{\RR^D}|\mathcal{F}(f)(\bomega)|^2\sigma_n^{-2m_0}e^{3\sigma_n^2\|\bomega\|_2^2}{\rm d}\bomega\nonumber\\
\leq & C_6\sigma_n^{-2m_0-D}\|h_2\|_{\cH_{\sqrt{3}\sigma_n}(\RR^D)}^2,
\end{align}
for $h_1\in \cH_{\tilde K_S}(\RR^D)$ and $h_2\in \cH_{\sqrt{3}\sigma_n}(\RR^D)$, where $C_4$ and $C_6$ does not depend on $\sigma_n$. \hfill\BlackBox

\subsection{Proof of Lemma \ref{lem:approx1}}\label{app:pflemapprox1}

By \eqref{eq:pflemapprox1NNe3}, the Fourier inversion theorem, and Parseval's identity, it can be shown that
\begin{align*}
& \|f^*-f_n^*\|_{L_2(P_\Xb)}^2 + \lambda_n\|f_n^*\|_{\cH_{\tilde K_S}(\Omega)}^2\nonumber\\
  \leq  & C_1\left(\|f^*-f_n^*\|_{L_2(\RR^D)}^2 + \lambda_n\|f_n^*\|_{\cH_{\tilde K_S}(\RR^D)}^2\right)\nonumber\\
  = & C_1\left(\int_{\RR^D} |\mathcal{F}(f^*)(\bomega)-\mathcal{F}(f_n^*)(\bomega)|^2 + \lambda_n \frac{|\mathcal{F}(f_n^*)(\bomega)|^2}{\cF(\tilde K_S(\bx))(\bomega)}d\bomega\right)\nonumber\\
    \leq & C_1\left(\int_{\RR^D} |\mathcal{F}(f^*)(\bomega)-\mathcal{F}(\tilde g_n^*)(\bomega)|^2 + \lambda_n \frac{|\mathcal{F}(\tilde g_n^*)(\bomega)|^2}{\cF(\tilde K_S(\bx))(\bomega)}d\bomega\right)\nonumber\\
    \leq & C_1\left(\int_{\RR^D} |\mathcal{F}(f^*)(\bomega)-\mathcal{F}(\tilde g_n^*)(\bomega)|^2 + C_2\lambda_n |\mathcal{F}(\tilde g_n^*)(\bomega)|^2\sigma_n^{-2m_0}e^{3\sigma_n^2\bomega^T\bomega}d\bomega\right)\nonumber\\
    = & C_1\left(\int_{\RR^D} \frac{C_2\lambda_n\sigma_n^{-2m_0}e^{3\sigma_n^2\bomega^T\bomega}}{1+C_2\lambda_n\sigma_n^{-2m_0}e^{3\sigma_n^2\bomega^T\bomega}}|\mathcal{F}(f^*)(\bomega)|^2d\bomega\right)\nonumber\\
    \leq & C_3\left(\int_{\Omega_1} \lambda_n\sigma_n^{-2m_0}e^{3\sigma_n^2\bomega^T\bomega}|\mathcal{F}(f^*)(\bomega)|^2d\bomega + \int_{\Omega_1^C} |\mathcal{F}(f^*)(\bomega)|^2d\bomega\right)\nonumber\\
    = & C_3\left(I_1 + I_2\right),
\end{align*}
where $\tilde g_n^*$ minimizes
\begin{align*}
    \int_{\RR^D} |\mathcal{F}(f^*)(\bomega)-\mathcal{F}(g)(\bomega)|^2 + C_2\lambda_n |\mathcal{F}(g)(\bomega)|^2\sigma_n^{-2m_0}e^{3\sigma_n^2\bomega^T\bomega}d\bomega,
\end{align*}
$\Omega_1 = \{\bomega: C_2\lambda_n\sigma_n^{-2m_0}e^{3\sigma_n^2\bomega^T\bomega}\leq 1\}$, which is the same as $\Omega_1 = \{\bomega: \|\bomega\|^2 < \frac{2m_0\log \sigma_n - \log (C_2\lambda_n)}{3\sigma_n^2}\}$, provided that $C_2\lambda_n\sigma_n^{-2m_0} < 1$, and the third inequality is because of \eqref{eq:lemapp1lbFour}.

Let $g(u) = 3\sigma_n^2u - m_f\log(1+u)$. Taking the derivative, we obtain
\begin{align*}
    g'(u) = 3\sigma_n^2 - \frac{m_f}{1+u},
\end{align*}
which is smaller than zero when $u\in [0,\frac{m_f}{3\sigma_n^2}-1)$, and larger than zero when $u\in (\frac{m_f}{3\sigma_n^2}-1,\infty)$. 

Since $g(0) = 0$ and 
\begin{align*}
    & g\left(\frac{2m_0\log \sigma_n - \log (C_2\lambda_n)}{3\sigma_n^2}\right) = 2m_0\log \sigma_n - \log (C_2\lambda_n) - m_f \log\left(1+\frac{2m_0\log \sigma_n - \log (C_2\lambda_n)}{3\sigma_n^2}\right)\nonumber\\
    \leq & 2m_0\log \sigma_n - \log (C_2\lambda_n) - m_f \log\left((2m_0\log \sigma_n - \log (C_2\lambda_n))/3\right) + 2m_f\log \sigma_n\nonumber\\
    \leq & (2m_0+2m_f)\log \sigma_n - \log (C_2\lambda_n),
\end{align*}
where the last inequality is because $C_2\lambda_n\sigma_n^{-2m_0}=o(1)$, which implies $\log\left((2m_0\log \sigma_n - \log (C_2\lambda_n))/3\right) > 0$ as $n$ becomes large.

Therefore, for $u\in [0,\frac{2m_0\log \sigma_n - \log (C_2\lambda_n)}{3\sigma_n^2}]$, we have
\begin{align*}
    g(u) \leq \max(0, \log(\sigma_n^{(2m_0+2m_f)}(C_2\lambda_n)^{-1})),
\end{align*}
which implies
\begin{align*}
    e^{3\sigma_n^2\|\bomega\|_2^2}\leq  \max(1,\sigma_n^{(2m_0+2m_f)}(C_2\lambda_n)^{-1})(1+\|\bomega\|_2^2)^{m_f}
\end{align*}
for $\bomega\in \Omega_1$. Thus, the term $I_1$ can be bounded by
\begin{align}\label{eq:lemap1I1}
    I_1\leq & \max(\lambda_n\sigma_n^{-2m_0},C_2^{-1}\sigma_n^{2m_f})\int_{\Omega_1} (1+|\bomega|^2)^{m_f}|\mathcal{F}(f^*)(\bomega)|^2d\bomega. 
\end{align}
The term $I_2$ can be bounded by
\begin{align}\label{eq:lemap1I2}
    I_2 \leq & \frac{3\sigma_n^{2m_f}}{(2m_0\log \sigma_n - \log (C_2\lambda_n))^{m_f}}\int_{\Omega_1^C} (1+|\bomega|^2)^{m_f}|\mathcal{F}(f^*)(\bomega)|^2d\bomega\nonumber\\
    \leq & 3C_4\sigma_n^{2m_f}\int_{\Omega_1^C}(1+|\bomega|^2)^{m_f}|\mathcal{F}(f^*)(\bomega)|^2d\bomega,
\end{align}
where the first inequality is because on $\Omega_1^C$, we have $\|\bomega\|^2 \geq \frac{2m_0\log \sigma_n - \log (C_2\lambda_n)}{3\sigma_n^2}$, which implies for sufficiently large $n$,
\begin{align*}
   (1+\|\bomega\|^2)^{m_f} \geq \frac{(2m_0\log \sigma_n - \log (C_2\lambda_n))^{m_f}}{3\sigma_n^{2m_f}},
\end{align*}
and the last inequality is because $C_2\lambda_n\sigma_n^{-2m_0}=o(1)$. Combining \eqref{eq:lemap1I1} and \eqref{eq:lemap1I2} leads to
\begin{align*}
    I_1 +I_2 \leq & C_5\max(\lambda_n\sigma_n^{-2m_0},\sigma_n^{2m_f})\int_{\RR^D} (1+|\bomega|^2)^{m_f}|\mathcal{F}(f^*)(\bomega)|^2d\bomega\nonumber\\
    \leq & C_6\max(\lambda_n\sigma_n^{-2m_0},\sigma_n^{2m_f})\|f^*\|_{\cW^{m_f}(\Omega)}^2,
\end{align*}
which finishes the proof.\hfill\BlackBox

\subsection{Proof of Lemma \ref{lem:approx2}}\label{app:pflemapprox2}

We first present a lemma used in this proof, which states the entropy numbers of RKHSs generated by the Gaussian kernels. Lemma \ref{lem:ENofGRKHS} is an intermediate step of the proof of Theorem A.2 of \cite{hamm2021adaptive}. Lemma \ref{lem:lem84geer} is a direct result of the proof of Lemma 8.4 of \cite{geer2000empirical} and Lemma \ref{lem:ENofGRKHS}. 
\begin{lemma}\label{lem:ENofGRKHS}
Let $4\sigma^2\leq 1$. Then for all $0< p < 2$, there exists a constant $C_1>0$ only depending on $D$ such that for all $\delta>0$, we have
\begin{align*}
    H(\delta,\cB_{\cH_\sigma(\Omega)},\|\cdot\|_{L_\infty(\Omega)})\leq C_1\sigma^{-d}p^{-D-1}\delta^{-p}.
\end{align*}
\end{lemma}

\begin{lemma}\label{lem:lem84geer}
Suppose conditions of Theorem \ref{thm:soboGN} are fulfilled. Then for some constant $C_2 > 0$ only related to the Assumption \ref{assum:sub-G} and for $\delta > 0$ with
\begin{align*}
    \sqrt{n}\delta > 2C_2\max\left(\int_0^1 H(u,\cB_{\cH_\sigma(\Omega)},\|\cdot\|_{L_\infty(\Omega)})^{1/2}{\rm d}u,1\right),
\end{align*}
we have for all $0< p < 2$
\begin{align*}
    \PP\left(\sup_{g\in \cB_{\cH_\sigma(\Omega)}}\frac{\langle g, \bepsilon\rangle_n}{\|g\|_n^{1-\frac{p}{2}}}\geq \delta\right)\leq C_2p^{-1}\exp\left(-\frac{n\delta^2}{C_2}\right).
\end{align*}
\end{lemma}

\noindent\textit{Proof of Lemma \ref{lem:lem84geer}.} In order to characterize the role of $p$ in Lemma \ref{lem:lem84geer}, we note that in the last step of the proof of Lemma 8.4 of \cite{geer2000empirical}, we use
\begin{align*}
    & \sum_{s=1}^\infty C_2\exp\left(-\frac{n\delta^2}{16C_2^2}2^{sp}\right)\leq \sum_{s=1}^\infty C_2\exp\left(-\frac{n\delta^2}{16C_2^2}e^{sp/2}\right)\nonumber\\
    \leq & \sum_{s=1}^\infty C_2\exp\left(-\frac{n\delta^2}{16C_2^2}(1+\frac{sp}{2})\right) = C_2\exp\left(-\frac{n\delta^2}{16C_2^2}\right)\frac{\exp\left(-\frac{np\delta^2}{32C_2^2}\right)}{1-\exp\left(-\frac{np\delta^2}{32C_2^2}\right)}\nonumber\\
    \leq & \frac{32C_2^3}{np\delta^2}\exp\left(-\frac{n\delta^2}{16C_2^2}\right)\leq \frac{8C_2}{p}\exp\left(-\frac{n\delta^2}{16C_2^2}\right),
\end{align*}
where the second and the third inequalities are by $e^u>1+u$ for all $u\in \RR$, and the last inequality is by $n\delta^2>4C_2^2$.

Then if $C_2 \geq 1$,
\begin{align*}
    \frac{8C_2}{p}\exp\left(-\frac{n\delta^2}{16C_2^2}\right) \leq \frac{16C_2^2}{p}\exp\left(-\frac{n\delta^2}{16C_2^2}\right).
\end{align*}
and if $0 < C_2 < 1$,
\begin{align*}
    \frac{8C_2}{p}\exp\left(-\frac{n\delta^2}{16C_2^2}\right) \leq \frac{16C_2}{p}\exp\left(-\frac{n\delta^2}{16C_2}\right).
\end{align*}
The rest of the proof is similar to the proof of Lemma 8.4 of \cite{geer2000empirical}.\hfill\BlackBox

\noindent\textit{Proof of Lemma \ref{lem:approx2}.} Since $\hat f$ is the solution to the optimization problem \eqref{eq:lem2approx2}, we have that
\begin{align}\label{eq:lemapprx2bi1}
    \|\hat f - \by\|_n^2 + \lambda_n\|\hat f\|_{\cH_{\tilde K_S}(\Omega)}^2 \leq \|f_n^* - \by\|_n^2 + \lambda_n\|f_n^*\|_{\cH_{\tilde K_S}(\Omega)}^2,
\end{align}
where $f_n^*$ is as in Lemma \ref{lem:approx1}. By rearrangement, \eqref{eq:lemapprx2bi1} implies
\begin{align*}
    \|f-\hat f_n\|_n^2 + \lambda_n\|\hat f\|_{\cH_{\tilde K_S}(\Omega)}^2 \leq  \|f-f_n^*\|_n^2 + \lambda_n\|f_n^*\|_{\cH_{\tilde K_S}(\Omega)}^2 + 2\langle \bepsilon, \hat f - f_n^*\rangle_n.
\end{align*}
Theorem 10.46 of \cite{wendland2004scattered} states that every RKHS defined on $\Omega$ possesses a natural extension to $\RR^D$ with equivalent norms. Applying this natural extension to $\cH_{\tilde K_S}(\Omega)$, we obtain that 
\begin{align}\label{eq:lemapprx2bi4}
    \|f-\hat f_n\|_n^2 + C_3\lambda_n\|\hat f\|_{\cH_{\tilde K_S}(\Omega)}^2 \leq  \|f-f_n^*\|_n^2 + C_4\lambda_n\|f_n^*\|_{\cH_{\tilde K_S}(\RR^D)}^2 + 2\langle \bepsilon, \hat f - f_n^*\rangle_n.
\end{align}
By assumption, we have
\begin{align*}
    \|f-f_n^*\|_{L_2(P_\Xb)}^2 + \lambda_n\|f_n^*\|_{\cH_{\tilde K_S}(\Omega)}^2\leq T.
\end{align*}
Then Lemma \ref{lem:approx1} implies $\|f_n^*\|_{\cH_{\tilde K_S}(\RR^D)}^2 = O(\lambda_n^{-1}T)$. Taking $p=(\log n)^{-1}\in (0,2)$ and $\delta_n = C_5 \sigma_n^{-d/2}p^{-(D+1)/2}n^{-1/2}$ (where $C_5$ is a constant only depending on $D$), we have
$
    \sqrt{n}\delta_n = C_5 \sigma_n^{-d/2}p^{-(D+1)/2}.
$
Applying Lemma \ref{lem:lem84geer}, we obtain that with probability at least 
$$
    C_6(\log n) \exp(-C_6^{-1}C_5^2\sigma_n^{-2d}p^{-2D-2}),
$$
we have
\begin{align}\label{eq:lemapprx2efs}
    2\langle \bepsilon, \hat f - f_n^*\rangle_n \leq C_7\|\hat f - f_n^*\|_n^{1-\frac{p}{2}}(\|\hat f\|_{\cH_{\sigma_n/\sqrt{2}}(\Omega)} + \|f_n^*\|_{\cH_{\sigma_n/\sqrt{2}}(\Omega)})^{\frac{p}{2}}C_5 \sigma_n^{-d/2}p^{-(D+1)/2}n^{-1/2}.
\end{align} 
Plugging \eqref{eq:lemapprx2efs} into \eqref{eq:lemapprx2bi4} yields
\begin{align}\label{eq:lemapprx2bi5}
    & \|f-\hat f\|_n^2 + \lambda_n \|\hat f\|_{\cH_{\tilde K_S}(\Omega)}^2\nonumber\\
    \leq &  \|f-f_n^*\|_n^2 + \lambda_n \|f_n^*\|_{\cH_{\tilde K_S}(\RR^D)}^2 + C_{8} \sigma_n^{-d/2}p^{-(D+1)/2}n^{-1/2}\|\hat f - f_n^*\|_n^{1-\frac{p}{2}}(\|\hat f\|_{\cH_{\sigma_n/\sqrt{2}}(\Omega)} + \|f_n^*\|_{\cH_{\sigma_n/\sqrt{2}}(\Omega)})^{\frac{p}{2}}.
\end{align}
Now we consider bounding the difference between $\|f-f_n^*\|_n$ and $\|f-f_n^*\|_{L_2(P_\Xb)}$. Since $f_n^*$ does not depend on $\bx_j$ and $\bepsilon$, we can directly apply Lemma \ref{lem:Berforsingleg} to
$\|f-f_n^*\|_n$ and obtain that
\begin{align*}
    \left|\|f-f_n^*\|_n^2 - \|f-f_n^*\|_{L_2(P_\Xb)}^2\right| = O_{\PP}(n^{-1/2})\|f-f_n^*\|_{L_2(P_\Xb)},
\end{align*}
which, together with Lemma \ref{lem:approx1}, yields
\begin{align}\label{eq:lemapprx2fn1}
    \|f-f_n^*\|_n^2 = O_{\PP}\left(T + n^{-1/2}T^{1/2}\right).
\end{align}
Plugging \eqref{eq:lemapprx2fn1} into \eqref{eq:lemapprx2bi5}, together with Lemma \ref{lem:approx1}, gives us
\begin{align}\label{eq:lemapprx2bi6}
    & \|f-\hat f\|_n^2 + \lambda_n \|\hat f\|_{\cH_{\tilde K_S}(\Omega)}^2\nonumber\\
    \leq &  O_{\PP}\left(T + n^{-1/2}T^{1/2}\right) + C_{8} \sigma_n^{-d/2 - \frac{pD}{4}}p^{-(D+1)/2}n^{-1/2}\|\hat f - f_n^*\|_n^{1-\frac{p}{2}}(\|\hat f\|_{\cH_{\tilde K_S}(\Omega)} + \|f_n^*\|_{\cH_{\tilde K_S}(\Omega)})^{\frac{p}{2}},
    % \nonumber\\
    % & C_2 \sigma^{-\rho/2}p^{-(d+1)/2}n^{-1/2}\|\hat f - f_n^*\|_n^{1-\frac{p}{2}}(\|\hat f\|_{\cH_{\sigma_n/\sqrt{2}}(\Omega)} + \|f_n^*\|_{\cH_{\sigma_n/\sqrt{2}}(\Omega)})^{\frac{p}{2}}.
\end{align}
where we also use $\sigma_n^{-D/2}\|f_n^*\|_{\cH_{\tilde K_S}(\Omega)}\geq C_{8}\|f_n^*\|_{\cH_{\sigma_n/\sqrt{2}}(\Omega)}$. Then \eqref{eq:lemapprx2bi6} implies either
\begin{align}\label{eq:lemapp1fub1}
    \|f-\hat f\|_n^2 + \lambda_n \|\hat f\|_{\cH_{\tilde K_S}(\Omega)}^2 = O_{\PP}\left(T + n^{-1/2}T^{1/2}\right)
\end{align}
or
\begin{align}\label{eq:lemapp1fub2}
    & \|f-\hat f\|_n^2 + \lambda_n \|\hat f\|_{\cH_{\tilde K_S}(\Omega)}^2\nonumber\\
    \leq & 4C_{8} \sigma_n^{-d/2 - \frac{pD}{4}}p^{-(D+1)/2}n^{-1/2}\|\hat f - f_n^*\|_n^{1-\frac{p}{2}}(\|\hat f\|_{\cH_{\tilde K_S}(\Omega)} + \|f_n^*\|_{\cH_{\tilde K_S}(\Omega)})^{\frac{p}{2}},
\end{align}
In order to solve \eqref{eq:lemapp1fub2}, we consider two cases. 

Case 1: $\|\hat f\|_{\cH_{\tilde K_S}(\Omega)} \geq \|f_n^*\|_{\cH_{\tilde K_S}(\Omega)}$. In this case, we have  
\begin{align}\label{eq:lemapp1fub21}
    & \|f-\hat f\|_n^2 + \lambda_n \|\hat f\|_{\cH_{\tilde K_S}(\Omega)}^2\leq 8C_{8} \sigma_n^{-d/2 - \frac{pD}{4}}p^{-(D+1)/2}n^{-1/2}\|\hat f - f_n^*\|_n^{1-\frac{p}{2}}\|\hat f\|_{\cH_{\tilde K_S}(\Omega)}^{\frac{p}{2}}\nonumber\\
    \leq & 8C_{8} \sigma_n^{-d/2 - \frac{pD}{4}}p^{-(D+1)/2}n^{-1/2}\|f - f_n^*\|_n^{1-\frac{p}{2}}\|\hat f\|_{\cH_{\tilde K_S}(\Omega)}^{\frac{p}{2}}\nonumber\\
    & + 8C_{8} \sigma_n^{-d/2 - \frac{pD}{4}}p^{-(D+1)/2}n^{-1/2}\|f - \hat f\|_n^{1-\frac{p}{2}}\|\hat f\|_{\cH_{\tilde K_S}(\Omega)}^{\frac{p}{2}},
\end{align}
where the second equality is because of the basic inequality $(a+b)^q \leq a^q + b^q$ for $q\in (0,1)$.

It can be seen that \eqref{eq:lemapp1fub21} further implies 
\begin{align}\label{eq:lemapp1fub211}
    \|f-\hat f\|_n^2 + \lambda_n \|\hat f\|_{\cH_{\tilde K_S}(\Omega)}^2\leq   8C_{8} \sigma_n^{-d/2 - \frac{pD}{4}}p^{-(D+1)/2}n^{-1/2}\|f - f_n^*\|_n^{1-\frac{p}{2}}\|\hat f\|_{\cH_{\tilde K_S}(\Omega)}^{\frac{p}{2}},
\end{align}
or
\begin{align}\label{eq:lemapp1fub212}
    \|f-\hat f\|_n^2 + \lambda_n \|\hat f\|_{\cH_{\tilde K_S}(\Omega)}^2\leq  8C_{8} \sigma_n^{-d/2 - \frac{pD}{4}}p^{-(D+1)/2}n^{-1/2}\|f - \hat f\|_n^{1-\frac{p}{2}}\|\hat f\|_{\cH_{\tilde K_S}(\Omega)}^{\frac{p}{2}}.
\end{align}
Solving \eqref{eq:lemapp1fub212} yields
\begin{align}\label{eq:lemapp1fub2121}
    \|f-\hat f\|_n \leq & 8C_{8} \sigma_n^{-d/2 - \frac{pD}{4}}p^{-(D+1)/2}n^{-1/2}\lambda_n^{-\frac{p}{4}}, \nonumber\\
    \|\hat f\|_{\cH_{\tilde K_S}(\Omega)} \leq & 8C_{8} \sigma_n^{-d/2 - \frac{pD}{4}}p^{-(D+1)/2}n^{-1/2}\lambda_n^{-\frac{2+p}{4}}.
\end{align}
Plugging \eqref{eq:lemapprx2fn1} into \eqref{eq:lemapp1fub211}, we have
\begin{align}\label{eq:lemapp1fub2122}
    \|f-\hat f\|_n^2 + \lambda_n \|\hat f\|_{\cH_{\tilde K_S}(\Omega)}^2
    \leq 8C_{8} \sigma_n^{-d/2 - \frac{pD}{4}}p^{-(D+1)/2}n^{-1/2}(T + n^{-1/2}T^{1/2})^{\frac{1}{2}-\frac{p}{4}}\|\hat f\|_{\cH_{\tilde K_S}(\Omega)}^{\frac{p}{2}}.
\end{align}
Solving \eqref{eq:lemapp1fub2122} yields
\begin{align}\label{eq:lemapp1fub2122f}
    \|f-\hat f\|_n \leq & \lambda_n^{-\frac{p}{2(4-p)}}\left(8C_{8} \sigma_n^{-d/2 - \frac{pD}{4}}p^{-(D+1)/2}n^{-1/2}(T + n^{-1/2}T^{1/2})^{\frac{1}{2}-\frac{p}{4}}\right)^{\frac{2}{4-p}},\nonumber\\
    \|\hat f\|_{\cH_{\tilde K_S}(\Omega)} \leq &  \left(8C_{8}\lambda_n^{-1} \sigma_n^{-d/2 - \frac{pD}{4}}p^{-(D+1)/2}n^{-1/2}(T + n^{-1/2}T^{1/2})^{\frac{1}{2}-\frac{p}{4}}\right)^{\frac{2}{4-p}}.
\end{align}

Case 2: $\|\hat f\|_{\cH_{\tilde K_S}(\Omega)} < \|f_n^*\|_{\cH_{\tilde K_S}(\Omega)}$. In this case, \eqref{eq:lemapp1fub2} implies that
\begin{align}\label{eq:lemapp1fubX2}
     & \|f-\hat f\|_n^2 + \lambda_n \|\hat f\|_{\cH_{\tilde K_S}(\Omega)}^2\leq 8C_{8} \sigma_n^{-d/2 - \frac{pD}{4}}p^{-(D+1)/2}n^{-1/2}\|\hat f - f_n^*\|_n^{1-\frac{p}{2}}\|f_n^*\|_{\cH_{\tilde K_S}(\Omega)}^{\frac{p}{2}}\nonumber\\
    \leq & 8C_{8} \sigma_n^{-d/2 - \frac{pD}{4}}p^{-(D+1)/2}n^{-1/2}\|f - f_n^*\|_n^{1-\frac{p}{2}}\|f_n^*\|_{\cH_{\tilde K_S}(\Omega)}^{\frac{p}{2}}\nonumber\\
    & + 8C_{8} \sigma_n^{-d/2 - \frac{pD}{4}}p^{-(D+1)/2}n^{-1/2}\|f - \hat f\|_n^{1-\frac{p}{2}}\|f_n^*\|_{\cH_{\tilde K_S}(\Omega)}^{\frac{p}{2}},
\end{align}
where the second equality is because of the basic inequality $(a+b)^q \leq a^q + b^q$ for $q\in (0,1)$.

By \eqref{eq:lemapp1fubX2}, we have either
\begin{align}\label{eq:lemapp1fubX21}
    \|f-\hat f\|_n^2 + \lambda_n \|\hat f\|_{\cH_{\tilde K_S}(\Omega)}^2\leq & C_{9} \sigma_n^{-d/2 - \frac{pD}{4}}p^{-(D+1)/2}n^{-1/2}\|f - f_n^*\|_n^{1-\frac{p}{2}}\|f_n^*\|_{\cH_{\tilde K_S}(\Omega)}^{\frac{p}{2}},
\end{align}
or
\begin{align}\label{eq:lemapp1fubX22}
    \|f-\hat f\|_n^2 + \lambda_n \|\hat f\|_{\cH_{\tilde K_S}(\Omega)}^2\leq & C_{10} \sigma_n^{-d/2 - \frac{pD}{4}}p^{-(D+1)/2}n^{-1/2}\|f - \hat f\|_n^{1-\frac{p}{2}}\|f_n^*\|_{\cH_{\tilde K_S}(\Omega)}^{\frac{p}{2}}.
\end{align}
Combining \eqref{eq:lemapp1fubX21} and Lemma \ref{lem:approx1}, we have
\begin{align}\label{eq:lemapp1fubX211}
    % \|f-\hat f\|_n^2 = O_{\PP}\left((\lambda_n^{-1}T)^{\frac{p}{(2+p)}}n^{-\frac{2}{2+p}}\right),\quad \|\hat f\|_{\cH_{\tilde K_S}(\Omega)}^2= O_{\PP}\left(\lambda_n^{-1}\left((\lambda_n^{-1}T)^{\frac{p}{(2+p)}}n^{-\frac{2}{2+p}}\right)^{\frac{1}{2}-\frac{p}{4}}(\lambda_n^{-1}T)^{\frac{p}{4}}n^{-1/2}\right),
    \|f-\hat f\|_n^2 = & O_{\PP}\left(\sigma_n^{-d/2 - \frac{pD}{4}}p^{-(D+1)/2}n^{-1/2}(\lambda_n^{-1}T)^{\frac{p}{2}}(T+n^{-1/2}T^{1/2})^{1-\frac{p}{2}}\right),\nonumber\\ 
    \|\hat f\|_{\cH_{\tilde K_S}(\Omega)}^2= &  O_{\PP}\left(\lambda_n^{-1} \sigma_n^{-d/2 - \frac{pD}{4}}p^{-(D+1)/2}n^{-1/2}(\lambda_n^{-1}T)^{\frac{p}{2}}(T+n^{-1/2}T^{1/2})^{1-\frac{p}{2}}\right).
\end{align}
Combining \eqref{eq:lemapp1fubX22} and Lemma \ref{lem:approx1}, we have
\begin{align}\label{eq:lemapp1fubX221}
    \|f-\hat f\|_n = & O_{\PP}\left( \sigma_n^{-d/2 - \frac{pD}{4}}p^{-(D+1)/2}n^{-1/2})^{\frac{2}{2+p}}(\lambda_n^{-1}T)^{\frac{p}{2+p}}\right)\nonumber\\
   \|\hat f\|_{\cH_{\tilde K_S}(\Omega)}^2= & O_{\PP}\left(\lambda_n^{-1/2}(\sigma_n^{-d/2 - \frac{pD}{4}}p^{-(D+1)/2}n^{-1/2})^{\frac{2}{2+p}}(\lambda_n^{-1}T)^{\frac{p}{2+p}}\right).
\end{align}
By \eqref{eq:lemapp1fub1}, \eqref{eq:lemapp1fub2121}, \eqref{eq:lemapp1fub2122f}, \eqref{eq:lemapp1fubX211}, and \eqref{eq:lemapp1fubX221}, we finish the proof.\hfill\BlackBox

% Combining \eqref{eq:lemapp1fub211}, \eqref{eq:lemapp1fub222}, \eqref{eq:lemapp1fubX211}, and \eqref{eq:lemapp1fubX221}, and setting $\lambda_n \asymp n^{-\frac{2m_f+2m_0}{2m_f+d}}$, $\sigma_n\asymp n^{-\frac{1}{2m_f + d}}$, $p= \frac{d}{m_f-m_0}$, we have 
% \begin{align}\label{eq:lemapp1fubXX}
%     \|f-\hat f\|_n^2 = O_{\PP}\left(n^{-\frac{2m_f}{2m_f+d}}\right),\quad\|\hat f\|_{\cH_{\tilde K_S}(\Omega)}^2 = O_{\PP}\left(n^{-\frac{2m_0}{2m_f+d}}\right).
% \end{align}
% It remains to bound the difference between $\|f-\hat f\|_n^2$ and $\|f-\hat f\|_{L_2(P_\Xb)}^2$.

% ...

\subsection{Proof of Lemma \ref{lem:ineqGRKHS}}\label{app:pflemineqGRKHS}
For any function $g\in \cH_\sigma(\RR^D)$, the Fourier inversion theorem implies
\begin{align}\label{eq:pflemem1}
    |g(\bx)| = & \left|\int_{\RR^D}e^{i\bx^T\bomega}\cF(g)(\bomega) d\bomega\right| \leq \int_{\RR^D}\left|\cF(g)(\bomega) \right| d\bomega\nonumber\\
    = & \int_{\RR^D}\left|\cF(g)(\bomega) \right|^{1-r}(\cF(k_\sigma)(\bomega))^{r/2}\left|\cF(g)(\bomega) \right|^r(\cF(k_\sigma)(\bomega))^{-r/2} d\bomega\nonumber\\
    \leq & \left(\int_{\RR^D}\left|\cF(g)(\bomega) \right|^{\frac{2(1-r)}{2-r}}(\cF(k_\sigma)(\bomega))^{\frac{r}{2-r}}d\bomega\right)^{\frac{2-r}{2}}\left(\int_{\RR^D}\left|\cF(g)(\bomega) \right|^2(\cF(k_\sigma)(\bomega))^{-1} d\bomega\right)^{\frac{r}{2}}\nonumber\\
    \leq & 2^{\frac{Dr}{2}}\sigma^{\frac{Dr}{2}}\left(\int_{\RR^D}\left|\cF(g)(\bomega) \right|^{\frac{2(1-r)}{2-r}}e^{-\frac{r}{2-r}\sigma^2\|\bomega\|_2^2}d\bomega\right)^{\frac{2-r}{2}}\|g\|_{\cH_\sigma(\RR^D)}^r\nonumber\\
    \leq & 2^{\frac{Dr}{2}}\sigma^{\frac{Dr}{2}}\left(\int_{\RR^D}\left|\cF(g)(\bomega) \right|^2 d\bomega \right)^{\frac{1-r}{2}}\left(\int_{\RR^D}e^{-r\sigma^2\|\bomega\|_2^2}d\bomega\right)^{\frac{1}{2}}\|g\|_{\cH_\sigma(\RR^D)}^r\nonumber\\
    = & 2^{\frac{Dr}{2}}\sigma^{\frac{Dr}{2}}(4 \pi^{-1}r\sigma^2)^{-\frac{D}{4}}\|g\|_{L_2(\RR^D)}^{1-r}\|g\|_{\cH_\sigma(\RR^D)}^r\nonumber\\
    \leq & C_1r^{-\frac{D}{4}}\sigma^{\frac{D(r-1)}{2}}\|g\|_{L_2(\RR^D)}^{1-r}\|g\|_{\cH_\sigma(\RR^D)}^r\nonumber\\
    % |\langle f, k_\sigma(\bx - \cdot)\rangle_{\cH_\sigma(\RR^D)}| \nonumber\\
    % = & \left|\int_{\RR^D} \frac{\cF(g)(\bomega)\cF(k_\sigma(\bx - \cdot))(\bomega)}{\cF(k_\sigma)(\bomega)}d\bomega\right|\nonumber\\
    % = & \left|\int_{\RR^D} \frac{\cF(g)(\bomega)\cF(k_\sigma(\bx - \cdot))(\bomega)}{\cF(k_\sigma)(\bomega)}d\bomega\right|\nonumber\\
\end{align}
where the second and fourth inequalities are by H\"older's inequality, and the third equality is by Parseval's identity. This finishes the proof.
\hfill\BlackBox
% Taking $r=\frac{p}{2d}$ finishes the proof.

\section{Proof of Lemmas in Appendix \ref{app:pfthmsoboNN_tensor}}

\subsection{Proof of Lemma \ref{lem:approx1NN_tensor_1}}\label{app:pfapprox1NN_tensor_1}

By following the similar approach in Appendix \ref{app:pfapprox1NN}, we have
\begin{align}\label{eq:pflemapprox1NNb1_tensor}
    \|f^*-f_n^*\|_{L_2(P_\Xb)}^2 + \lambda_n\|f_n^*\|_{\cH_{\tilde K_S}(\Omega)}^2
    \leq & \max(C_1,1)\left(\|f^*-\tilde f_n^*\|_{L_2(\RR^D)}^2 + \lambda_n\|\tilde f_n^*\|_{\cH_{\tilde K_S}(\RR^D)}^2\right).
\end{align}
Therefore, it remains to bound 
$$\|f^*-\tilde f_n^*\|_{L_2(\RR^D)}^2 + \lambda_n\|\tilde f_n^*\|_{\cH_{\tilde K_S}(\RR^D)}^2.$$

Similar to Appendix \ref{app:pfapprox1NN}, we can use the Fourier inversion theorem to get
\begin{align}\label{eq:pfapx1Sobonoise1_tensor_1_1}
    & \|f^*-\tilde f_n^*\|_{L_2(\RR^D)}^2 + \lambda_n\|\tilde f_n^*\|_{\cH_{\tilde K_S}(\RR^D)}^2 = \int_{\RR^D} |\mathcal{F}(f^*)(\bomega)-\mathcal{F}(\tilde f_n^*)(\bomega)|^2 + \lambda_n \frac{|\mathcal{F}(\tilde f_n^*)(\bomega)|^2}{\cF(\tilde K_S(\bx))(\bomega)}{\rm d}\bomega\nonumber\\
    \leq &  \int_{\RR^D} \frac{C_2\lambda_n\prod_{j=1}^D(1+\omega_j^2)^{m_0}(1+\sigma_n^2w_j^2)^{m_\varepsilon}}{1+C_2\lambda_n\prod_{j=1}^D(1+\omega_j^2)^{m_0}(1+\sigma_n^2w_j^2)^{m_\varepsilon}}|\mathcal{F}(f^*)(\bomega)|^2{\rm d}\bomega \nonumber\\
   \leq & \sum_{|\bl|\geq 1}I^{<}_{\bl}+I^{\geq}_{\bl}
\end{align}
where $\bl=(l_1,...,l_D)\in\{0,1\}^D$,
% and based on $\bl$, we can divide $\RR^D$ in to the following regions:
\begin{align*}
    &\Omega_{l_j}=\begin{cases}
        & \{\omega_j: \sigma_n^2\omega_j^2<1\},\quad\text{if}\ l_j=0,\\
        & \{\omega_j: \sigma_n^2\omega_j^2\geq1\},\quad\text{otherwise},
    \end{cases}\\
    &\Omega^{<}_{\bl}=\left[\times_{j=1}^D\Omega_{l_j}\right]\bigcap \{\bomega: C_2\lambda_n\prod_{j=1}^D(1+\omega_j^2)^{m_0}(1+\sigma_n^2w_j^2)^{m_\varepsilon}<1 \},\\
    &\Omega^{\geq}_{\bl}=\left[\times_{j=1}^D\Omega_{l_j}\right]\bigcap \{\bomega: C_2\lambda_n\prod_{j=1}^D(1+\omega_j^2)^{m_0}(1+\sigma_n^2w_j^2)^{m_\varepsilon}\geq 1 \},\\
% \end{align*}
% \begin{align*}
    &I_{\bl}^{<}=\int_{\Omega_{\bl}^<} {C_2\lambda_n\left[\prod_{j=1}^D(1+\omega_j^2)^{m_0}(1+\sigma_n^2w_j^2)^{m_\varepsilon}\right]}|\mathcal{F}(f^*)(\bomega)|^2{\rm d}\bomega,\\
    &I_{\bl}^{\geq}=\int_{\Omega_{\bl}^\geq} |\mathcal{F}(f^*)(\bomega)|^2{\rm d}\bomega
\end{align*}
and the sum over all $\{|\bl|\geq 1\}$ is because on any $\Omega_{\bl}^<$ and $\Omega_{\bl}^\geq$, there must be at least one  $j^*$ and one $j^{**}$ such that $\sigma_n^2w_{j^*}<1$ and  $\sigma_n^2w_{j^{**}}\geq 1$, respectively.

Define $p=\frac{m_f}{m_0+m_\varepsilon}\leq 1$. On any $\Omega_{\bl}^{<}$, we have
\begin{align*}
     &\quad {C_2\lambda_n \prod_{j=1}^D(1+\omega_j^2)^{m_0}(1+\sigma_n^2w_j^2)^{m_\varepsilon} }\\
     \leq & \left({C_2 \prod_{j=1}^D\lambda_n^{\frac{1}{D}}(1+\omega_j^2)^{m_0}(1+\sigma_n^2w_j^2)^{m_\varepsilon} }\right)^p\\
    = & C_3\prod_{j=1}^D\lambda_n^{\frac{p}{D}}(1+\omega_j^2)^{m_0p}(1+\sigma_n^2w_j^2)^{m_\varepsilon p}\\
    \leq & C_4 \prod_{j=1}^D \left(\lambda_n^{\frac{p}{D}}(1+\omega_j^2)^{m_0p}\right)^{1-l_j}\left(\lambda_n^{\frac{p}{D}}(1+\omega_j^2)^{m_0p}(\sigma_n^2w_j^2)^{m_\varepsilon p}\right)^{l_j}
\end{align*}
From the fact that $m_0p=m_f\frac{m_0}{m_0+m_\varepsilon}\leq m_f $ and calculations similar to \eqref{eq:pfapx1SobonoiseI2}, we have
\begin{align*}
    & \lambda_n^{\frac{p}{D}}(1+\omega_j^2)^{m_0p}\leq \lambda_n^{\frac{p}{D}}(1+\omega_j^2)^{m_f} &&\text{when}\ l_j=0,\\
{\rm and~~~}    & \lambda_n^{\frac{p}{D}}(1+\omega_j^2)^{m_0p}(\sigma_n^2w_j^2)^{m_\varepsilon p}\leq (\lambda_n^{\frac{1}{D}}\sigma_n^{2m_\varepsilon})^p (1+\omega_j^2)^{m_f} &&\text{when}\ l_j=1.
\end{align*}
As a result, on $\Omega^{<}_{\bl}$, we have
\begin{align}\label{eq:pfapx1Sobonoise1_tensor_1_2}
    {C_2\lambda_n \prod_{j=1}^D(1+\omega_j^2)^{m_0}(1+\sigma_n^2w_j^2)^{m_\varepsilon} } &\leq C_4\prod_{j=1}^D\left( \lambda_n^{\frac{p}{D}}\right)^{1-l_j}\left(\lambda_n^{\frac{1}{D}}\sigma_n^{2m_\varepsilon}\right)^{pl_j}(1+\omega_j^2)^{m_f}\nonumber\\
    &=C_4\lambda_n^p \sigma_n^{2m_\varepsilon p|\bl|}\prod_{j=1}^D(1+\omega_j^2)^{m_f},
\end{align}
where $|\bl| = \sum_{j=1}^D l_j$.

On $\Omega^{\geq}_{\bl}$, we have
\begin{align}\label{eq:pfapx1Sobonoise1_tensor_1_3}
    1\leq &\left(C_2\lambda_n\prod_{j=1}^D(1+\omega_j^2)^{m_0}(1+\sigma_n^2\omega_j^2)^{m_\varepsilon}\right)^p\nonumber\\
    \leq & \left(C_5\lambda_n\prod_{j=1}^D(1+\omega_j^2)^{m_0}(\sigma_n^2\omega_j^2)^{m_\varepsilon l_j}\right)^p\nonumber\\
    \leq & C_6\lambda_n^p\sigma_n^{2m_\varepsilon p |\bl|}\prod_{j=1}^D(1+\omega_j)^{m_f}.
\end{align}

Plugging \eqref{eq:pfapx1Sobonoise1_tensor_1_2} and \eqref{eq:pfapx1Sobonoise1_tensor_1_3} into \eqref{eq:pfapx1Sobonoise1_tensor_1_1} finishes the proof.\hfill\BlackBox

\subsection{Proof of Lemma \ref{lem:approx2NN_tensor}}\label{app:pfapprox2NN_tensor}

Let $\Psi_{\sigma}(\|\cdot\|)\coloneqq \prod_{j=1}^D\psi_\sigma(|\cdot|)$ be tensor product of positive definite functions with 
\begin{align*}
    c_1(1+\sigma^2|\omega_j|^2)^{-(m_0+m_\varepsilon)} \leq  \cF(\psi_{\sigma})\leq c_2(1+\sigma^2|\omega_j|^2)^{-(m_0+m_\varepsilon)}, \forall \bomega\in\mathbb{R}^D, \forall j=1,\ldots, d
\end{align*}
and $\cN_{\sigma}(\Omega)$ be the RKHS generated by $\Psi_{\sigma}$. We will use the following lemmas. Lemma \ref{lem:ENofSobonoise_tensor} can be derived by Corollary A.8 of \cite{hamm2021adaptive} and (6.6) of \cite{dung2018hyperbolic}. Lemma \ref{lem:Berforsingleg_tensor} is a direct result of the proof of Lemma 8.4 of \cite{geer2000empirical} and Lemma \ref{lem:ENofSobonoise_tensor}. 

\begin{lemma}\label{lem:ENofSobonoise_tensor}
Let $4\sigma^2\leq 1$. Suppose the conditions of Lemma \ref{lem:approx2NN_tensor} are fulfilled. Let $\cB_{\cN_\sigma(\Omega)}$ be a unit ball in $\cN_{\sigma}(\Omega)$. Then there exists a constant $C_1>0$ only depending on $D$ and $\Omega$ such that for all $\delta>0$, we have
\begin{align*}
    H(\delta,\cB_{\cH_\sigma(\Omega)},\|\cdot\|_{L_\infty(\Omega)})\leq C_1\sigma^{-d}\delta^{-\frac{1}{m_0+m_\varepsilon}}|\log \delta|^{(D-1)+\frac{1}{2(m_0+m_\varepsilon)}}.
\end{align*}
\end{lemma}

\begin{lemma}\label{lem:Berforsingleg_tensor}
Suppose conditions of Theorem \ref{thm:soboNN_tensor} are fulfilled. Then for any $T$ large enough
we have 
\begin{align*}
    \PP\left(\sup_{g\in \cB_{\cH_\sigma(\Omega)}}\frac{\sqrt{n}\langle g, \bepsilon\rangle_n}{\|g\|_n^{1-p}|\log\|g\|_n |^{(D-1+p)/2}}\geq T\right)\leq C_2\exp\left(-\frac{T^2}{C_3}\right).
\end{align*}
where $p=\frac{1}{2(m_0+m_\varepsilon)}$, $C_2$ and $C_3$ are some constant independent of $T$ and $n$.
\end{lemma}

% \noindent\textit{Proof:} Lemma \ref{lem:Berforsingleg_tensor} can be proved by slight modification of Lemma 8.4 of \cite{geer2000empirical}.

\noindent\textit{Proof of Lemma \ref{lem:Berforsingleg_tensor}.} From Lemma \ref{lem:ENofSobonoise_tensor}, we can derive that for any $\delta\leq 1$,
\[\int_0^\delta H(u,\cB_{\cH_\sigma(\Omega)},\|\cdot\|_{L_\infty(\Omega)})^{1/2}{\rm d}u\lesssim\sigma^{-d}\delta^{1-p}\big|\log \delta\big|^{\frac{D-1+p}{2}}. \]
Then, by Corollary 8.3 of \cite{geer2000empirical},
we can derive that
\[\PP\left(\sup_{g\in \cB_{\cH_\sigma(\Omega)}}\sqrt{n}\big|\langle g, \bepsilon\rangle_n\big|\geq \sigma^{-d}\delta^{1-p}\big|\log \delta\big|^{\frac{D-1+p}{2}}\right)\lesssim \exp\left(-C_4\sigma_n^{-2d}\delta^{-2p}|\log \delta|^{D-1+p}\right).\]
We then can follow the peeling-off argument in Lemma 8.4 of \cite{geer2000empirical} to show  
\begin{align*}
    &\quad \PP\left(\sup_{g\in \cB_{\cH_\sigma(\Omega)}}\frac{\sqrt{n}\langle g, \bepsilon\rangle_n}{\|g\|_n^{1-p}|\log\|g\|_n |^{(D-1+p)/2}}\geq T\right)\\
    \leq & \sum_{s=1}^\infty  \PP\left(\sup_{g\in \cB_{\cH_\sigma(\Omega)}, \|g\|_n\leq 2^{-s+1}}\sqrt{n}\langle g, \bepsilon\rangle_n\geq T 2^{-s(1-p)}s^{\frac{D-1+p}{2}}\right)\\
    \lesssim & \sum_{s=1}^\infty \exp\left(-C_4T^2\sigma_n^{-2d}2^{4ps}|\log 2|^{D-1+p}\right)\\
    \lesssim & \sum_{s=1}^\infty \exp\left(-C_4T^2s\right)\\
    =& C_2 \exp\left(-\frac{T^2}{C_3}\right).
\end{align*}
\hfill\BlackBox

\noindent\textit{Proof  of Lemma \ref{lem:approx2NN_tensor}.} We can follow the proof of Lemma \ref{lem:approx2NN} to derive the following inequality using Lemmas \ref{lem:ENofSobonoise_tensor} and \ref{lem:Berforsingleg_tensor}:
\begin{align}\label{eq:lemapprx2Sobonoisebi2_tensor}
    & \|f^*-\hat f_n\|_n^2 + \lambda_n \|\hat f_n\|_{\cH_{\tilde K_S}(\Omega)}^2\nonumber\\
    \leq &  \|f^*-f_n^*\|_n^2 + \lambda_n \|f_n^*\|_{\cH_{\tilde K_S}(\Omega)}^2\nonumber\\
    &\quad + O_{\PP}(n^{-1/2})\sigma_n^{-d/2}\|\hat f_n - f_n^*\|_n^{1-\frac{p}{2}}|\log \|\hat f_n - f_n^*\|_n|^{\frac{D-1}{2}+\frac{p}{4}}(\|\hat f_n\|_{\cN_{\sigma_n}(\Omega)} + \|f_n^*\|_{\cN_{\sigma_n}(\Omega)})^{\frac{p}{2}},
\end{align}
where $p=\frac{1}{m_0+m_\varepsilon}$.
Notice that \eqref{eq:lemapprx2Sobonoisebi2_tensor} is similar to \eqref{eq:lemapprx2Sobonoisebi2} in  the proof of Lemma \ref{lem:approx2NN} except for the extra poly-log term $|\log \|\hat f_n - f_n^*\|_n|^{\frac{D-1}{2}+\frac{p}{4}}$. However, the extra poly-log term will not change the case-by-case analysis in our proof because it is always dominated by those polynomial terms in \eqref{eq:lemapprx2Sobonoisebi2_tensor}. Therefore, we can follow the same logic in the proof of Lemma \ref{lem:approx2NN} to get the final results.
% \section{Other notes}
% This should be used in the approximation error.
% Let $g=f_n^* - f$.

\subsection{Proof of Lemma \ref{lem:ineqTensorRKHS}}\label{app:pflemineqTensorRKHS}
For any function $g\in \cMW^m(\RR^D)$, the Fourier inversion theorem implies
\begin{align}\label{eq:pflemem1}
    |g(\bx)| = & \left|\int_{\RR^D}e^{i\bx^T\bomega}\cF(g)(\bomega) d\bomega\right| \leq \int_{\RR^D}\left|\cF(g)(\bomega) \right| d\bomega\nonumber\\
    = & \int_{\RR^D}\left|\cF(g)(\bomega) \right|^{1-r}(\cF(k_\sigma)(\bomega))^{r/2}\left|\cF(g)(\bomega) \right|^r(\cF(k_\sigma)(\bomega))^{-r/2} d\bomega\nonumber\\
    \leq & \left(\int_{\RR^D}\left|\cF(g)(\bomega) \right|^{\frac{2(1-r)}{2-r}}(\cF(k_\sigma)(\bomega))^{\frac{r}{2-r}}d\bomega\right)^{\frac{2-r}{2}}\left(\int_{\RR^D}\left|\cF(g)(\bomega) \right|^2(\cF(k_\sigma)(\bomega))^{-1} d\bomega\right)^{\frac{r}{2}}\nonumber\\
    \leq & \left(\int_{\RR^D}\left|\cF(g)(\bomega) \right|^{\frac{2(1-r)}{2-r}}\left|\prod_{j=1}^D(1+\omega_j^2)^{-m}\right|^{\frac{r}{2-r}}d\bomega\right)^{\frac{2-r}{2}}\|g\|_{\cH_\sigma(\RR^D)}^r\nonumber\\
    \leq & \left(\int_{\RR^D}\left|\cF(g)(\bomega) \right|^2 d\bomega \right)^{\frac{1-r}{2}}\left(\prod_{j=1}^D\int_{\RR}(1+\omega_j^2)^{-mr}d\bomega\right)^{\frac{1}{2}}\|g\|_{\cH_\sigma(\RR^D)}^r\nonumber\\
    = & C_r\|g\|_{L_2(\RR^D)}^{1-r}\|g\|_{\cH_\sigma(\RR^D)}^r\nonumber\\
    % |\langle f, k_\sigma(\bx - \cdot)\rangle_{\cH_\sigma(\RR^D)}| \nonumber\\
    % = & \left|\int_{\RR^D} \frac{\cF(g)(\bomega)\cF(k_\sigma(\bx - \cdot))(\bomega)}{\cF(k_\sigma)(\bomega)}d\bomega\right|\nonumber\\
    % = & \left|\int_{\RR^D} \frac{\cF(g)(\bomega)\cF(k_\sigma(\bx - \cdot))(\bomega)}{\cF(k_\sigma)(\bomega)}d\bomega\right|\nonumber\\
\end{align}
where the second and fourth inequalities are by H\"older's inequality, and the third equality is by Parseval's identity. $C_r<\infty$ for any $r>m^{-1}/2$. This finishes the proof.
\hfill\BlackBox
% \section{Proof of Lemmas in }

\section{Proof of Lemma \ref{lem:ENofSobonoise}}

% Let $\cB_{\cN_\sigma(\Omega)}$ be a unit ball in $\cN_{\sigma}(\Omega)$ (see Appendix \ref{app:pfapprox2NN} for the definition of $\cN_{\sigma}(\Omega)$). 

\begin{lemma}\label{lem:EnofSovolev_m}
Let the RKHS $\cH_m$ induced by the kernel function $K_m$ be equipped with norm satisfying 
\[\|f\|^2_{\cH_m}\leq C\int_{\RR^D}\left(1+\frac{\|\bomega\|^2}{m}\right)^m\left|\hat{f}(\bomega)\right|^2{\rm d}\bomega,\]
where $C$ is some constant independent of $m$. Then for any $m>D/2$, there exists a constant $C'$ independent of $m$ such that for all $\delta>0$, we have
\[H(\delta,\cB_{\cH_m([0,1]^D)},\|\cdot\|_{L_\infty([0,1]^D)})\leq C' (2m-D)^{-\frac{2D}{2m-D}}m^{\frac{2mD}{2m-D}}\delta^{-\frac{2D}{2m-D}}\log(1+\delta^{-1}). \]
\end{lemma}

\begin{remark}
If we treat $m$ as a constant, the upper bound in Lemma \ref{lem:EnofSovolev_m} is larger than that in \eqref{eq:EnofSovolev}. However, in the proofs of Lemmas \ref{lem:ENofSobonoise} and \ref{lem:approx2NN}, it turns out that the upper bound in Lemma \ref{lem:EnofSovolev_m} is sufficient.
\end{remark}

\textit{Proof of Lemma \ref{lem:ENofSobonoise}.} The proof follows Corollary A.8 of \cite{hamm2021adaptive}. Specifically, Corollary A.8 of \cite{hamm2021adaptive} states that for any $\delta>0$ annd $\sigma>0$, it holds that
\begin{align*}
    H(\delta,\cB_{\cH_\sigma(\Omega)},\|\cdot\|_{L_\infty(\Omega)})\leq & \cN_{\ell_\infty^D}(\sigma,\Omega) H(\delta,\cB_{\cH_m([0,1]^D)},\|\cdot\|_{L_\infty(\Omega)}),
\end{align*}
which, by Assumption \ref{assum:intriD} and Lemma \ref{lem:EnofSovolev_m}, leads to
\begin{align*}
    H(\delta,\cB_{\cH_\sigma(\Omega)},\|\cdot\|_{L_\infty(\Omega)})\leq & C\sigma^{-d}(2m-D)^{-\frac{2D}{2m-D}}m^{\frac{2mD}{2m-D}}\delta^{-\frac{2D}{2m-D}}\log(1+\delta^{-1}),
\end{align*}
where the constant $C$ is independent with $m_\varepsilon$, and $m=m_\varepsilon + m_0$.
\hfill\BlackBox

\section{Proof of Lemma \ref{lem:EnofSovolev_m}}
For any $f\in \cH_m([0,1]^D)$, we have the following representation of $f$ by Fourier series
\[f=\sum_{\bzeta\in\NN^D}f_{\bzeta} \psi_{\bzeta}\]
where $\psi_{\bzeta}$ is the Fourier basis associated to $\bzeta$ and $f_{\bzeta}$ is the projection of $f$ on $\psi_{\bzeta}$. Then transference from $L_2(\RR^D)$ to $L_2([0,1]^D)$ by Fourier multiplier (see theorem 3.4 in \cite{l1977transference})
shows that the RKHS norm of $f$ embedded on $[0,1]^D$ can be written as
\[\|f\|_{\cH_m}^2\leq\sum_{\bzeta\in\NN^D}(1+\frac{\|\bzeta\|^2}{m})^mf_{\bzeta}^2.\]

We first define a projection $P_M$ as follows:
\[P_Mf=\sum_{\bzeta\in[M]^D}f_{\bzeta}\psi_{\bzeta}.\]
Then for the embedding operator $\cI:\cH([0,1]^D)\to L_\infty([0,1]^D)$, we have
\begin{align}
    \|\cI\|^2=&\sup_{f\in\cB_{\cH_m([0,1]^D)}}\sup_{\bx\in[0,1]^D}|f(\bx)|^2\nonumber\\
    &\leq 2\sup_{f\in\cB_{\cH_m([0,1]^D)}}\sup_{\bx\in[0,1]^D}|P_Mf(\bx)|^2+2\sup_{f\in\cB_{\cH_m([0,1]^D)}}\sup_{\bx\in[0,1]^D}|f(\bx)-P_Mf(\bx)|^2\label{eq:EnofF_Sobolev_m_1}.
\end{align}
For the first term of \eqref{eq:EnofF_Sobolev_m_1}, it is obvious that
\[2\sup_{f\in\cB_{\cH_m([0,1]^D)}}\sup_{\bx\in[0,1]^D}|P_Mf(\bx)|^2\leq 2\|\cI\|^2\leq 2K_m(\bx,\bx). \]
For the second term of \eqref{eq:EnofF_Sobolev_m_1}, we have
\begin{align*}
    \|\cI-P_M\| =& \sup_{f\in\cB_{\cH_m([0,1]^D)}}\sup_{\bx\in[0,1]^D}|f(\bx)-P_Mf(\bx)|\\
    %=& \sup_{f\in\cB_{\cH_m([0,1]^D)}}\sup_{\bx\in[0,1]^D}|\sum_{\bzeta\in\NN^D-[M]^D}f_{\bzeta}\psi_{\bzeta}(\bx)|\\
    \leq & \sup_{f\in\cB_{\cH_m([0,1]^D)}}\sum_{\bzeta\in\NN^D-[M]^D}|f_{\bzeta}|\\
    \leq & \left(\sum_{\bzeta\in\NN^D-[M]^D}(1+\frac{\|\bzeta\|^2}{m})^{-m}\right)^\frac{1}{2}
\end{align*}
where the last line is from H\"older inequality and  $\forall f\in \cB_{\cH_m([0,1]^D)}$, $\|f\|_{\cH_m([0,1]^D)}\leq 1$.

Notice that for $m>D/2$, we have
\begin{align*}
    \sum_{\bzeta\in\NN^D-[M]^D}(1+\frac{\|\bzeta\|^2}{m})^{-m}&=\sum_{\zeta_1\geq M+1}\cdots\sum_{\zeta_D\geq M+1}\left(1+\frac{\sum_{j=1}^D\zeta_j^2}{m}\right)^{-m}\\
    &\leq \underbrace{\int_{M}^{\infty}\cdots\int_{M}^{\infty}}_{D\text{\ terms}}(1+\frac{\|\bzeta\|^2}{m})^{-m}{\rm d}\bzeta\\
    & = \int_{0}^{2\pi}\cdots\int_{0}^{2\pi}\int_{M}^{\infty}(1+\frac{r^2}{m})^{-m}\det(J(r,\boldsymbol{\theta})){\rm d}r{\rm d}\boldsymbol{\theta}\\
    & \leq  \int_{0}^{2\pi}\cdots\int_{0}^{2\pi}\int_{M}^{\infty}(1+\frac{r^2}{m})^{-m}r^{D-1}{\rm d}r{\rm d}\boldsymbol{\theta}\\
    & \leq C\frac{1}{2m-D} m^{m} M^{-2m + D}.
\end{align*}
%Multiply $\|\cI-P_M\|^2$ by $(1+M/m)^{2m}$, we get
%\begin{align}
%    \sup_{m>D/2}(1+\frac{M}{m})^{2m}\|\cI-P_M\|^2=& \sup_{m>D/2}  \sum_{\bzeta\in\NN^D-[M]^D}\left(\frac{m^2}{m^2+M^2+2mM}\frac{m+\|\bzeta\|^2}{m}\right)^{-m}\nonumber\\
%    =& \sup_{m>D/2}\sum_{\bzeta\in\NN^D-[M]^D}\left(\frac{m^2/M^2+1+2m/M}{m^2/M^2+m\|\bzeta\|^2/M^2}\right)^m\nonumber\\
%    \leq & \sup_{m>D/2}\sum_{\bzeta\in\NN^D}\left(\frac{m/M^2+2}{m/M^2+\|\bzeta\|^2}\right)^m\nonumber \\
%    \leq & \sup_{m>D/2}\sum_{\bzeta\in\NN^D}\left(\frac{3}{1+\|\bzeta\|^2/m}\right)^m\label{eq:Enof_Sobolev_m_2}
%\end{align}
%Integral test of the last line gives the Mat\'ern kernel with smoothness parameter $m-D/2$, which is bounded for any $m\in(D/2,\infty]$. Therefore, 
Therefore, we can conclude that $\|\cI-P_M\|\leq C\frac{1}{2m-D} m^{m} M^{-2m + D}$ for come $C$ independent of $m$. Given any $\delta>0$, we can select integer $M=\lceil \left((2m-D)m^{-m}\delta\right)^{-\frac{2}{2m-D}}\rceil$ so that
\[\|\cI-P_M\|\leq \delta,\]
where $\lceil r\rceil$ denotes the ceiling round up of $r$. Then we can apply Lemma 1 in \cite{kuhn2011covering} to get
\begin{align*}
    H(\delta,\cB_{\cH_m([0,1]^D)},\|\cdot\|_{L_\infty([0,1]^D)})\leq &\text{rank}(P_M)\log(1+\delta^{-1})\\
    \leq& M^D\log(1+\delta^{-1})\\
    \leq& C'\left((2m-D)m^{-m}\delta\right)^{-\frac{2D}{2m-D}}\log(1+\delta^{-1})\\
    = & C' (2m-D)^{-\frac{2D}{2m-D}}m^{\frac{2mD}{2m-D}}\delta^{-\frac{2D}{2m-D}}\log(1+\delta^{-1}),
\end{align*}
for some $C'$ independent of $m$.\hfill\BlackBox

\section{Appendix for Detailed Experiments}\label{app:experiments}

% As stated in Section \ref{sec:num}, we adopt two-hidden-layer neural networks with ReLU activation as our predictor. Each hidden layer of the neural network has 100 nodes, and all weights are initialized by Kaiming Initialization \citep{he2015delving}. We try both non-smooth Laplace noise and smooth Gaussian noise for random smoothing. Specifically, each element of $\bvarepsilon_k$ is sampled from either $\mathcal{N}(0,\sigma^2)$ or $Laplace(0,b)$. 
% For convenience, $\sigma$ and $b$ are called the smoothing scale in this section. 

In this section, we present more details of numerical experiments conducted in Section \ref{sec:num}.

Note that in the experiments, our goal is specified by minimizing the $l_2$ loss in the form of \eqref{eq:loss}. We train the neural network using stochastic gradient descent (SGD) with momentum (0.9), small batch size (10), and learning rate $\beta=0.01$. We choose a constant weight decay strength ($10^{-4}$) to focus on the influence of random smoothing in cases with weight decay.
We set the number of augmented samples $N=1000$ and conduct a grid search for the smoothing scale from 0 to 0.6. The simulated data are divided into the training set, validation set, and test set. 
The validation set is sampled as half the size of the training set, while the size of the test set is fixed at 500. The test results are selected based on the validation set unless otherwise specified and we repeat each experiment 15 times and report the average loss on the test set. 

Considering stochastic gradient descent with weight decay, we adopt a candidate list of weight decay strength $\{10^{-3}, 10^{-4}, 10^{-5}\}$. To make a fair comparison, we choose a consistent number of iterations instead of epochs for different training sizes, i.e., given a batch size, the number of epochs gets smaller when the training size becomes larger. Specifically, the number of iterations in cases with weight decay is 10,000. For early stopping without weight decay, we evaluate the validation error every 200 gradient descent steps during training and select the model with the smallest validation error. The maximal step for SGD with early stopping is 100,000. We repeat each experiment 15 times and report the average loss on the test set.

\end{document}